\documentclass[10pt,twocolumn,twoside]{IEEEtran}

\usepackage{epsfig}
\usepackage{graphicx}
\usepackage{amsmath}
\usepackage{amssymb}
\usepackage{algorithm}
\usepackage{algorithmicx}
\usepackage{algpseudocode}
\usepackage{subfig}
\usepackage{xcolor}
\usepackage{caption}
\def\etal{et al. }
\def\eg{e.g. }

\begin{document}

\title{Video Tracking Using Learned Hierarchical Features}

\author{Li~Wang,~\IEEEmembership{Member,~IEEE,}
        Ting~Liu,~\IEEEmembership{Student Member,~IEEE,}
        Gang~Wang,~\IEEEmembership{Member,~IEEE,}
        ~Kap~Luk~Chan,~\IEEEmembership{Member,~IEEE,}
        and Qingxiong Yang,~\IEEEmembership{Member,~IEEE}
        
\thanks{L. Wang, T. Liu, G. Wang and K.L. Chan are with the School of Electrical and Electronic Engineering, Nanyang Technological University, Singapore (e-mail: wa0002li@e.ntu.edu.sg; liut0016@e.ntu.edu.sg; wanggang@ntu.edu.sg; eklchan@ntu.edu.sg).}
\thanks{G. Wang is also with the Advanced Digital Science Center, Singapore.}
\thanks{Q. Yang is with the Department of Computer Science, City University of Hong Kong, China (e-mail: qiyang@cityu.edu.hk).}}

\maketitle

\begin{abstract}
In this paper, we propose an approach to learn hierarchical features for visual object tracking. First, we offline learn features robust to diverse motion patterns from auxiliary video sequences. The hierarchical features are learned via a two-layer convolutional neural network. Embedding the temporal slowness constraint in the stacked architecture makes the learned features robust to complicated motion transformations, which is important for visual object tracking. Then, given a target video sequence, we propose a domain adaptation module to online adapt the pre-learned features according to the specific target object. The adaptation is conducted in both layers of the deep feature learning module so as to include appearance information of the specific target object. As a result, the learned hierarchical features can be robust to both complicated motion transformations and appearance changes of target objects. We integrate our feature learning algorithm into three tracking methods. Experimental results demonstrate that significant improvement can be achieved by using our learned hierarchical features, especially on video sequences with complicated motion transformations.
\end{abstract}


\IEEEpeerreviewmaketitle

\section{Introduction}

\IEEEPARstart{L}{earning} hierarchical feature representation (also called deep learning) has emerged recently as a promising research direction in computer vision and machine learning. Rather than using hand-crafted features, deep learning aims to learn data-adaptive, hierarchical, and distributed representation from raw data. The learning process is expected to extract and organize discriminative information from data. Deep learning has achieved impressive performance on image classification \cite{DBLP:conf/nips/KrizhevskySH12}, action recognition \cite{DBLP:conf/cvpr/LeZYN11}, and speech recognition \cite{hinton2012deep}, etc.

Feature representation is an important component for visual object tracking. Deep learning usually requires a lot of training data to learn deep structure and its related parameters. However, in visual tracking, only the annotation of the target object in the first frame of the test video sequence is available. Recently, Wang and Yeung \cite{DBLP:conf/nips/WangY13} have proposed a so-called deep learning tracker (DLT). They propose to offline learn generic features from auxiliary natural images. However, using unrelated images for training, they cannot obtain deep features with temporal invariance, which is actually very important for visual object tracking. Moreover, they do not have an integrated objective function to bridge offline training and online tracking. They transfer knowledge from offline training to online tracking by simply feeding the deep features extracted from the pre-trained encoder to the target object classifier and tune the parameters of the pre-trained encoder when significant changes of object appearances are detected.

\begin{figure*}[t]
    \begin{center}
    \epsfig{file=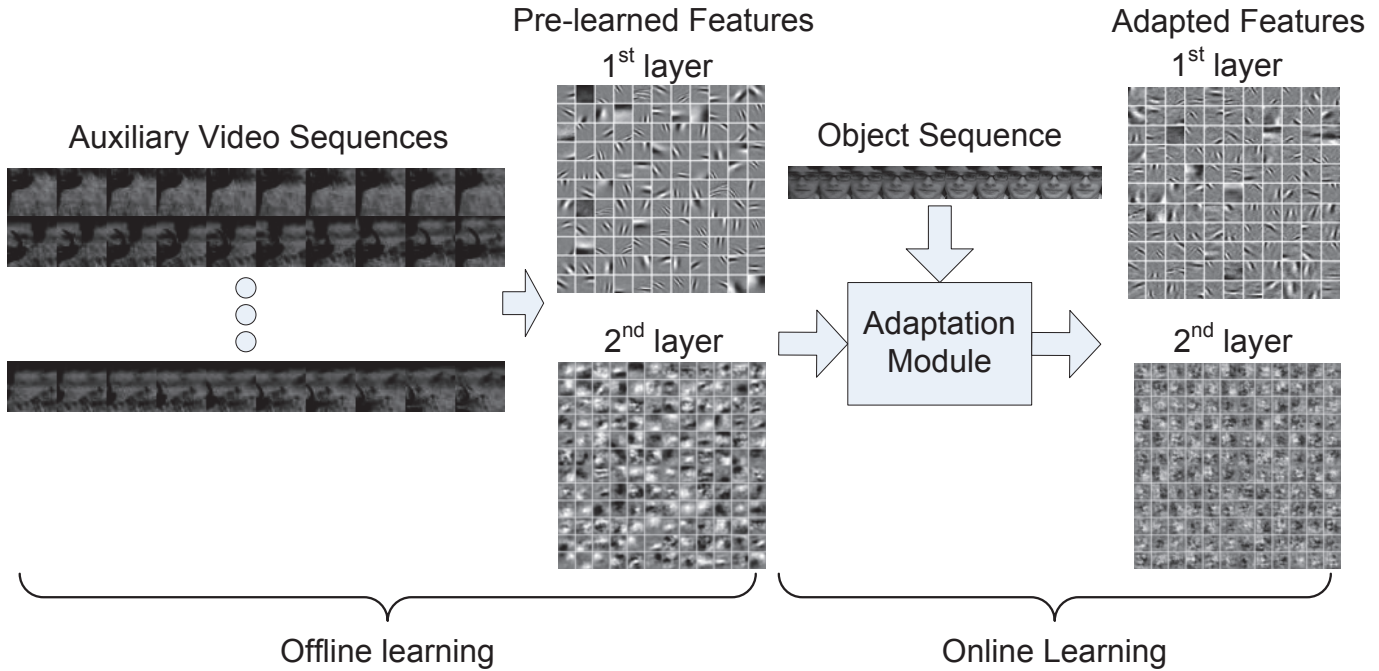,width=1.0\textwidth}
    \end{center}
    \caption{Overview of the proposed feature learning algorithm. First, we pre-learn generic features from auxiliary data obtained from Hans van Hateren natural scene videos \cite{DBLP:conf/nips/CadieuO08}. A number of learned feature filters from two layers are visualized. Then, we adapt the generic features to a specific object sequence. The adapted feature filters are also visualized, from which we can find that the adapted features are more relevant to the specific object ``face" as they contain more facial edges and corners in the first layer and more semantic elements which look like faces or face parts in the second layer.}
    \label{fig:flowchart}
\end{figure*}

To address these two issues in DLT \cite{DBLP:conf/nips/WangY13}, we propose a domain adaptation based deep learning method to learn hierarchical features for model-free object tracking. Figure~\ref{fig:flowchart} presents an overview of the proposed feature learning method. First, we aim to learn deep features robust to complicated motion transformations of the target object, which are not considered by DLT \cite{DBLP:conf/nips/WangY13}. Also, we intend to learn features which can handle a wide range of motion patterns in the test video sequences. Therefore, we adopt the feature learning method proposed in Zou \etal \cite{DBLP:conf/nips/ZouNZY12} as a basic model to pre-learn features robust to diverse motion patterns from auxiliary video sequences (offline learning part shown in Figure~\ref{fig:flowchart}). Given the corresponding patches in the training video sequences, the basic model learns patch features invariant between two consecutive frames. As a result, high-level features which are robust to non-linear motion patterns can be discovered. Zou \etal \cite{DBLP:conf/nips/ZouNZY12} employ the learned features for generic object
recognition. We argue that this method is also beneficial to object tracking, as temporal robustness can help a tracker to find corresponding patches reliably.

As stated above, Wang and Yeung \cite{DBLP:conf/nips/WangY13} do not have an extra united objective function connecting offline learning and online tracking. As a result, the learned features from their method do not include appearance information of specific target objects. To solve this issue, we propose a domain adaptation module to effectively adapt the pre-learned features according to the specific target object (online learning part shown in Figure~\ref{fig:flowchart}). The adaptation module is seamlessly incorporated into both layers of the stacked architecture of our deep learning model. As a result, the adapted features can be robust to both complicated motion transformations and appearance changes of the target object. As shown in Figure~\ref{fig:flowchart}, we can observe that the adapted features are more relevant to the specific object ``face" as they contain more facial edges and corners in the first layer and more semantic elements which look like faces or face parts in the second layer. 

In order to capture appearance changes of specific target objects, we online adapt pre-learned generic features according to the new coming data of the test video sequence. Due to high dimensions of the parameter space in our deep learning model, we employ the limited memory BFGS (L-BFGS) algorithm \cite{nocedal1980updating} to solve the optimization problem in the adaptation module. As a result, convergence can be quickly reached in each adaptation.

We validate the proposed method on benchmark test video sequences. Experimental results demonstrate that significant improvement can be obtained by using our learned hierarchical features for object tracking.

\section{Related Work}
\noindent \textbf{Object tracking} For decades, many interesting methods have been proposed for object tracking which has a wide range of applications, \eg video surveillance \cite{DBLP:journals/tifs/LuWM14} \cite{DBLP:conf/cvpr/WangWCW14}. Eigentracker \cite{DBLP:journals/ijcv/BlackJ98} has had a deep impact on subspace based trackers \cite{DBLP:conf/cvpr/HoLYK04} \cite{DBLP:journals/ijcv/RossLLY08}. The method named as ``Condensation" \cite{DBLP:journals/ijcv/IsardB98} is well-known because it is the first one to apply particle filter \cite{doucet2001introduction} to object tracking. In \cite{DBLP:journals/pami/ComaniciuRM03}, mean-shift \cite{DBLP:journals/pami/ComaniciuM02} is used to optimize the target localization problem in visual tracking. The ``Lucas-Kanade" algorithm \cite{DBLP:journals/ijcv/BakerM04} is famous for defining the cost function by using the sum of squared difference (SSD). Another pioneering method \cite{DBLP:journals/pami/JepsonFE03} paves the way for the subsequent trackers based on the adaptive appearance model (AAM).

Recently, the tracking problem has also been considered as a binary classification problem due to the significant improvement on object recognition \cite{DBLP:journals/pami/WangHF12} \cite{DBLP:journals/pami/WangFH13}. In \cite{DBLP:journals/pami/Avidan04}, the Support Vector Machine (SVM) is integrated into an optical-flow based tracker. Subsequently, the ensemble tracker \cite{DBLP:conf/cvpr/Avidan05} trains an ensemble of weak classifiers online to label pixels as objects or backgrounds. In \cite{DBLP:conf/eccv/GrabnerLB08}, an online semi-supervised boosting method is proposed to handle the drifting problem caused by inaccuracies from updating the tracker. In \cite{DBLP:conf/cvpr/BabenkoYB09}, on-line multiple instance learning is also proposed to solve the drifting problem. P-N learning \cite{DBLP:conf/cvpr/KalalMM10} is proposed to train a binary classifier from labeled and unlabeled examples which are iteratively corrected by positive (P) and negative (N) constraints. Also, correlation filters \cite{DBLP:conf/cvpr/LiuWY15} have achieved very promising results for visual object tracking.

Many advanced trackers are also developed based on sparse representation \cite{DBLP:conf/cvpr/YangYGH09}. $\ell_1$ tracker \cite{DBLP:conf/iccv/MeiL09} solves an $\ell_1$-regularized least squares problem to achieve the sparsity for target candidates, in which the one with the smallest reconstruction error is selected as the target in the next frame. Two pieces of works \cite{DBLP:conf/cvpr/MeiLWBB11} \cite{DBLP:conf/cvpr/LiSS11} focus on accelerating the $\ell_1$ tracker \cite{DBLP:conf/iccv/MeiL09} because the $\ell_1$ minimization requires high computational costs. There are some other promising sparse trackers \cite{DBLP:conf/eccv/LiuYHMGK10} \cite{DBLP:conf/cvpr/LiuHYK11} \cite{DBLP:conf/cvpr/JiaLY12}. The tracker \cite{DBLP:conf/cvpr/JiaLY12} employing the adaptive structural local sparse appearance model (ASLA) achieves especially good performance, and this is used as the baseline tracking system in this paper.

\noindent \textbf{Feature representation} Some tracking methods focus on feature representation. In \cite{DBLP:journals/pami/CollinsLL05}, an online feature ranking mechanism is proposed to select features which are capable of discriminating between object and background for visual tracking. Similarly, an online AdaBoost feature selection algorithm is proposed in \cite{DBLP:conf/bmvc/GrabnerGB06} to handle appearance changes in object tracking. In \cite{DBLP:conf/cvpr/GrabnerGB07}, keypoint descriptors in the region of the interested object are learned online together with background information. The compressive tracker (CT)
\cite{DBLP:conf/eccv/Zhang0Y12} employs random projections to extract data independent features for the appearance model and separates objects from backgrounds using a naive Bayes classifier. Recently, Wang and Yeung \cite{DBLP:conf/nips/WangY13} has proposed to learn deep compact features for visual object tracking.

\noindent \textbf{Deep learning} Deep learning \cite{DBLP:journals/neco/HintonOT06} \cite{hinton2006reducing} has recently attracted much attention in machine learning. It has been successfully applied in many computer vision applications, such as visual object tracking \cite{DBLP:journals/tip/WangLWCY15}, shape modeling \cite{DBLP:conf/cvpr/EslamiHW12}, action recognition \cite{DBLP:journals/pami/ShahroudyWNY15}, image set classification \cite{DBLP:conf/cvpr/LuWDMZ15}, attribute prediction \cite{DBLP:journals/tmm/AbdulnabiWLJ15}, face recognition \cite{DBLP:journals/tifs/LuLWM15}, scene image classification \cite{DBLP:journals/pr/ZuoWSZY15} and scene labeling \cite{DBLP:conf/cvpr/ShuaiWZWZ15} \cite{DBLP:journals/tip/0001LCWC15}. Deep learning aims to replace hand-crafted features with high-level and robust features learned from raw pixel values, which is also known as unsupervised feature learning \cite{DBLP:conf/nips/CoatesKN12} \cite{DBLP:conf/nips/CoatesN11} \cite{DBLP:conf/nips/LeKNN11} \cite{DBLP:conf/icml/LeRMDCCDN12}. In \cite{DBLP:conf/nips/ZouNZY12}, the temporal slowness constraint \cite{li2008unsupervised} is combined with deep neural networks to learn hierarchical features. Inspired by this work, we intend to learn deep features to handle complicated motion transformations in visual object tracking.

\noindent \textbf{Domain adaptation} Recently, there have been increasing interests in visual domain adaptation problems. Saenko \etal \cite{DBLP:conf/eccv/SaenkoKFD10} apply domain adaptation to learn object category models. In \cite{DBLP:conf/mm/YangYH07}, domain adaptation techniques are developed to detect video concepts. In \cite{DBLP:conf/cvpr/DuanXTL10}, Duan \etal adapt learned models from web data to recognize visual events. Recently, Glorot \etal \cite{glorot2011domain} develop a meaningful representation for large-scale sentiment classification by combining deep learning and domain adaptation. Domain adaptation has also been applied in object tracking. Wang \etal \cite{DBLP:journals/tip/WangCYXY12} pre-learn an over-complete dictionary and transfer the learned visual prior for tracking specific objects.

\noindent \textbf{Our method} The principles behind our method are deep learning and domain adaptation learning. We first utilize the temporal slowness constraint to offline learn generic hierarchical features robust to complicated motion transformations. Then, we propose a domain adaptation module to adapt the pre-learned features according to the specific target object. The differences between DLT \cite{DBLP:conf/nips/WangY13} and our method are as follows. First, their method pre-learns features from untracked images. In contrast, our method uses tracked video sequences and focuses on learning features robust to complex motion patterns. Second, their method does not have a united objective function with the regularization term for domain adaptation, whereas our method has an adaptation module integrating the specific target object's appearance information into the pre-learned generic features. Our method is also different from \cite{DBLP:journals/tip/WangCYXY12}, in which the dictionary is pre-defined and the tracking object is reconstructed by the patterns in the pre-defined dictionary. The method in \cite{DBLP:journals/tip/WangCYXY12} may fail if the pre-defined dictionary does not include the visual patterns of the target object. Last, it is necessary to mention that Zou \etal \cite{DBLP:conf/nips/ZouNZY12} learn hierarchical features from video sequences with tracked objects for image classification whereas our method focuses on visual object tracking.

\section{Tracking System Overview}
We aim to learn hierarchical features to enhance the state-of-the-art tracking methods. The tracking system with the adaptive structural local sparse appearance model (ASLA) \cite{DBLP:conf/cvpr/JiaLY12} achieves very good performance. Hence, we integrate our feature learning method into this system. But note that our feature learning method is general for visual tracking, and it can be used with other tracking systems as well by replacing original feature representations.

In this section, we briefly introduce the tracking system. Readers may refer to \cite{DBLP:conf/cvpr/JiaLY12} for more details. Suppose we have an observation set of target $x_{1:t}=\{x_1, \dots, x_t\}$ up to the $t^{th}$ frame and a corresponding feature representation set $z_{1:t}=\{z_1, \dots, z_t\}$, we can calculate the target state $y_t$ as follows

\begin{equation}
y_t = \arg \max_{y_t^i} p\left( y_t^i | z_{1:t} \right )
\end{equation}

\noindent where $y_t^i$ denotes the state of the $i^{th}$ sample in the $t^{th}$ frame. The posterior probability $p\left( y_t | z_{1:t} \right)$ can be inferred by the Bayes' theorem as follows

\begin{equation}
p\left( y_t | z_{1:t} \right)\varpropto p\left( z_t | y_t \right)\int p\left( y_t | y_{t-1} \right) p\left( y_{t-1} | z_{1:t-1} \right) dy_{t-1} \label{equ:postp}
\end{equation}

\noindent where $z_{1:t}$ denotes the feature representation, $p\left( y_t | y_{t-1} \right)$ denotes the motion model and $p\left( z_t | y_t \right)$ denotes the appearance model. In \cite{DBLP:conf/cvpr/JiaLY12}, the representations $z_{1:t}$ simply use raw pixel values. In contrast, we propose to learn hierarchical features from raw pixels for visual tracking.

\section{Learning Features for Video Tracking}
Previous tracking methods usually use raw pixel values or hand-crafted features to represent target objects. However, such features cannot capture essential information which is invariant to non-rigid object deformations, in-plane and out-of-plane rotations in object tracking. We aim to enhance tracking performance by learning hierarchical features which have the capability of handling complicated motion transformations. To achieve this, we propose a domain adaptation based feature learning algorithm for visual object tracking. We first adopt the approach proposed in \cite{DBLP:conf/nips/ZouNZY12} to learn features from auxiliary video sequences offline. These features are robust to complicated motion transformations. However, they do not include appearance information of specific target objects. Hence, we further use a domain adaptation method to adapt pre-learned features according to specific target objects.

We integrate our feature learning method into the tracking system ASLA \cite{DBLP:conf/cvpr/JiaLY12} and its details are given in Algorithm~\ref{alg:our_trac_sys}.

\begin{algorithm}[h]
\caption{\textbf{Our tracking method}} \label{alg:our_trac_sys}
\begin{algorithmic}[1]

\State \textbf{Input:} the previous tracking state $y_{t-1}$, the existing feature learning parameter $\hat{\Theta}$ and the exemplar library.

\State Apply the affine transformation on $y_{t-1}$ to obtain a number of tracking states $y_{t}^i$ and the corresponding candidate image patches $x_t^i$.

\State Extract feature representations $z_t^i$ from the candidate image patches $x_t^i$ under the existing feature learning parameter $\hat{\Theta}$.

\State Calculate the posterior probability $p\left( y_t^i | z_{1:t} \right)$ according to Equation~\ref{equ:postp}.

\State Predict the tracking state by $y_t = \arg \max_{y_t^i} p\left( y_t^i | z_{1:t} \right )$.

\State Update the feature learning parameter and the exemplar library every $M$ frames.

\State \textbf{Output:} the predicted tracking state $y_{t}$, the up-to-date feature learning parameter $\Theta$ and the up-to-date exemplar library.

\end{algorithmic}
\end{algorithm}

\begin{figure*}[t]
    \begin{center}
    \epsfig{file=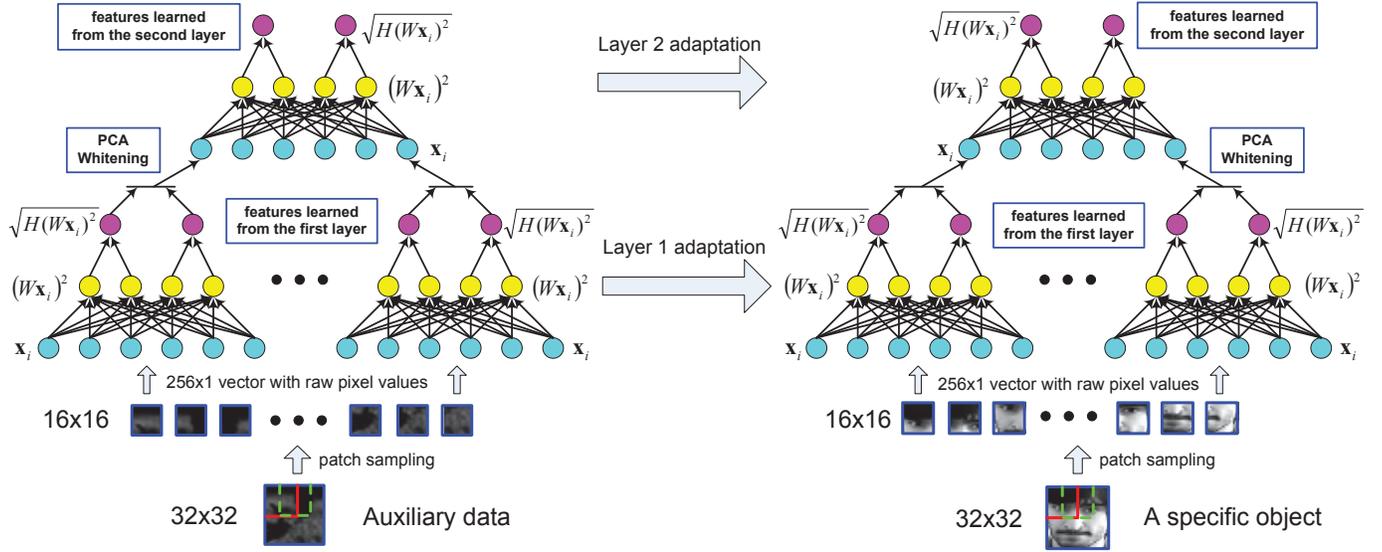,width=1.0\textwidth}
    \end{center}
    \caption{Stacked architecture of our deep feature learning algorithm. The output of the first layer is whitened using PCA and then used as the input of the second layer. For the adaptation module, given a specific object sequence, the pre-learned features learned from auxiliary data are adapted respectively in two layers by minimizing the objective function in Equation~\ref{equ:our_equ}.}
    \label{fig:learn_feature}
\end{figure*}

\subsection{Pre-Learning Generic Features from Auxiliary Videos}
Since the appearance of an object could change significantly due to its motion, a good tracker desires features robust to motion transformations. Inspired by \cite{DBLP:conf/nips/ZouNZY12}, we believe that there exist generic features robust to diverse motion patterns. Therefore, we employ the deep learning model in \cite{DBLP:conf/nips/ZouNZY12} to learn hierarchical features from auxiliary videos \cite{DBLP:conf/nips/CadieuO08} to handle diverse motion transformations of objects in visual tracking. Note that this is performed offline.

The deep learning model has two layers as illustrated in Figure~\ref{fig:learn_feature}. In our case, the first layer works on smaller patches ($16 \times 16$). The second layer works on larger patches ($32 \times 32$). We learn the feature transformation matrix $W$ of each layer as below.

Given the offline training patch $\mathbf{x}_i$ from the $i^{th}$ frame, we denote the corresponding learned feature as $\mathbf{z}_i=\sqrt{H(W\mathbf{x}_i)^2}$, where $H$ is the pooling matrix and $(W\mathbf{x}_i)^2$ is the element-wise square on the output of the linear network layer. To better explain the basic learning module in each layer, we make use of the illustration in Figure~\ref{fig:learn_feature}. First, it is necessary to mention that the blue, yellow and purple circles denote the input vector $\mathbf{x}_i$, the intermediate vector $(W\mathbf{x}_i)^2$ and the output vector $\sqrt{H(W\mathbf{x}_i)^2}$ respectively w.r.t. the basic learning module. Then, $H$ can be illustrated as the transformations between the intermediate vector (yellow circles) and the output one (purple circles). The pooling mechanism is to calculate the summation of two adjacent feature dimensions of the intermediate vector (yellow circles) in a non-overlapping fashion. Also, $W$ can be illustrated as the transformations between the input vector (blue circles) and the intermediate one (yellow circles). Essentially, each row of the feature transformation matrix $W$ can be converted to an image patch filter as shown in Figure~\ref{fig:flowchart}. The feature transformation matrix $W$ is learned by solving the following unconstrained minimization problem,

\begin{equation}
\min_W \lambda \sum_{i=1}^{N-1} \Vert \mathbf{z}_i-\mathbf{z}_{i+1} \Vert_1 + \sum_{i=1}^N \Vert \mathbf{x}_i - W^TW\mathbf{x}_i \Vert_2^2 \label{equ:zou_equ},
\end{equation}

\noindent where $\mathbf{z}_{i+1}$ denotes the learned feature from the $(i+1)^{th}$ frame and $N$ is the total length of all video sequences in the auxiliary data. Essentially, multiple video sequences are organized sequence-by-sequence. Between two sequences, our learning algorithm does not take into account the differences between the non-continuous frames, the last frame $\mathbf{z}_i$ of the current sequence and the first frame $\mathbf{z}_{i+1}$ of the next sequence. The first term forces learned features to be temporally continuous and the second term is an auto-encoder reconstruction cost \cite{DBLP:conf/nips/LeKNN11}. As a result, we obtain the feature $\mathbf{z}$ which is robust to complicated motion transformations.

The input of the first layer is the raw pixel values of smaller patches ($16\times16$). We can learn the feature transformation matrix $W^{L1}$ for the first layer by Equation~\ref{equ:zou_equ}. Then, we apply $W^{L1}$ to convolve with the larger patches ($32 \times 32$). The larger patch is divided into a number of sub-patches ($16\times16$). We use $W^{L1}$ to conduct feature mapping for each sub-patch and concatenate features of all the sub-patches to represent the larger patch. Next, PCA whitening is applied to the concatenated feature vector. Finally, we use the whitened feature vector of the larger patch as the input to the second layer and learn the feature transformation matrix $W^{L2}$ for the second layer.

The first layer can extract features robust to local motion patterns \eg translations. From the second layer, we could extract features robust to more complicated motion transformations \eg non-linear warping and out-of-plane rotations (See Figure~\ref{fig:flowchart}). We concatenate features from two layers as our generic features. Moreover, we pre-learn the generic features from a lot of auxiliary video data. As a result, the pre-learned features can provide our tracker with capabilities of handling diverse motion patterns.

\subsection{Domain Adaption Module}
Although the generic features are robust to non-linear motion patterns in visual tracking, they do not include appearance information of specific target objects, \eg shape and texture. Hence, we propose a domain adaptation module to adapt the generic features according to specific target objects. The domain adaptation module is illustrated in Figure~\ref{fig:learn_feature}.

Given a target video sequence, we employ ASLA \cite{DBLP:conf/cvpr/JiaLY12} to track the target object in the first $N$ frames and use the tracking results as the training data for the adaptation module. The adapted feature is denoted as $\mathbf{z}_i^{adp}=\sqrt{H(W\mathbf{x}_i^{obj})^2}$, where $\mathbf{x}_i^{obj}$ indicates the object image patch in the $i^{th}$ frame of the training data for adaptation and $W$ is the feature transformation matrix to be learned. We formulate the adaptation module by adding a regularization term as follows,

\begin{eqnarray}
\nonumber W_{adp}&=&\arg\min_W \lambda \sum_{i=1}^{N-1} \Vert \mathbf{z}_i^{adp}-\mathbf{z}_{i+1}^{adp} \Vert_1 \\ \nonumber &+&\gamma \sum_{i=1}^N \Vert W \mathbf{x}_i^{obj} - W_{old} \mathbf{x}_i^{obj} \Vert_2^2 \\ &+&\sum_{i=1}^N \Vert \mathbf{x}_i^{obj} - W^TW \mathbf{x}_i^{obj} \Vert_2^2 \label{equ:our_equ},
\end{eqnarray}

\noindent where $W_{old}$ denotes the pre-learned feature transformation matrix. The second term refers to the adaptation module and aims to make the adapted feature close to the old one for the sake of preserving the pre-learned features' robustness to complicated motion transformations. Meanwhile, using the training data $\mathbf{x}_i^{obj}$ is intended to include the appearance information of the specific target object, \eg shape and texture. $\gamma$ is the trade-off parameter which controls the adaptation level.

We adapt the generic features in a two-layer manner. It means that we conduct the minimization in Equation~\ref{equ:our_equ} with respect to $W$ in both layers respectively.

\subsection{Optimization and Online Learning}
Succinctly, we denote the objective function of the adaptation module as $f(\mathbf{X};\Theta,\hat{\Theta})$, where $\mathbf{X}$ denotes a number of training images of object regions for the adaptation, $\Theta=\{w_{ij}|i,j=1,\dots,N\}$ indicates the parameter set representing all entries in the transformation matrix $W$ and $\hat{\Theta}$ refers to the known parameter set w.r.t. $W_{old}$. We employ limited-memory BFGS (L-BFGS) algorithm \cite{nocedal1980updating} to optimize the objective function $f(\mathbf{X};\Theta,\hat{\Theta})$ w.r.t. the parameter set $\Theta$.

\begin{algorithm}[h]
\caption{\textbf{Calculation on L-BFGS search direction $p_k$}} \label{alg:L-BFGS_2loop}
\begin{algorithmic}[1]
\State \textbf{Input:} the derivative $\nabla f_k$ of the objective function $f$ w.r.t. $\Theta_k$, the curvature information from $m$ most recent iterations $\{s_i,y_i|i=k-m,\dots,k-1\}$.

\State $p_k=-\nabla f_k$;

\For{$i=k-1, k-2, \dots, k-m$}

\State $\alpha_i=\frac{s_i^T p_k}{y_i^T s_k}$;

\State $p_k=p_k-\alpha_i y_i$;

\EndFor

\State $p_k=B_0^{-1} p_k$

\For{$i=k-m,k-m+1,\dots,k-1$}

\State $\beta=\frac{y_i^T p_k}{y_i^T s_i}$;

\State $p_k=p_k+s_i(\alpha_i-\beta)$;

\EndFor

\State \textbf{Output:} L-BFGS search direction $p_k$

\end{algorithmic}
\end{algorithm}

The Quasi-Newton methods, such as BFGS algorithm \cite{Nocedal2006NO}, need to update the approximate Hessian matrix $B_k$ at the $i^{th}$ iteration to calculate the search direction $p_k=-B_k^{-1} \nabla f_k$, where $\nabla f_k$ is the derivative of the objective function $f$ w.r.t. $\Theta_k$ at the $k^{th}$ iteration. The cost of storing the approximate Hessian matrix $B_k$ ($N^2 \times N^2$) is prohibitive in our case because the dimension $N^2$ of the parameter set $\Theta$ is high ($\approx 10^4$). Therefore, we use L-BFGS in which the search direction $p_k$ is calculated based on the current gradient $\nabla f_k$ and the curvature information from $m$ most recent iterations $\{s_i=\Theta_{i+1}-\Theta_i,y_i=\nabla f_{i+1} - \nabla f_i |i=k-m,\dots,k-1\}$. Algorithm~\ref{alg:L-BFGS_2loop} presents calculation on L-BFGS search direction $p_k$. In our implementation, $m$ is set to $5$.

Given the search direction $p_k$ obtained from Algorithm~\ref{alg:L-BFGS_2loop}, we compute $\Theta_{k+1}=\Theta_{k}+\alpha_k p_k$, where $\alpha_k$ is chosen to satisfy the Wolfe conditions \cite{Nocedal2006NO}. When $k>m$, we discard the curvature information $\{s_{k-m},y_{k-m}\}$ and compute and save the new one $\{s_k=\Theta_{k+1}-\Theta_k,y_k=\nabla f_{k+1}-\nabla f_k$\}. Using L-BFGS to optimize the adaptation formulation, the convergence can be reached after several iterations.

To capture appearance changes of target objects, we online learn the parameter set $\Theta$ of the adaptation module every $M$ frames. We also use L-BFGS algorithm to solve the minimization problem $\arg min_{\Theta} f(\Theta;\mathbf{X},\tilde{\Theta})$, where $\mathbf{X}=\{\mathbf{x}_1:\mathbf{x}_M\}$ denotes training data within object regions from $M$ most recent frames and $\tilde{\Theta}$ indicates the old parameter set. The learned parameter set $\Theta$ converges quickly in the current group of $M$ frames and it will be used as the old parameter set $\tilde{\Theta}$ in the next group of M frames. In our implementation, $M$ is set to $20$ in all test video sequences.

\subsection{Implementation Details}
\emph{Auxiliary data} We pre-learn the generic features using the auxiliary data from Hans van Hateren natural scene videos \cite{DBLP:conf/nips/CadieuO08}. As mentioned in \cite{DBLP:conf/nips/ZouNZY12}, features learned from sequences containing tracked objects can encode more useful information such as non-linear warping. Hence, we employ video sequences containing tracked objects for pre-learning features (see Figure~\ref{fig:flowchart}).

\emph{Initialization} We use tracking results from ASLA \cite{DBLP:conf/cvpr/JiaLY12} in the first $20$ frames as the initial training data for our adaptation module. It is fair to compare with other methods under this setting. Many tracking methods have this sort of initialization. For example, Jia \etal \cite{DBLP:conf/cvpr/JiaLY12} utilize a k-d tree matching scheme to track target objects in first $10$ frames of sequences and then build exemplar libraries and patch dictionaries based on these tracking results.

\emph{Computational cost} Learning generic features consumes much time (about $20$ minutes) due to the large training dataset. However, it is conducted offline, hence it does not matter. For the online adaptation part, we initialize the transformation matrix $W$ to be learned with the pre-learned $W_{old}$. Based on the training data collected online, each update of the adaptation module takes only several iterations to achieve the convergence. Another part is feature mapping, in which it is required to extract features from candidate image patches. ASLA \cite{DBLP:conf/cvpr/JiaLY12} requires to sample $600$ candidate patches in each frame. We find that it is very expensive if we conduct feature mapping for all candidate patches. Therefore, we conduct a coarse-to-fine searching strategy, in which we first select a number of (\eg $20$) promising candidates in each frame according to the tracking result from ASLA \cite{DBLP:conf/cvpr/JiaLY12} using raw pixel values and then refine the ranking of candidates based on our learned hierarchical features. We run the experiments on a PC with a Quad-Core $3.30$ GHz CPU and $8$ GB RAM. However, we do not use the multi-core setting of the PC. The speed of our tracker (about $0.8$ fps) is roughly twice slower than the one of ASLA \cite{DBLP:conf/cvpr/JiaLY12} (about $1.6$ fps) due to the additional feature extraction step. The time (about $625$ ms) spent on the feature extraction is about same as on the other parts of our tracker. Note that the main objective here is to show that our learned hierarchical features can improve tracking accuracy. The efficiency of our tracker could be improved further because feature mapping for different patches could be conducted in parallel by advanced techniques, \eg GPU. Finally, we empirically tune the trade-off parameters of $\lambda$ and $\gamma$ in Equation~\ref{equ:our_equ} for different sequences. However, the parameters change in a small range. $\lambda$ and $\gamma$ are tuned in $[1,10]$ and $[90, 110]$ respectively.

\begin{table}
\caption{Average center error (in pixels). The best two results are shown in red and blue fonts. We compare our tracker using learned features with $4$ state-of-the-art trackers using other feature representations: the raw pixel values (ASLA\cite{DBLP:conf/cvpr/JiaLY12}\_RAW), the hand-crafted HOG feature (ASLA\cite{DBLP:conf/cvpr/JiaLY12}\_HOG), the sparse representation ($\ell_1$\_APG \cite{DBLP:conf/cvpr/BaoWLJ12}) and the data-independent feature (CT\_DIF \cite{DBLP:conf/eccv/Zhang0Y12}). We also present the results of the variant of our tracker (Ours\_VAR) which does not use the temporal slowness constraint in feature learning.}
\vspace{-0.1in}
\begin{center}
\tabcolsep 0.036in \scriptsize
\begin{tabular}{|c||c|c|c|c||c|c|}
\hline
\textbf{Sequence}&\tiny{\textbf{ASLA}\cite{DBLP:conf/cvpr/JiaLY12}\textbf{\_RAW}}&\tiny{\textbf{ASLA}\cite{DBLP:conf/cvpr/JiaLY12}\textbf{\_HOG}}&\tiny{\textbf{$\ell_1$\_APG}\cite{DBLP:conf/cvpr/BaoWLJ12}}&\tiny{\textbf{CT\_DIF}\cite{DBLP:conf/eccv/Zhang0Y12}}&\tiny{\textbf{Ours\_VAR}}&\tiny{\textbf{Ours}}\\
\hline \hline
Basketball& 70.2 & 245.6 & 107.2 & 123.6 & \textcolor{blue}{14.0} & \textcolor{red}{11.2} \\
Biker&88.0&68.2&79.6&80.7&\textcolor{blue}{19.1}&\textcolor{red}{11.7}\\
David2& 24.0& 25.1& 44.1&59.8&\textcolor{blue}{12.4}& \textcolor{red}{1.5}\\
FleetFace& 129.2& 180.0& 123.5& 145.5&\textcolor{blue}{30.2}& \textcolor{red}{24.8}\\
Freeman1& 79.2& 69.8& 21.7& 17.8&\textcolor{blue}{9.2}& \textcolor{red}{8.7}\\
Freeman3& 21.4& 43.4& 15.6& 65.6&\textcolor{blue}{6.5}& \textcolor{red}{4.4}\\
Kitesurf& 40.3& 38.5& 71.7& 62.1&\textcolor{blue}{22.6}& \textcolor{red}{13.2}\\
Lemming& 165.5& 155.4& 138.2& 149.3&\textcolor{blue}{8.6}& \textcolor{red}{6.7}\\
MountainBike& 185.9& 155.2& 210.1& 212.9&\textcolor{blue}{10.3}& \textcolor{red}{9.2}\\
Shaking& 86.5& 86.6& 113.0& \textcolor{blue}{30.9}&37.9& \textcolor{red}{15.2}\\
Skating1&14.6&15.6&72.3&87.9&\textcolor{blue}{6.6}&\textcolor{red}{6.5}\\
Sylvester& 27.2& 27.7& 31.6& \textcolor{blue}{17.6}&52.1& \textcolor{red}{6.4}\\
Tiger1& 71.2& 112.9& \textcolor{blue}{61.7}& 85.6&70.3& \textcolor{red}{40.9}\\
Tiger2& 61.8& 96.2& 58.4& 83.3&\textcolor{blue}{27.1}& \textcolor{red}{19.2}\\
Trellis& 31.9& 18.8& 62.5& 47.4&\textcolor{blue}{13.2}& \textcolor{red}{3.0}\\
\hline
Average& 73.1& 89.3& 80.7& 84.7&\textcolor{blue}{22.7}& \textcolor{red}{12.2}\\
\hline
\end{tabular}
\end{center}
\label{tab:ace_motion}
\end{table}

\begin{table}
\caption{Average overlap rate (\%). The best two results are shown in red and blue fonts. We compare our tracker using learned features with $4$ state-of-the-art trackers using other feature representations: the raw pixel values (ASLA\cite{DBLP:conf/cvpr/JiaLY12}\_RAW), the hand-crafted HOG feature (ASLA\cite{DBLP:conf/cvpr/JiaLY12}\_HOG), the sparse representation ($\ell_1$\_APG \cite{DBLP:conf/cvpr/BaoWLJ12}) and the data-independent feature (CT\_DIF \cite{DBLP:conf/eccv/Zhang0Y12}). We also present the results of the variant of our tracker (Ours\_VAR) which does not use the temporal slowness constraint in feature learning.}
\vspace{-0.1in}
\begin{center}
\tabcolsep 0.036in \scriptsize
\begin{tabular}{|c||c|c|c|c||c|c|}
\hline
\textbf{Sequence}&\tiny{\textbf{ASLA}\cite{DBLP:conf/cvpr/JiaLY12}\textbf{\_RAW}}&\tiny{\textbf{ASLA}\cite{DBLP:conf/cvpr/JiaLY12}\textbf{\_HOG}}&\tiny{\textbf{$\ell_1$\_APG}\cite{DBLP:conf/cvpr/BaoWLJ12}}&\tiny{\textbf{CT\_DIF}\cite{DBLP:conf/eccv/Zhang0Y12}}&\tiny{\textbf{Ours\_VAR}}&\tiny{\textbf{Ours}}\\
\hline \hline
Basketball&0.46& 0.27& 0.29&0.36&\textcolor{blue}{0.54}&\textcolor{red}{0.59} \\
Biker&\textcolor{blue}{0.52}&0.41&0.29&0.40&0.51&\textcolor{red}{0.67}\\
David2& 0.59& 0.26& 0.29&0.05&\textcolor{blue}{0.61}& \textcolor{red}{0.85}\\
FleetFace& 0.13& 0.13&0.41& 0.24&\textcolor{blue}{0.58}& \textcolor{red}{0.60}\\
Freeman1& 0.32& 0.28& 0.34& 0.27&\textcolor{blue}{0.45}& \textcolor{red}{0.51}\\
Freeman3& 0.47& \textcolor{blue}{0.52}& 0.45& 0.13&0.49& \textcolor{red}{0.59}\\
Kitesurf& 0.35& 0.36& 0.34& 0.24&\textcolor{blue}{0.50}& \textcolor{red}{0.63}\\
Lemming& 0.42& 0.41& 0.40& 0.41&\textcolor{blue}{0.69}& \textcolor{red}{0.76}\\
MountainBike& 0.27& 0.37& 0.40& 0.25&\textcolor{blue}{0.61}& \textcolor{red}{0.68}\\
Shaking& 0.10& 0.06& 0.06& 0.33&\textcolor{blue}{0.50}& \textcolor{red}{0.58}\\
Skating1&\textcolor{blue}{0.38}&0.35&0.31&0.26&0.35&\textcolor{red}{0.42}\\
Sylvester& 0.35& 0.47& 0.37& \textcolor{blue}{0.55}&0.39& \textcolor{red}{0.71}\\
Tiger1& 0.25& 0.24& \textcolor{blue}{0.42}& 0.15&0.27& \textcolor{red}{0.50}\\
Tiger2& 0.32& 0.21& 0.38& 0.10&\textcolor{blue}{0.44}& \textcolor{red}{0.53}\\
Trellis& \textcolor{blue}{0.49}& 0.49& 0.28& 0.28&\textcolor{blue}{0.69}& \textcolor{red}{0.79}\\
\hline
Average& 0.36& 0.32& 0.34& 0.27&\textcolor{blue}{0.51}&\textcolor{red}{0.63}\\
\hline
\end{tabular}
\end{center}
\label{tab:aor_motion}
\end{table}

\section{Experiments}
First, we evaluate our learned hierarchical features to demonstrate its robustness to complicated motion transformations. Second, we evaluate the temporal slowness constraint and the adaptation module in our feature learning algorithm. Third, we evaluate our tracker's capability of handling typical problems in visual tracking. Then, we compare our tracker with $14$ state-of-the-art trackers. Moreover, we present the comparison results between DLT \cite{DBLP:conf/nips/WangY13} and our tracker. Finally, we present the generalizability of our feature learning algorithm on the other $2$ tracking methods.

We use two measurements to quantitatively evaluate tracking performances. The first one is called center location error which measures distances of centers between tracking results and ground truths in pixels. The second one is called overlap rate which is calculated according to $\frac{area(R_T \cap R_G)}{area(R_T \cup R_G)}$ and indicates extent of region overlapping between tracking results $R_T$ and ground truths $R_G$. It is necessary to mention that there are often subjective biases in evaluating tracking algorithms as indicated in \cite{DBLP:conf/iccv/PangL13}.

\subsection{Evaluation on Our Learned Feature's Robustness to Complicated Motion Transformations}

\begin{figure*}
\centering \subfloat[Basketball]{
\epsfig{file=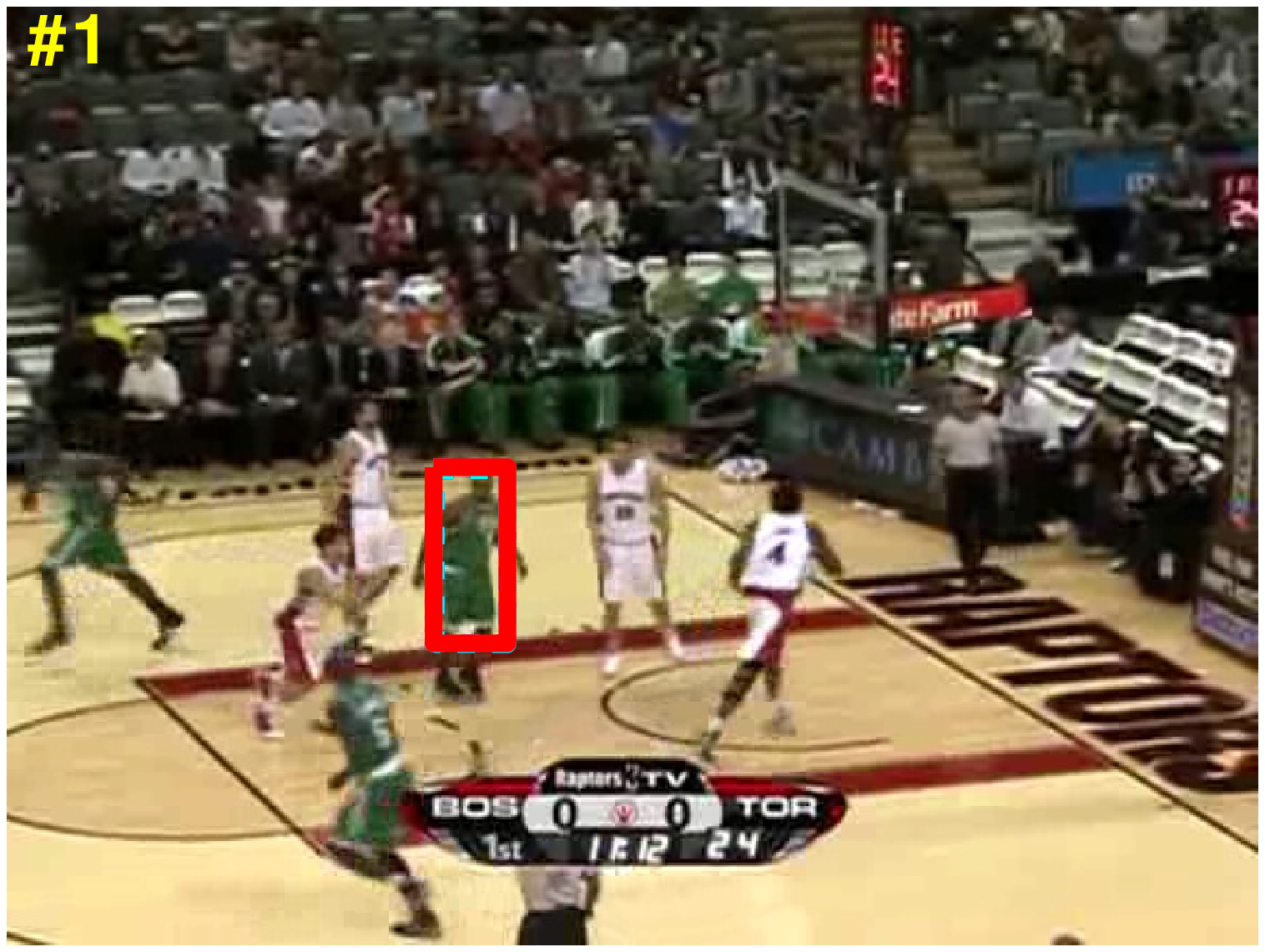,width=0.16\textwidth}
\epsfig{file=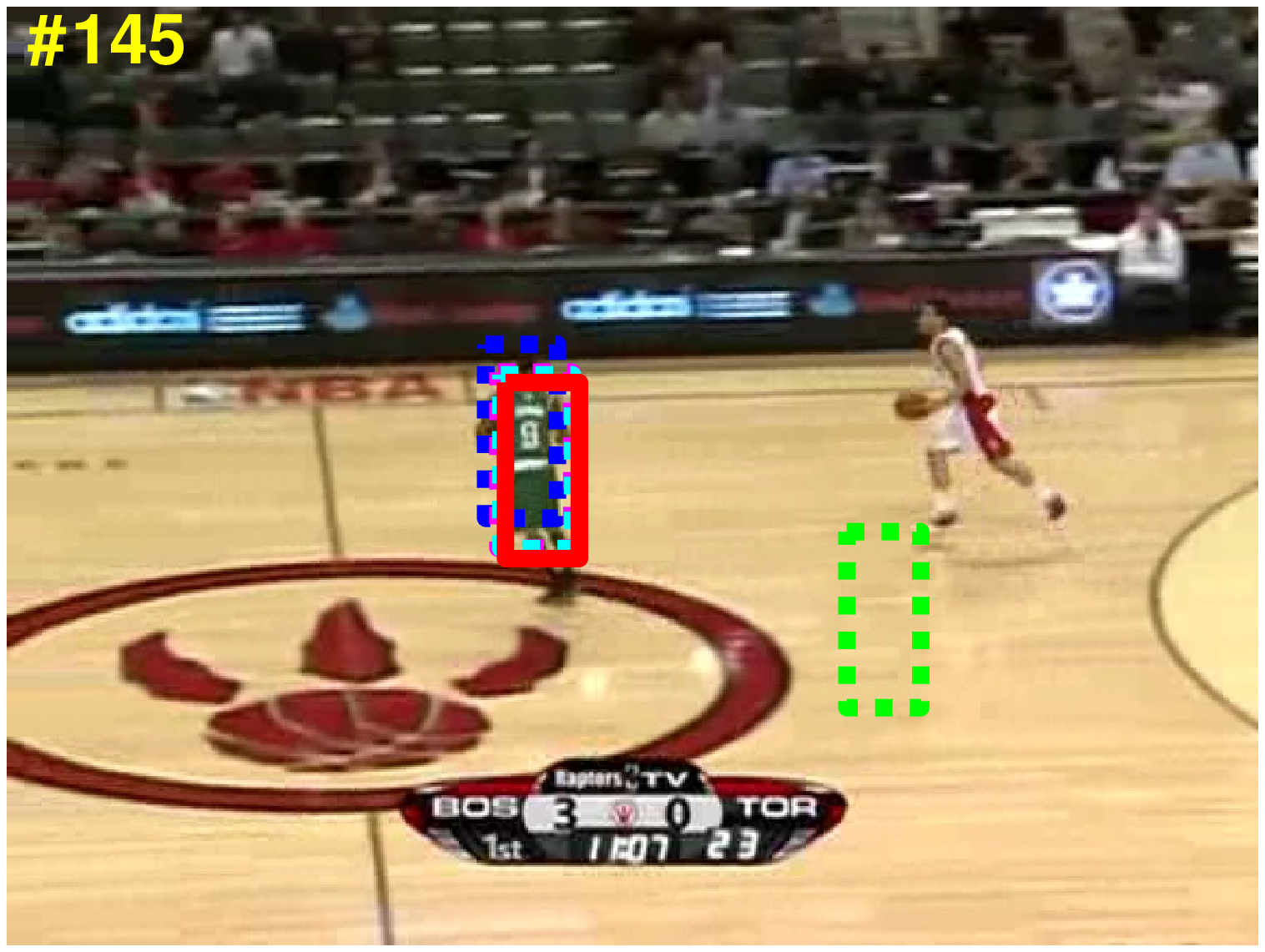,width=0.16\textwidth}
\epsfig{file=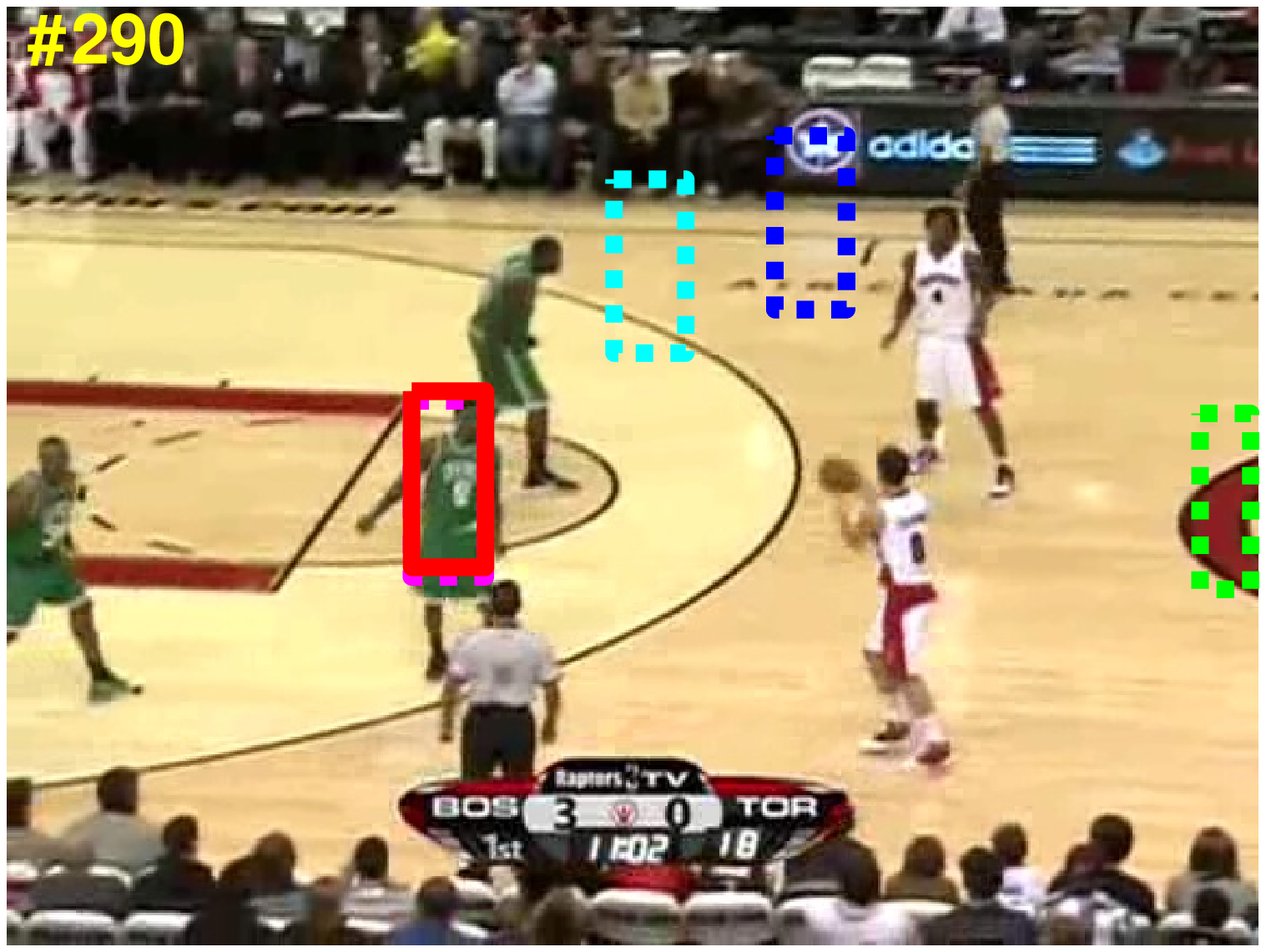,width=0.16\textwidth}
\epsfig{file=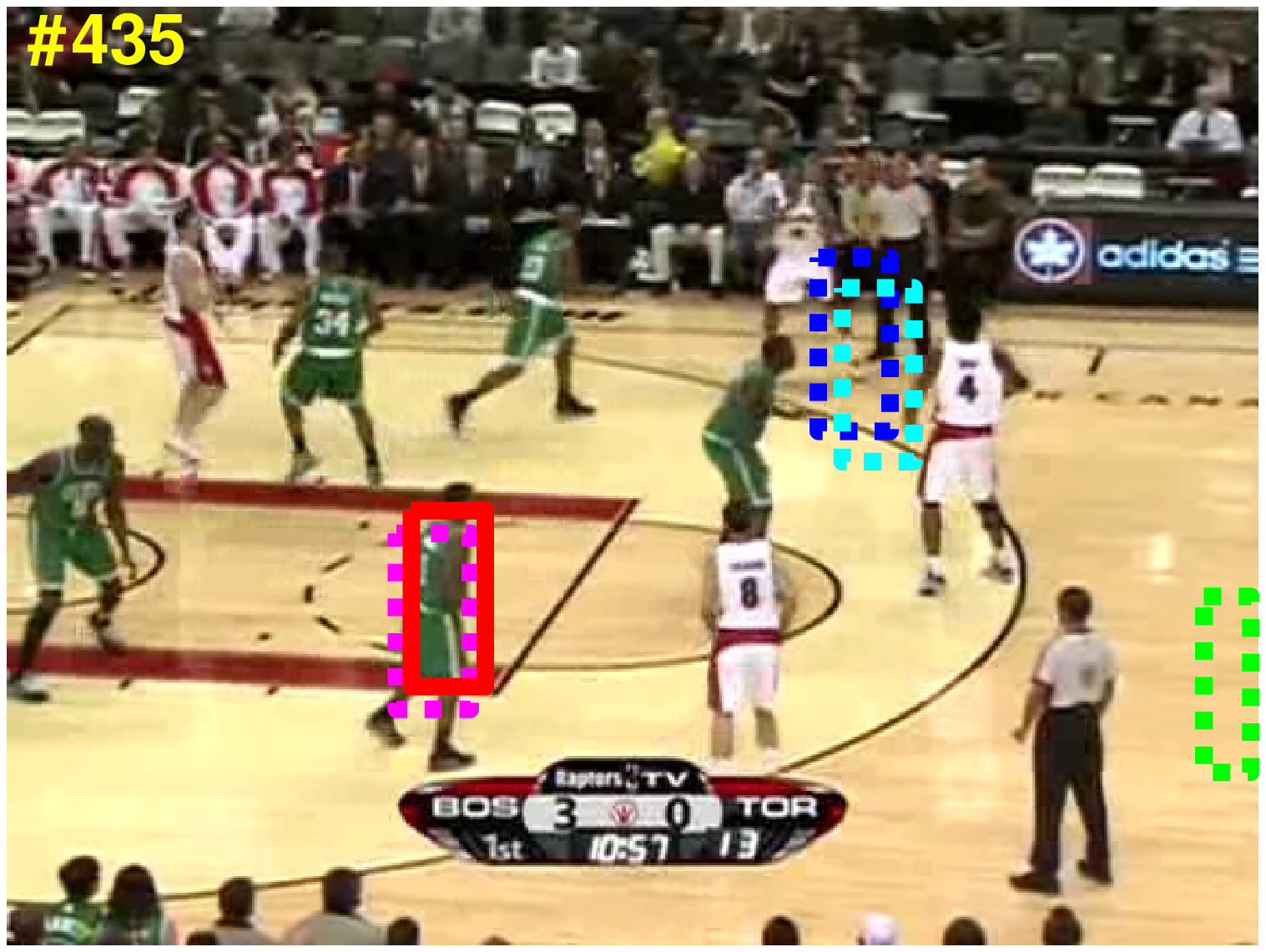,width=0.16\textwidth}
\epsfig{file=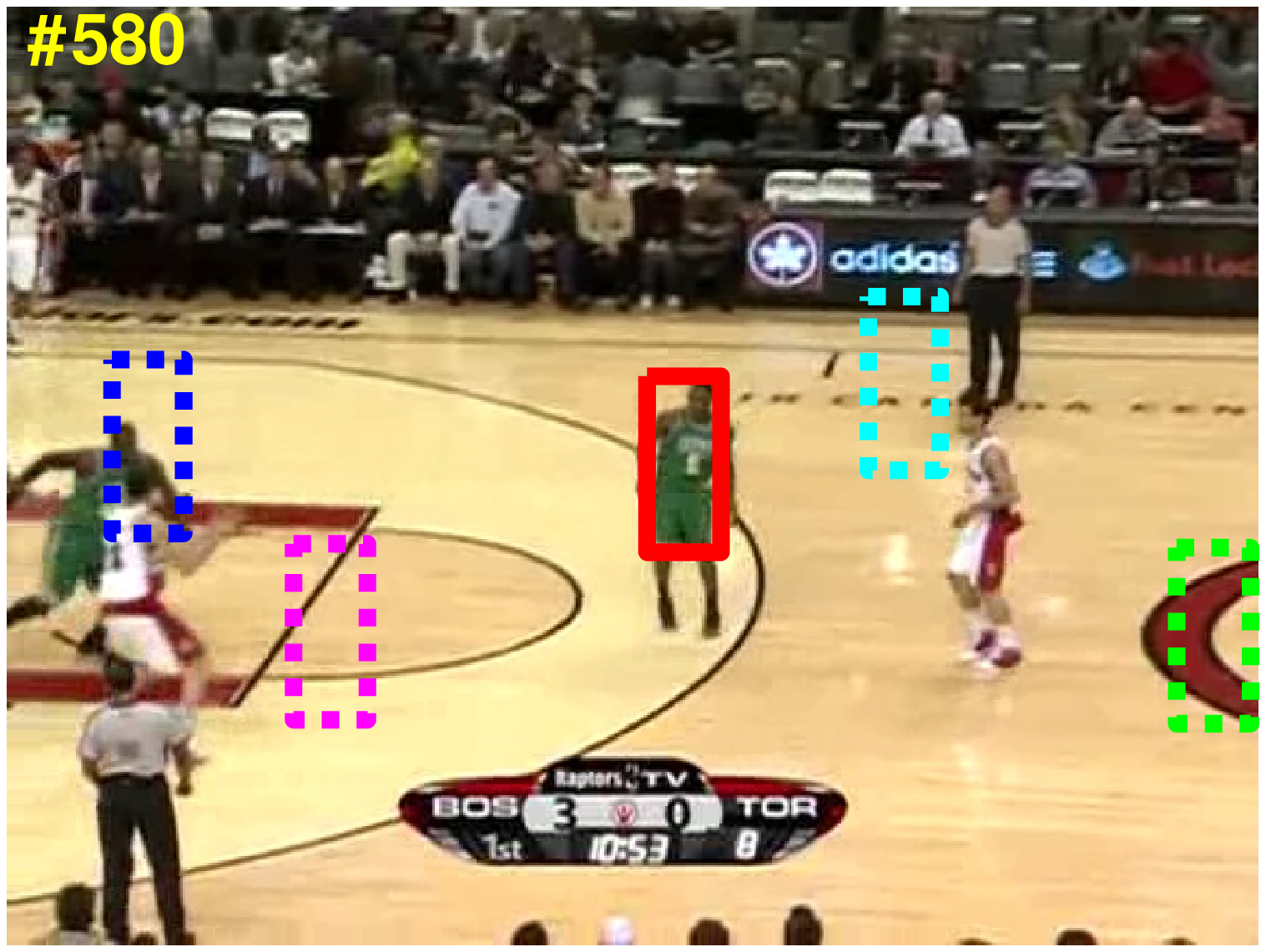,width=0.16\textwidth}
\epsfig{file=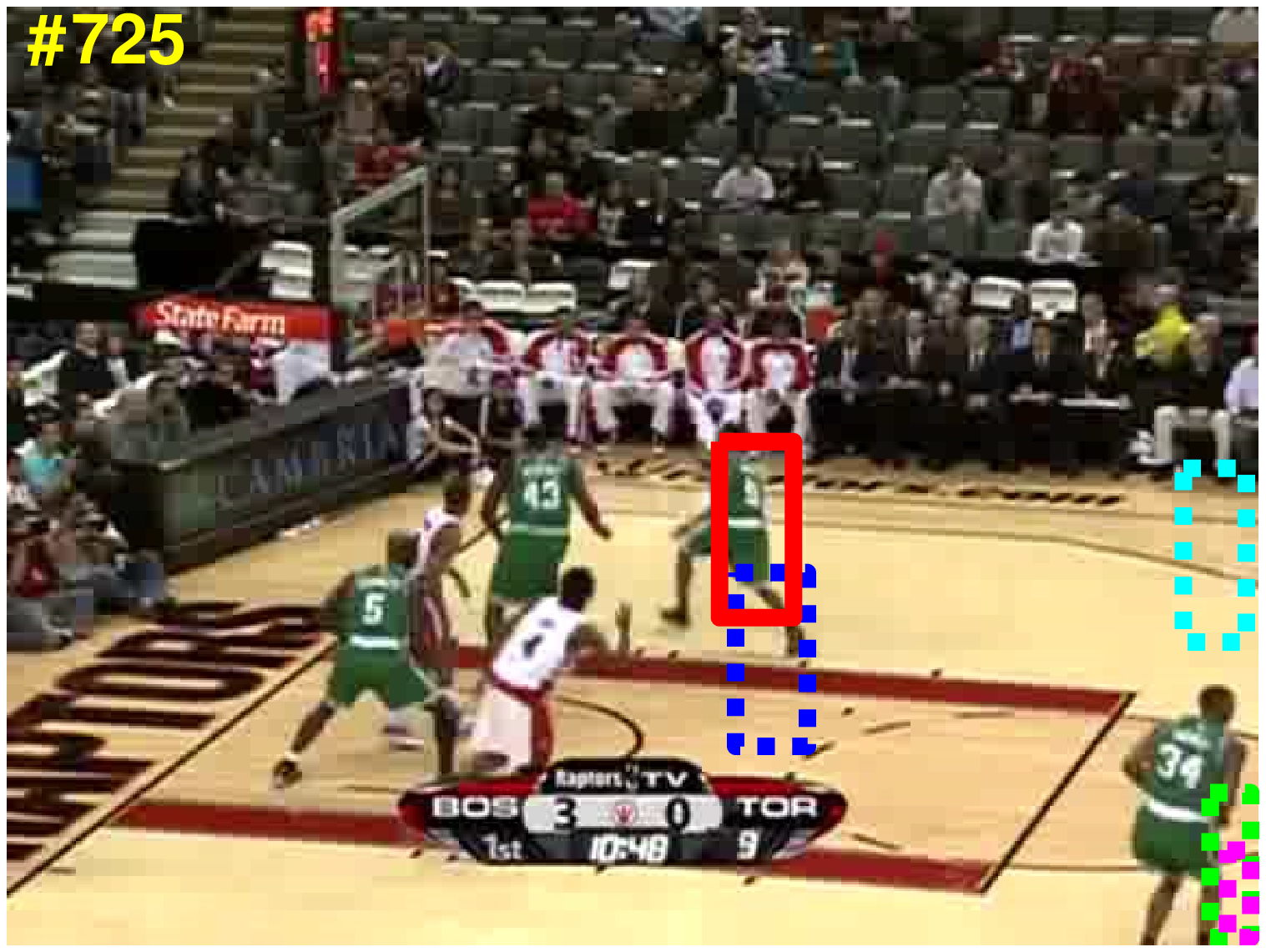,width=0.16\textwidth}}
\\ \vspace{-0.1in}

\centering \subfloat[Biker]{
\epsfig{file=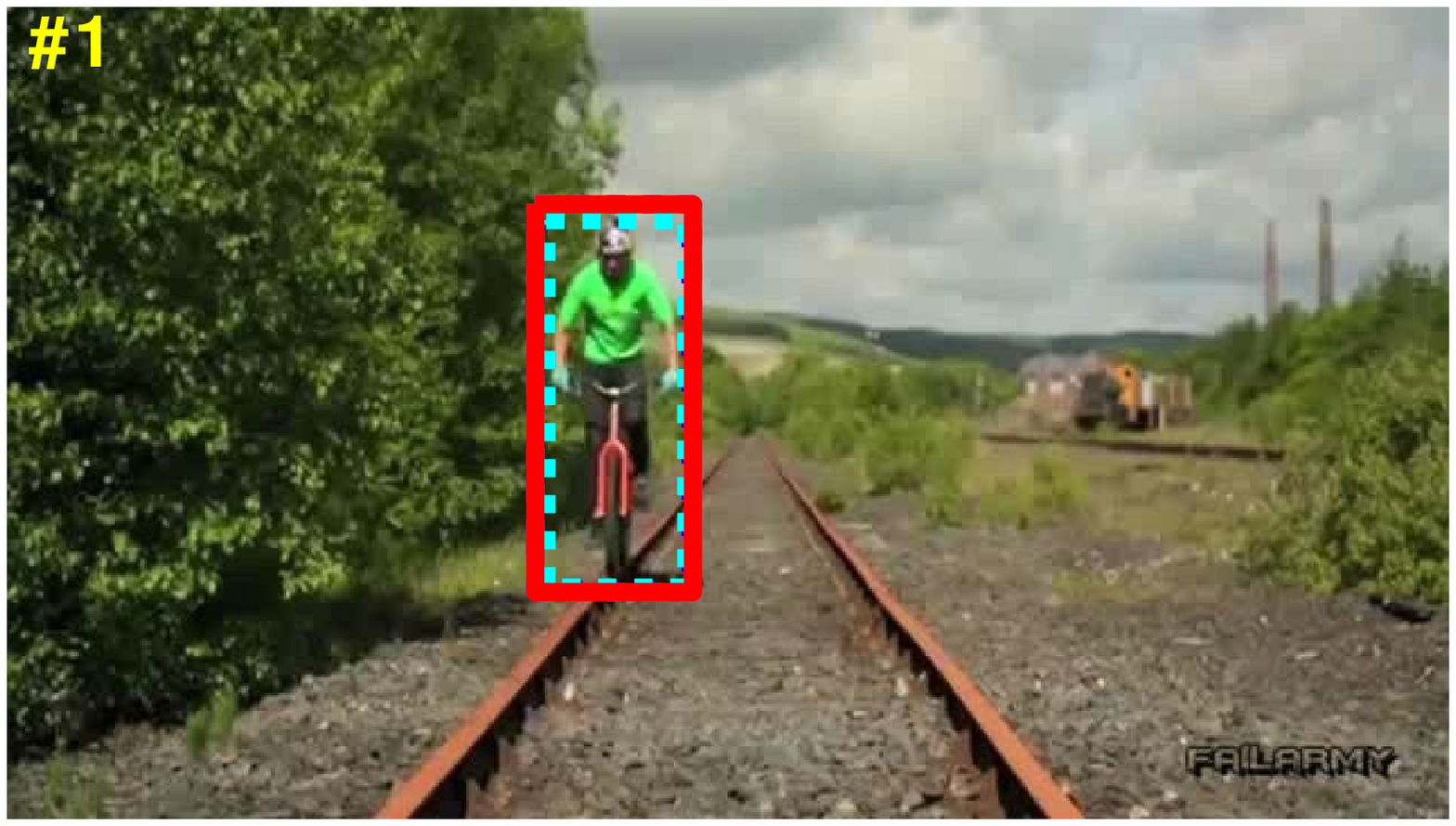,width=0.16\textwidth}
\epsfig{file=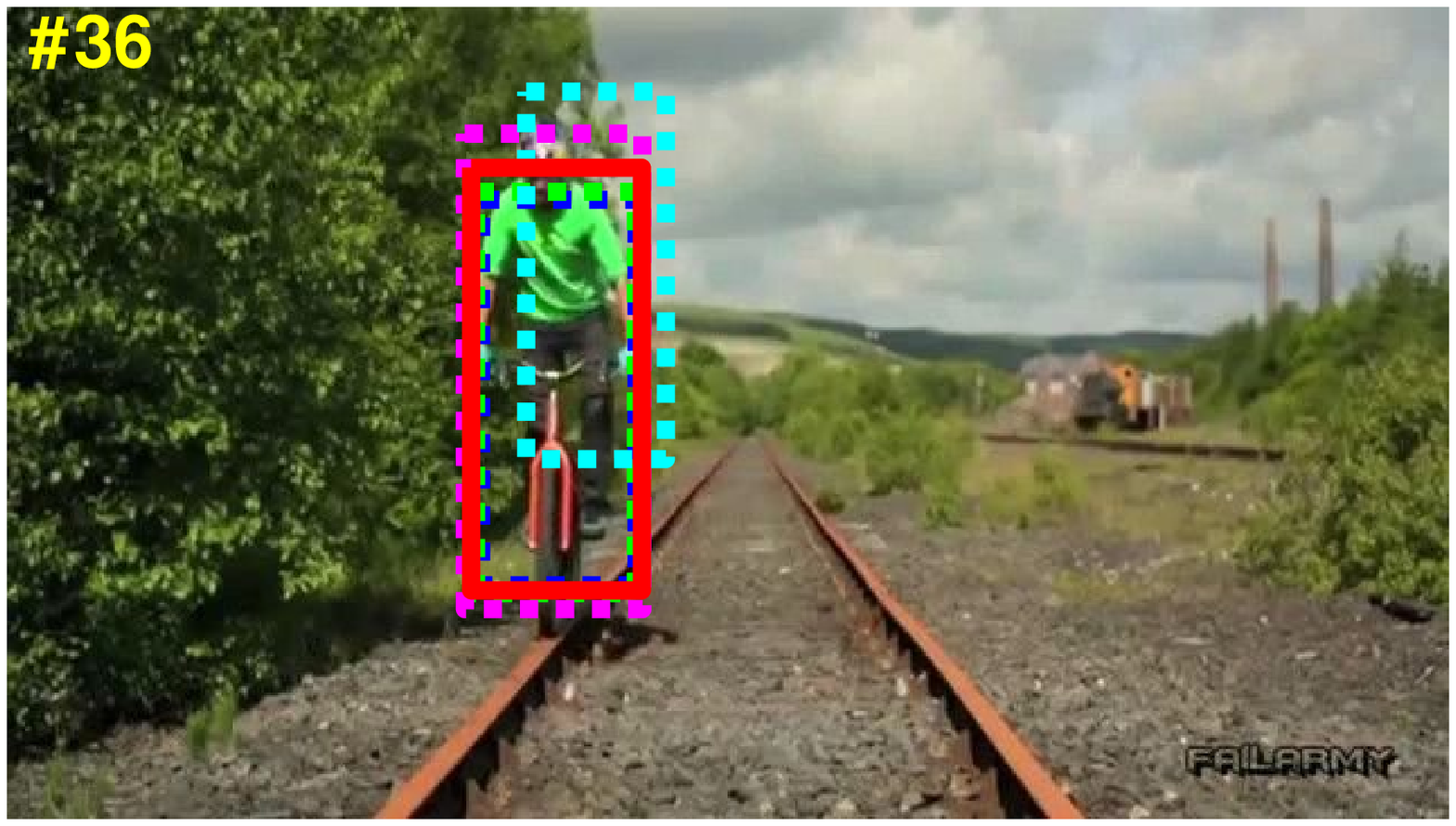,width=0.16\textwidth}
\epsfig{file=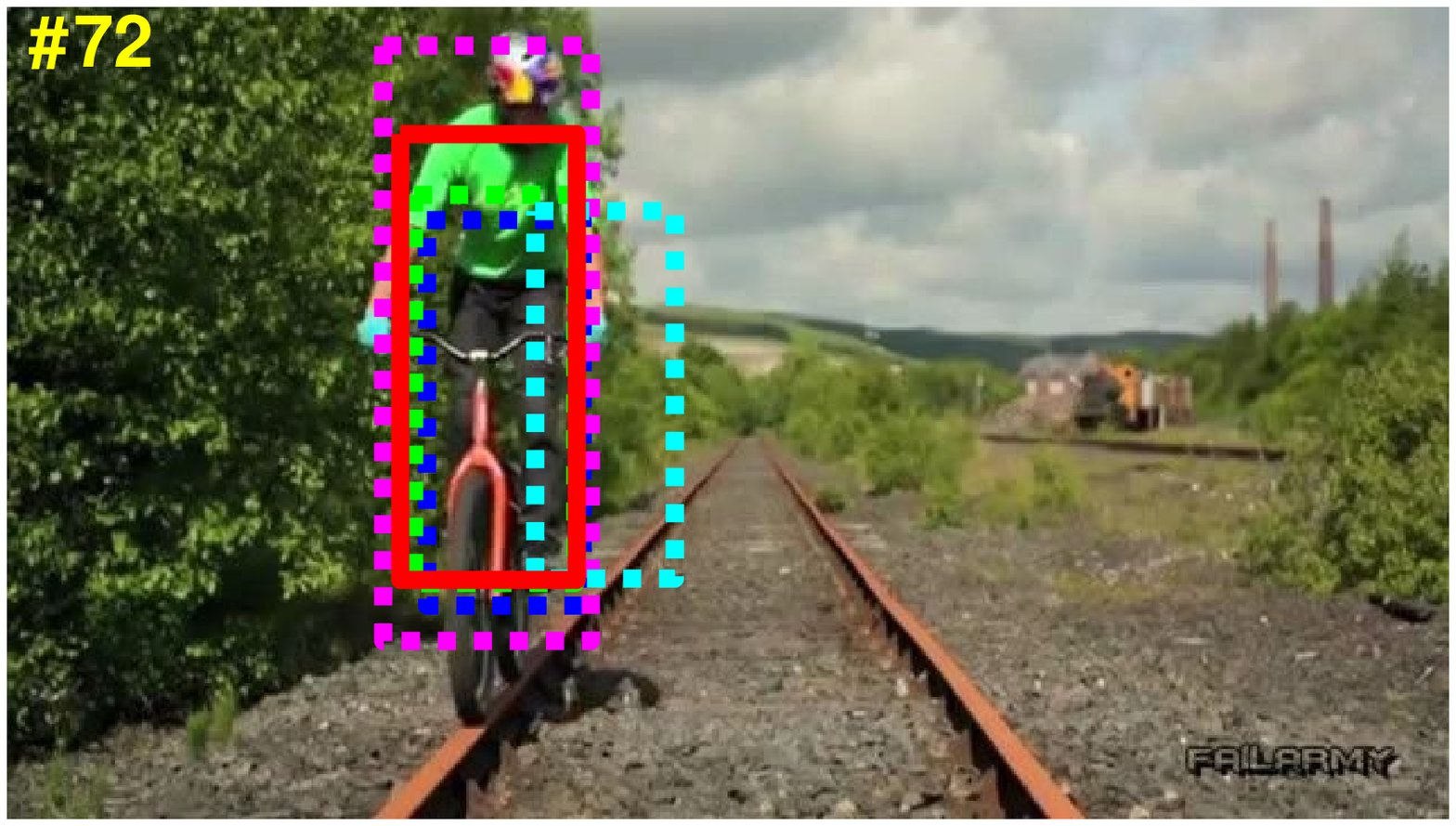,width=0.16\textwidth}
\epsfig{file=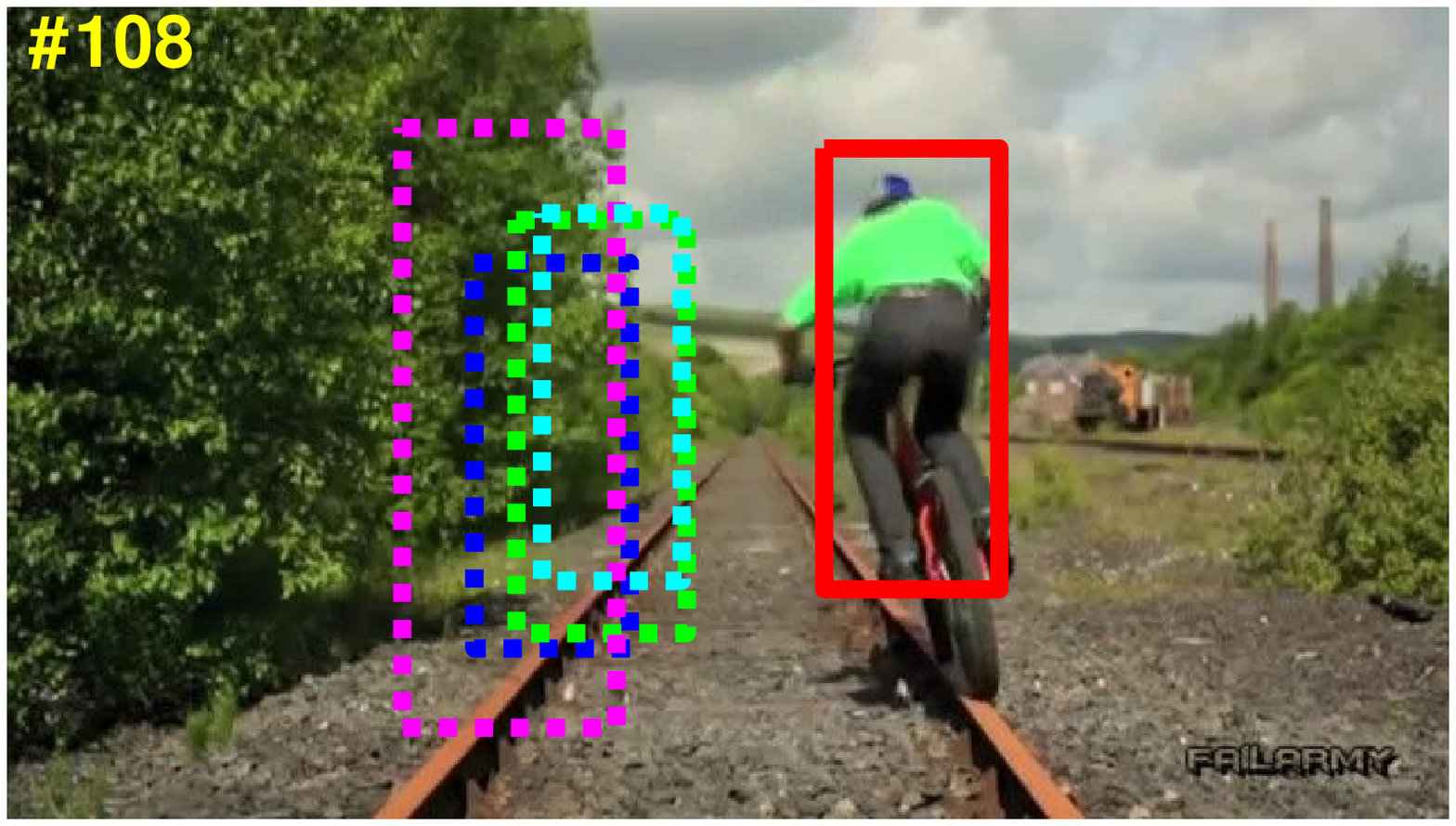,width=0.16\textwidth}
\epsfig{file=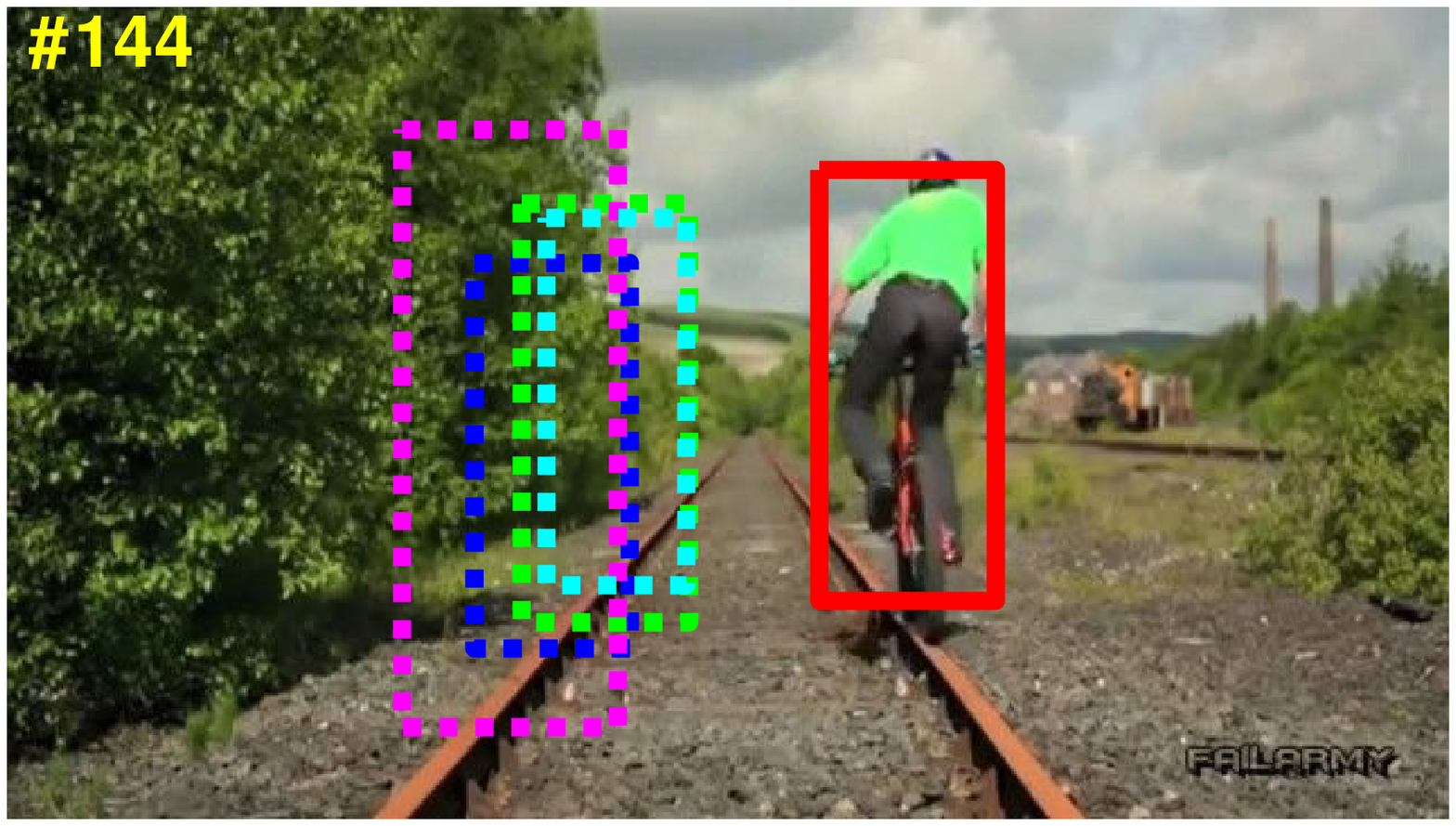,width=0.16\textwidth}
\epsfig{file=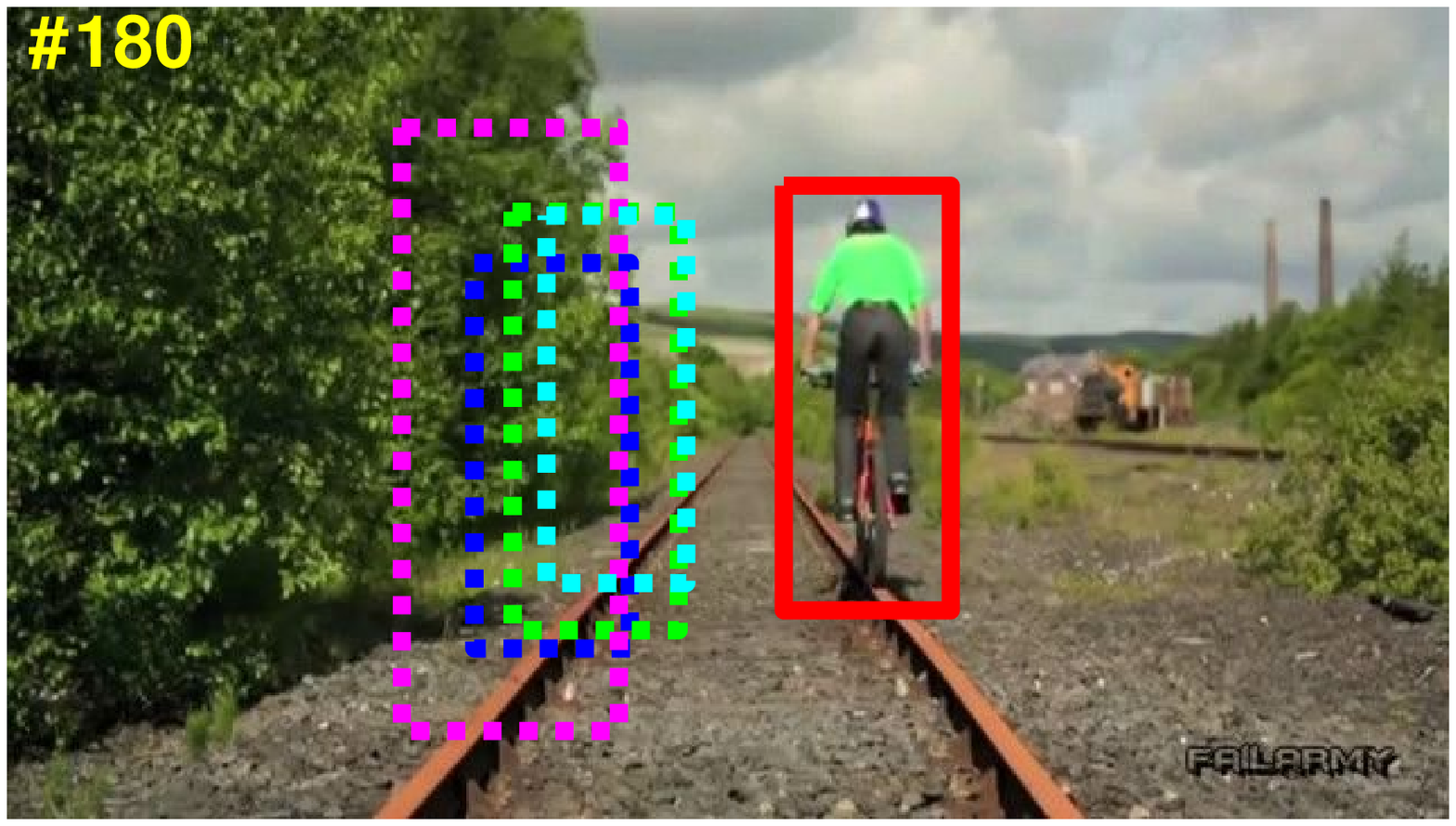,width=0.16\textwidth}}
\\ \vspace{-0.1in}

\centering \subfloat[FleetFace]{
\epsfig{file=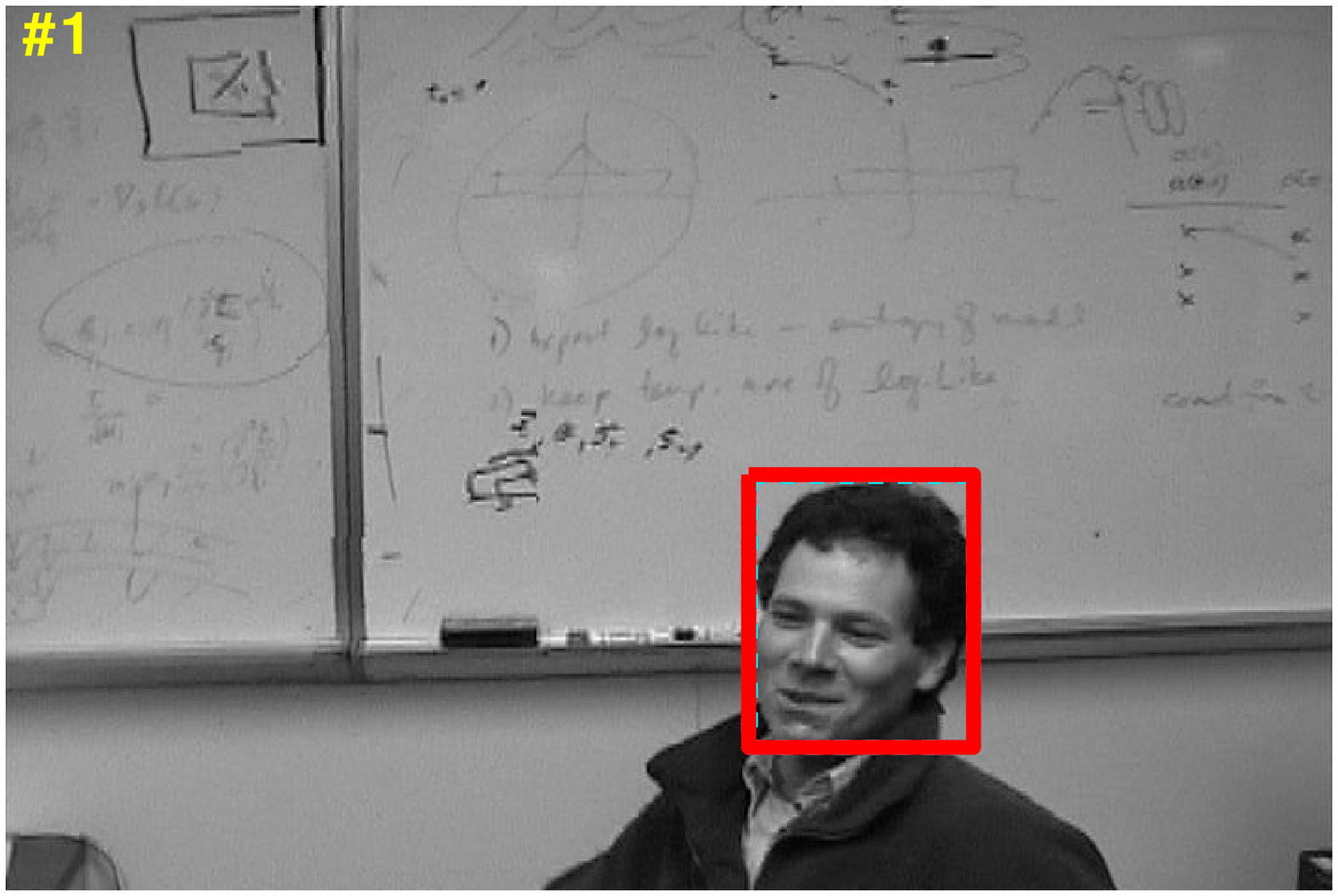,width=0.16\textwidth}
\epsfig{file=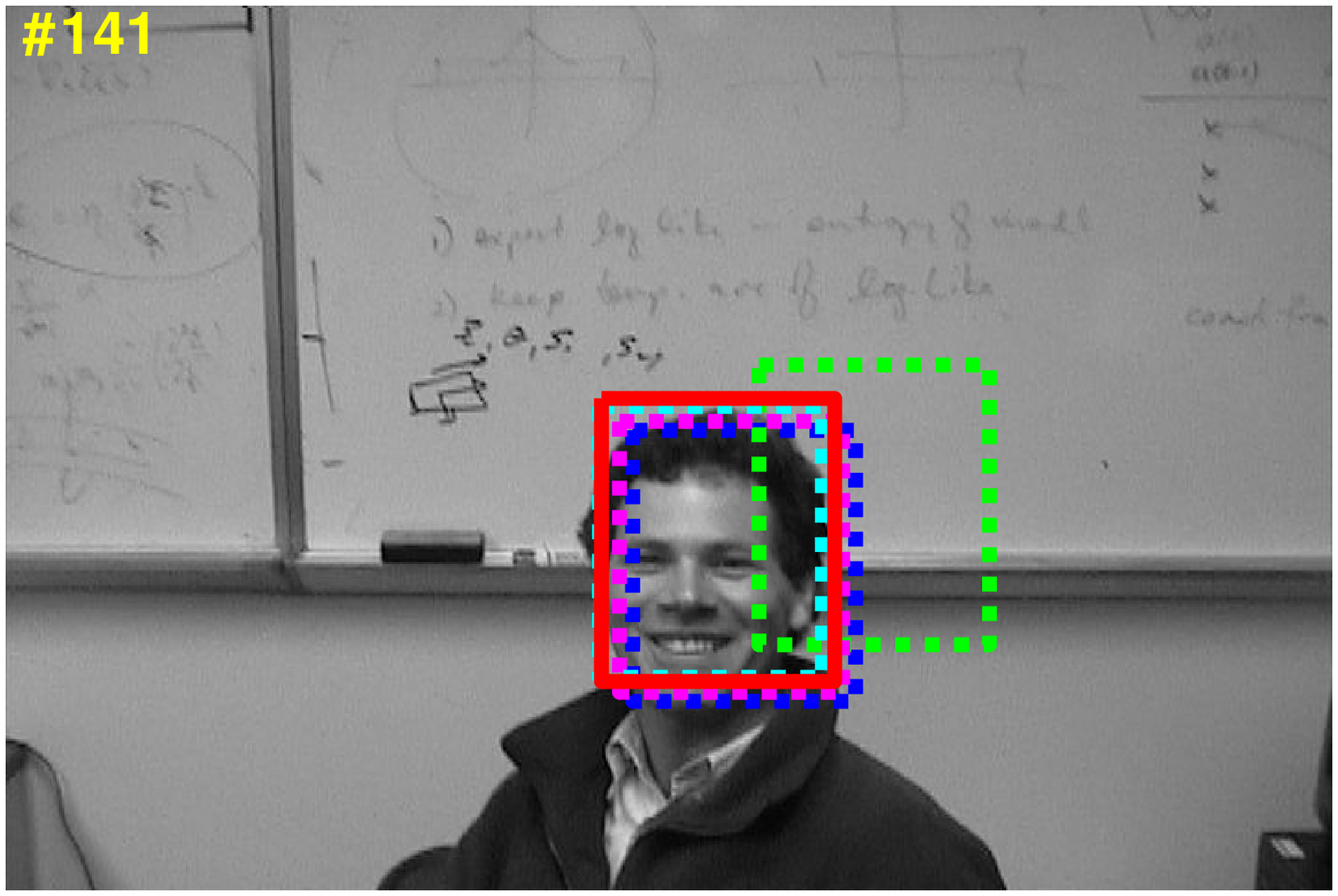,width=0.16\textwidth}
\epsfig{file=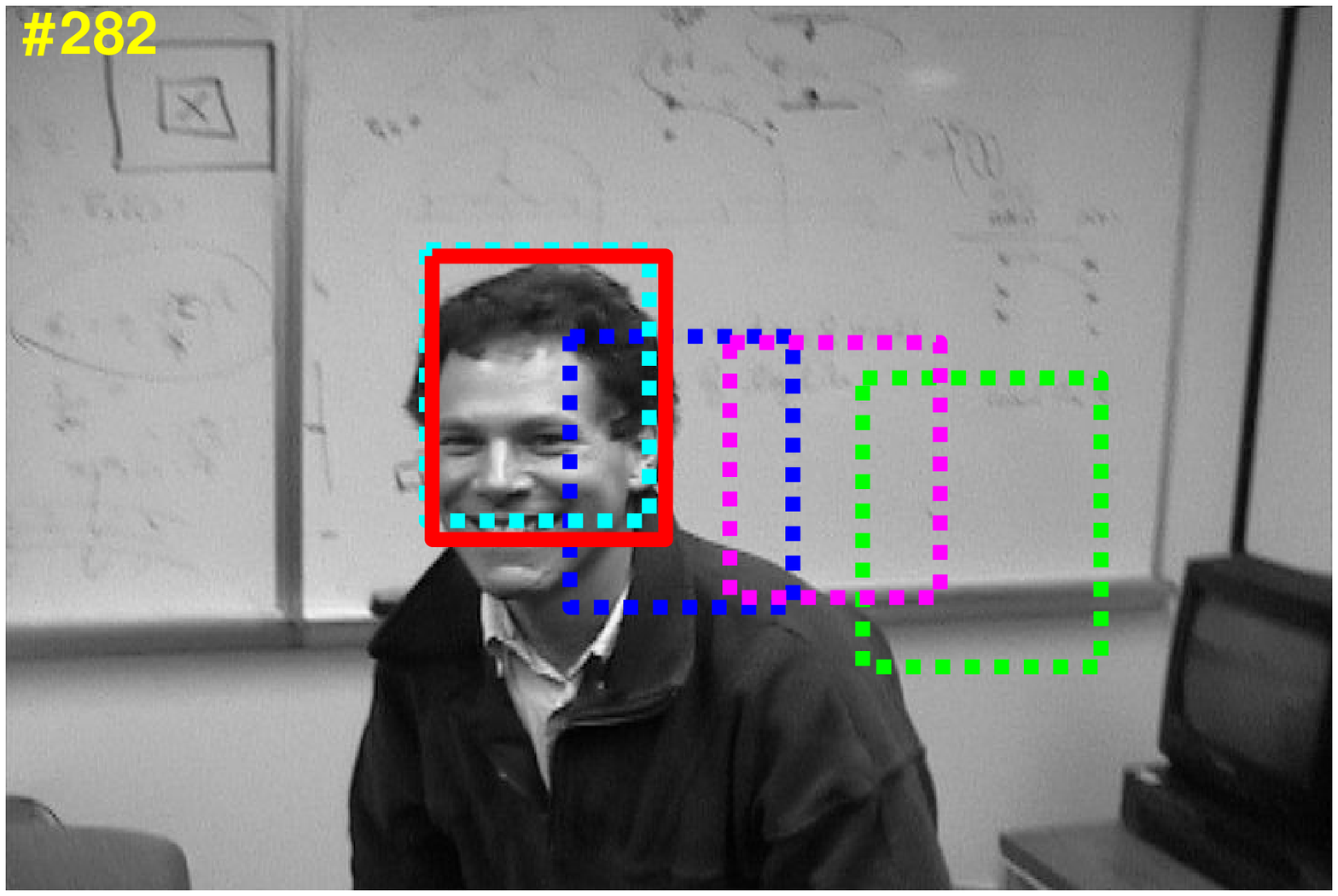,width=0.16\textwidth}
\epsfig{file=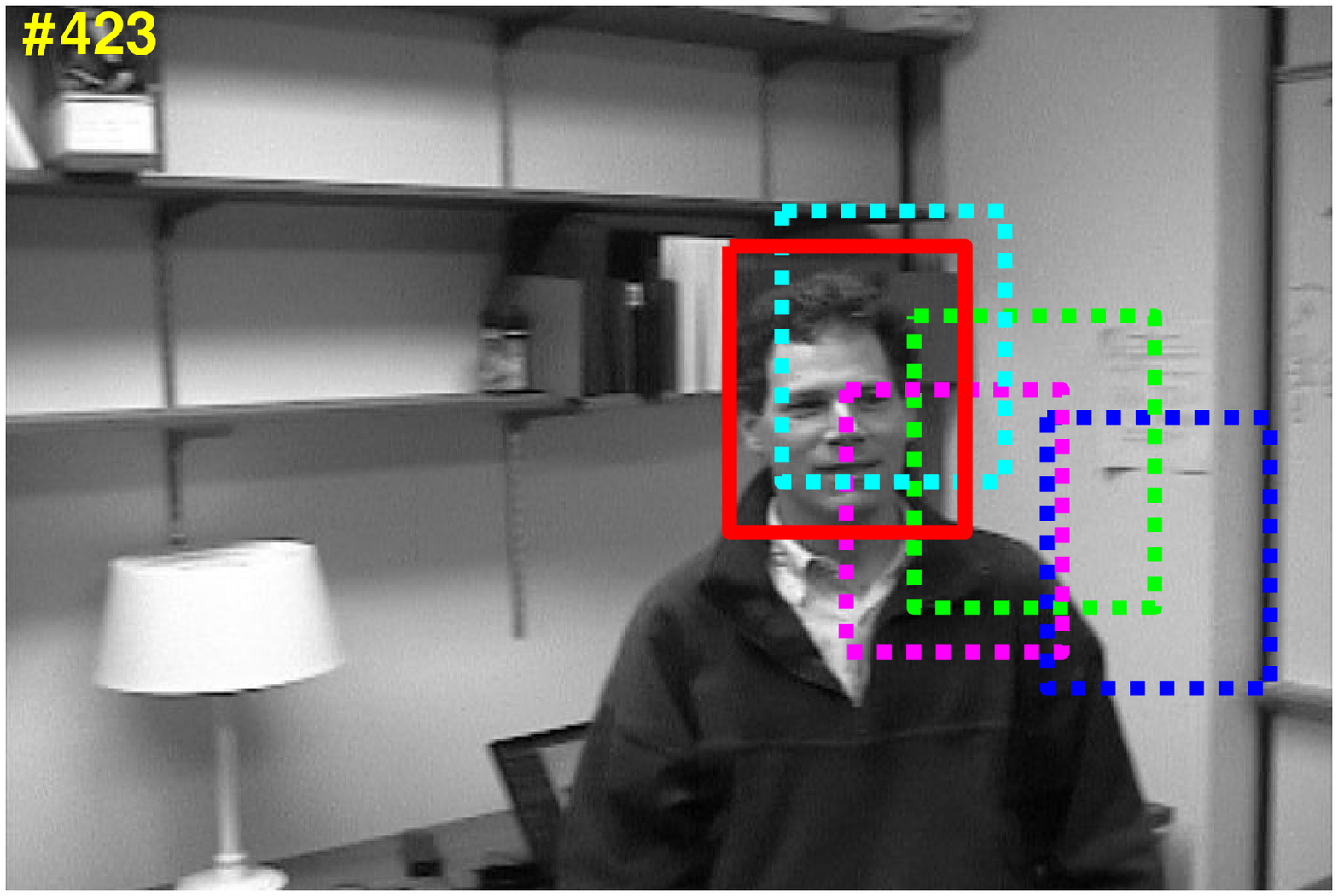,width=0.16\textwidth}
\epsfig{file=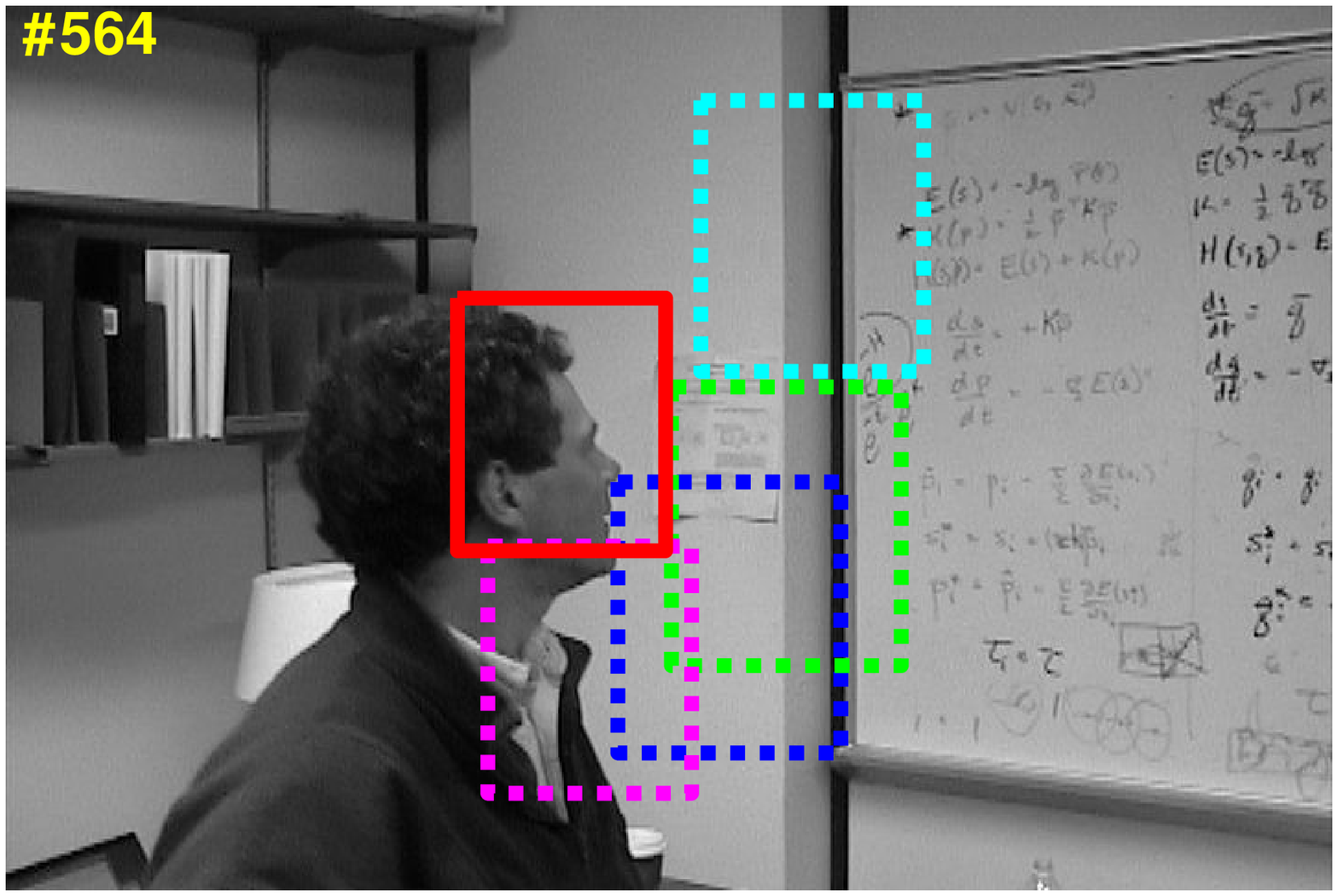,width=0.16\textwidth}
\epsfig{file=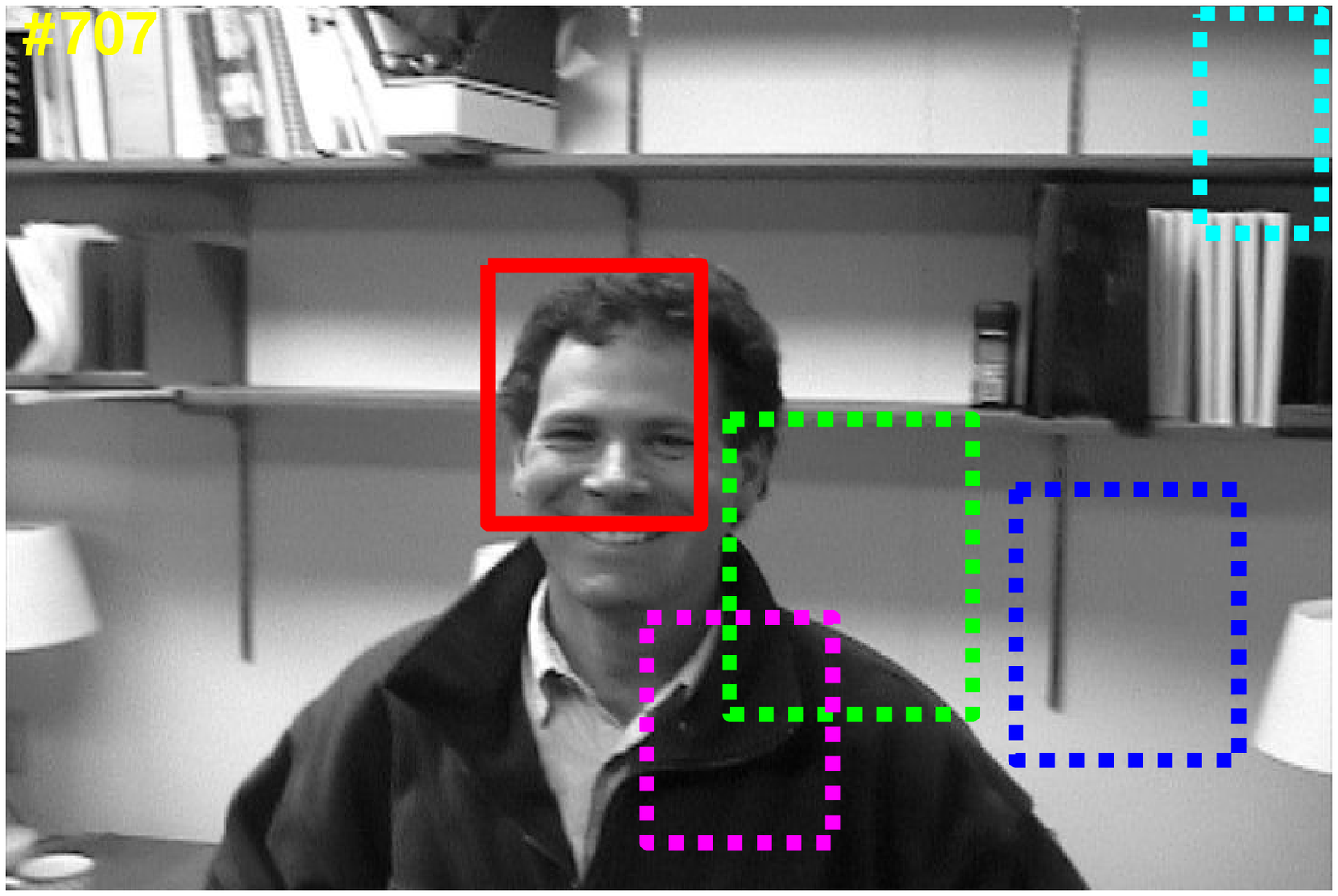,width=0.16\textwidth}}
\\ \vspace{-0.1in}

\centering \subfloat[Kitesurf]{
\epsfig{file=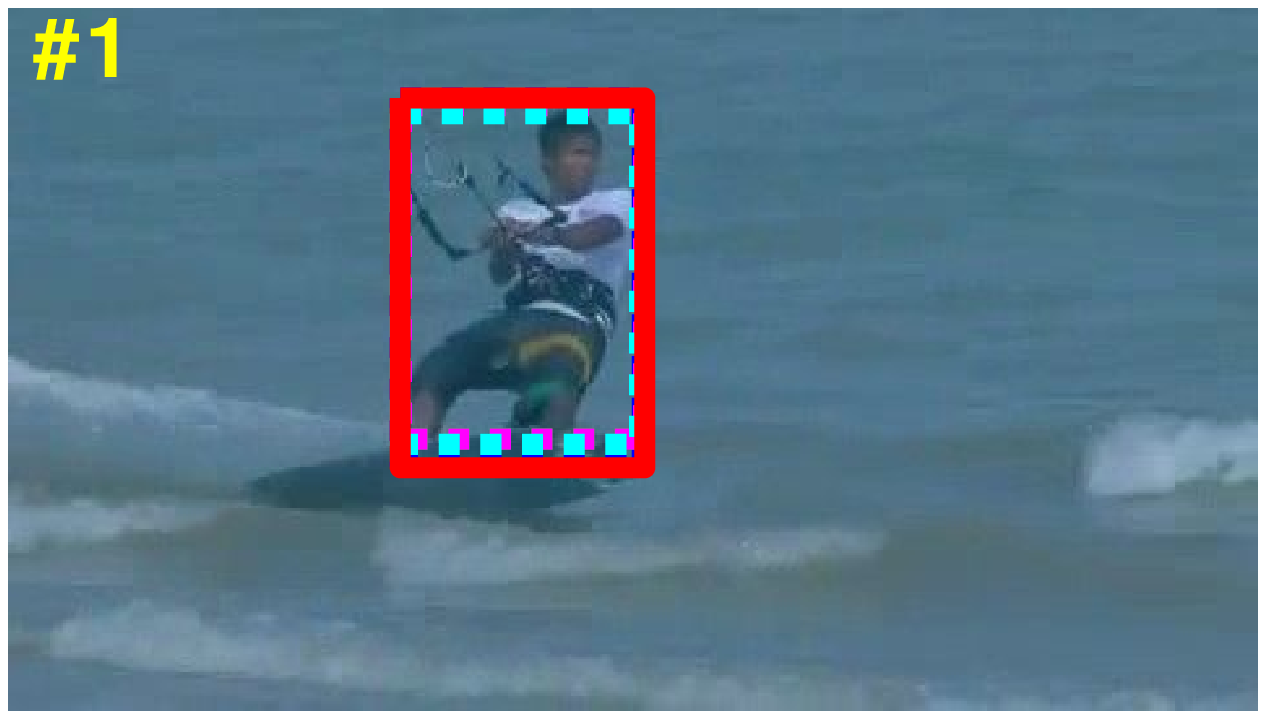,width=0.16\textwidth}
\epsfig{file=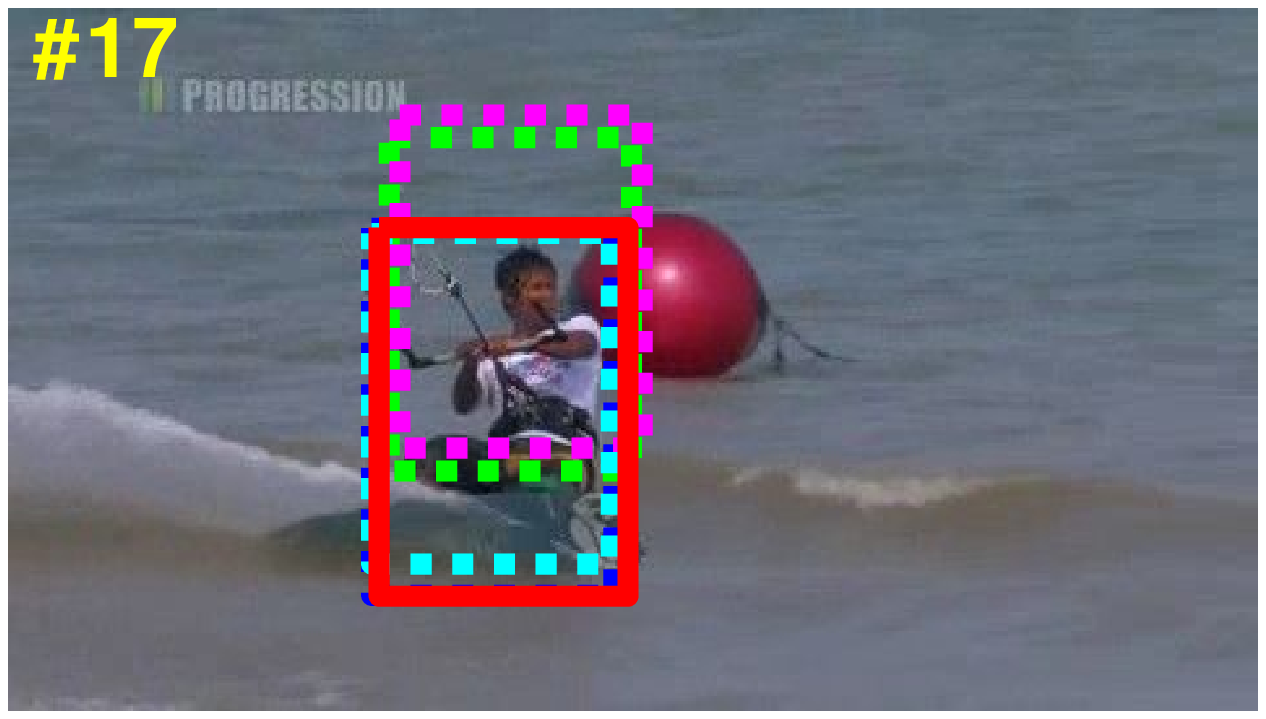,width=0.16\textwidth}
\epsfig{file=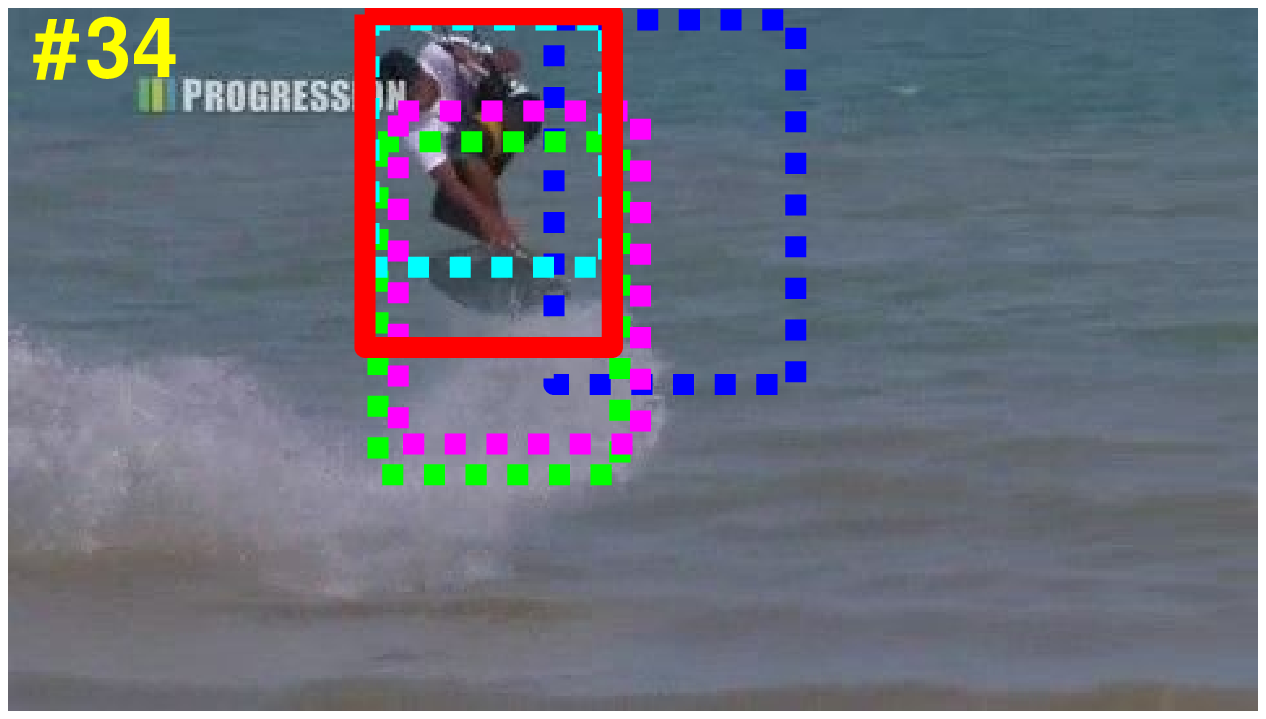,width=0.16\textwidth}
\epsfig{file=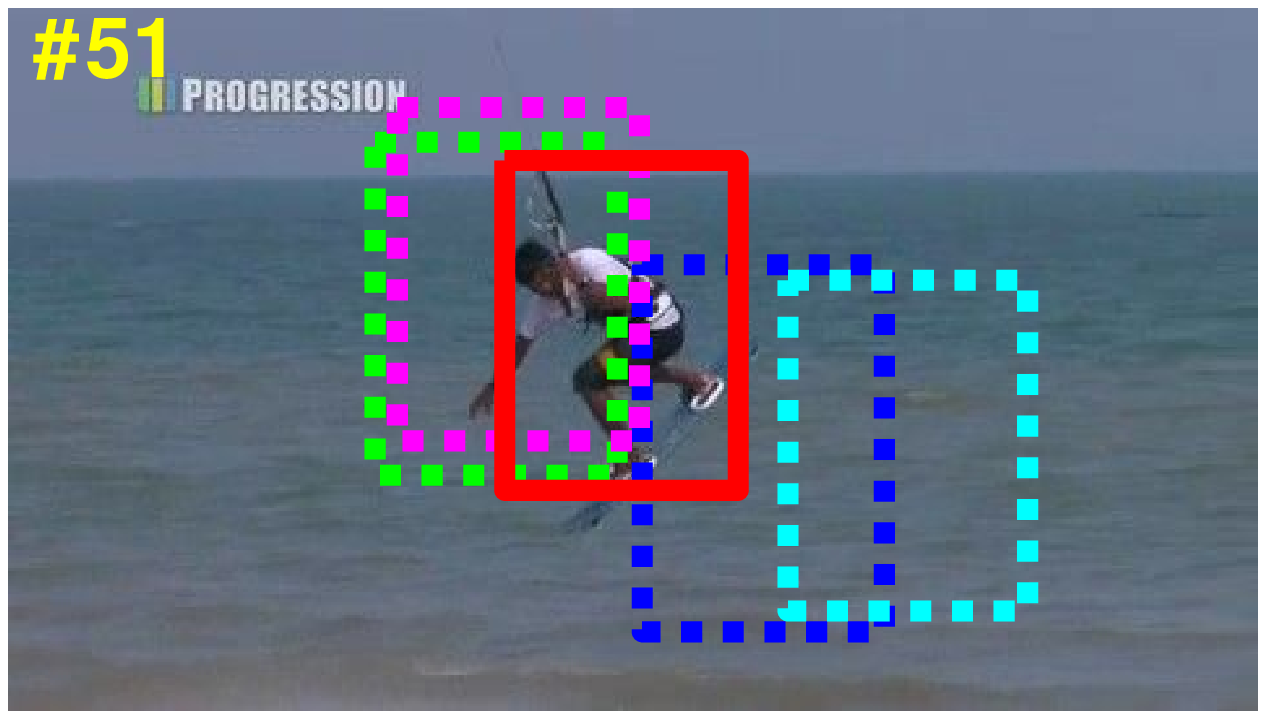,width=0.16\textwidth}
\epsfig{file=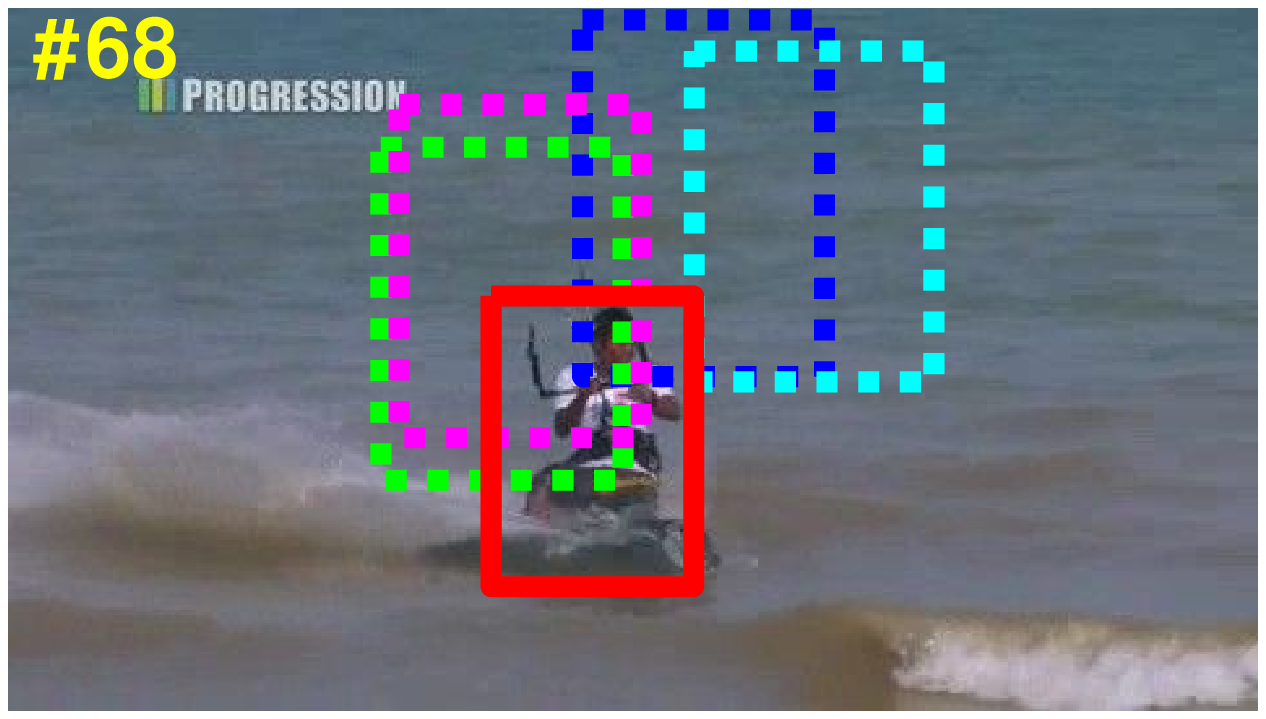,width=0.16\textwidth}
\epsfig{file=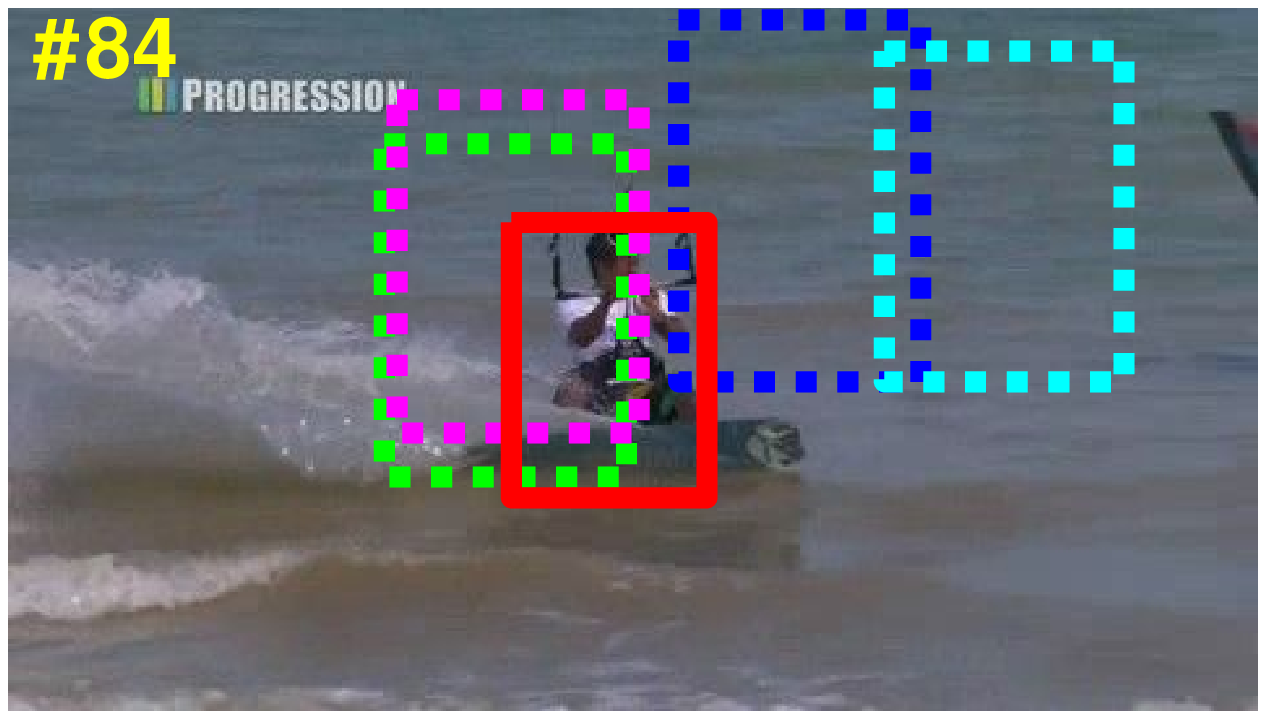,width=0.16\textwidth}}
\\ \vspace{-0.1in}

\centering \subfloat[Skating1]{
\epsfig{file=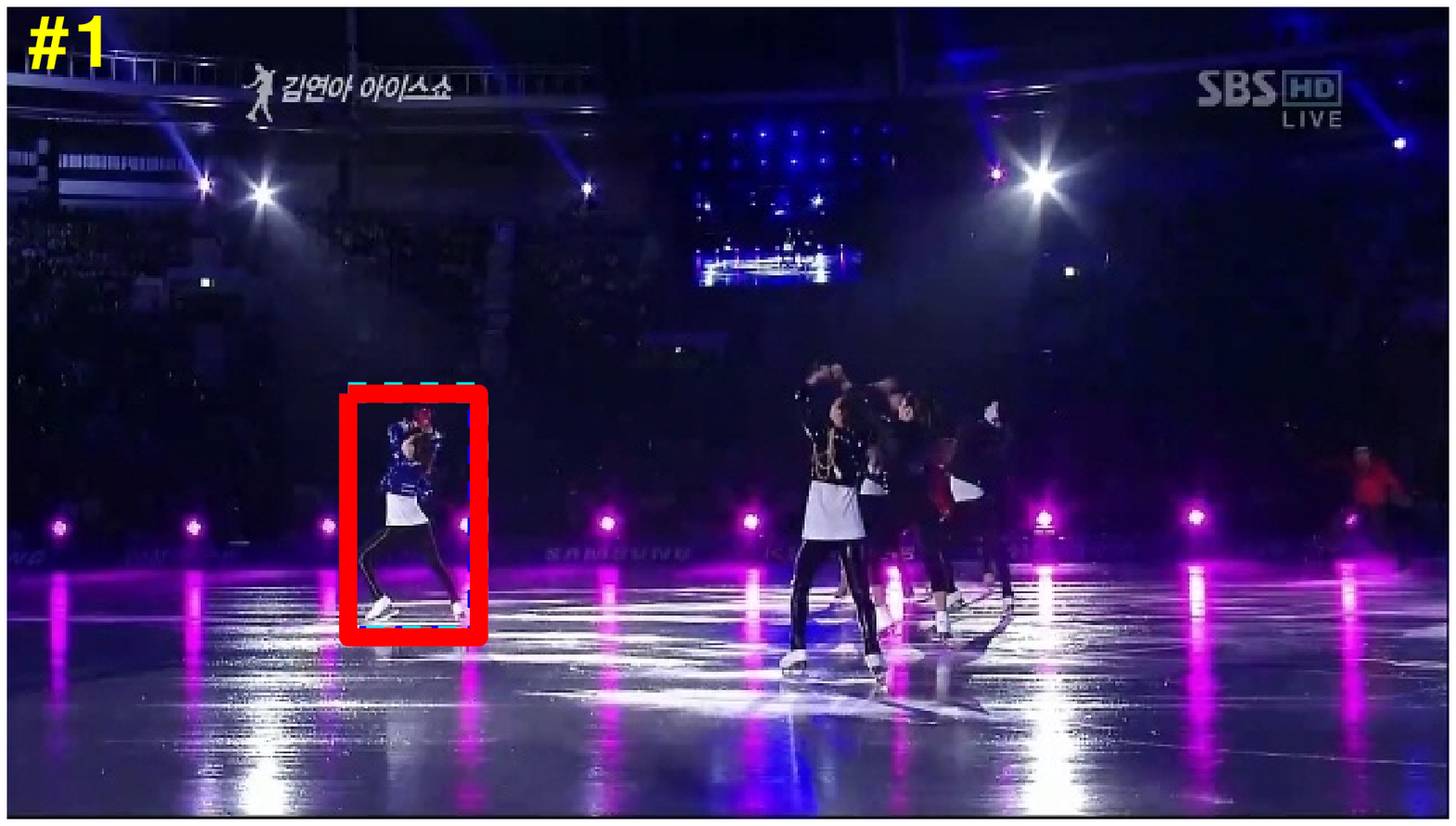,width=0.16\textwidth}
\epsfig{file=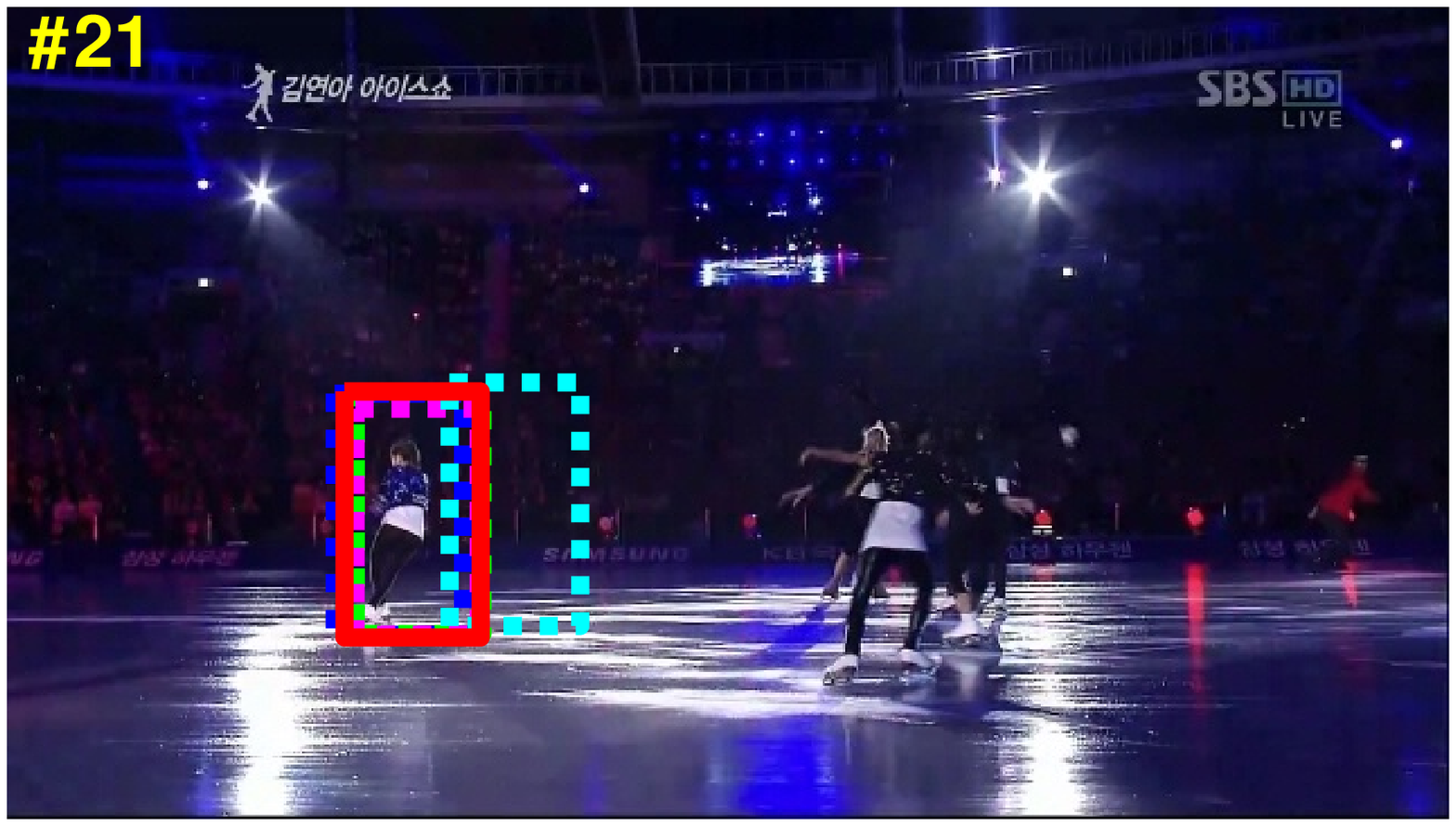,width=0.16\textwidth}
\epsfig{file=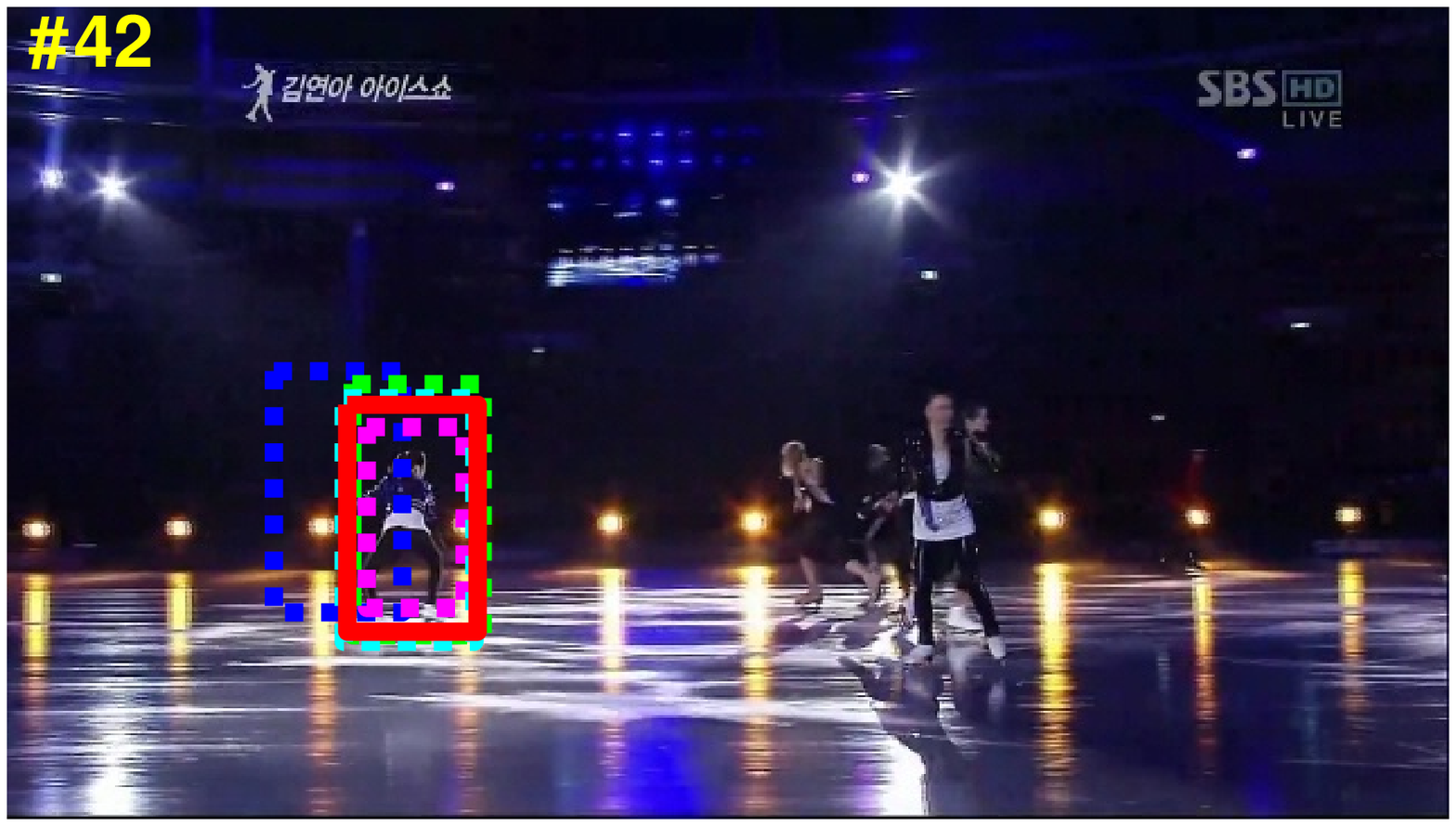,width=0.16\textwidth}
\epsfig{file=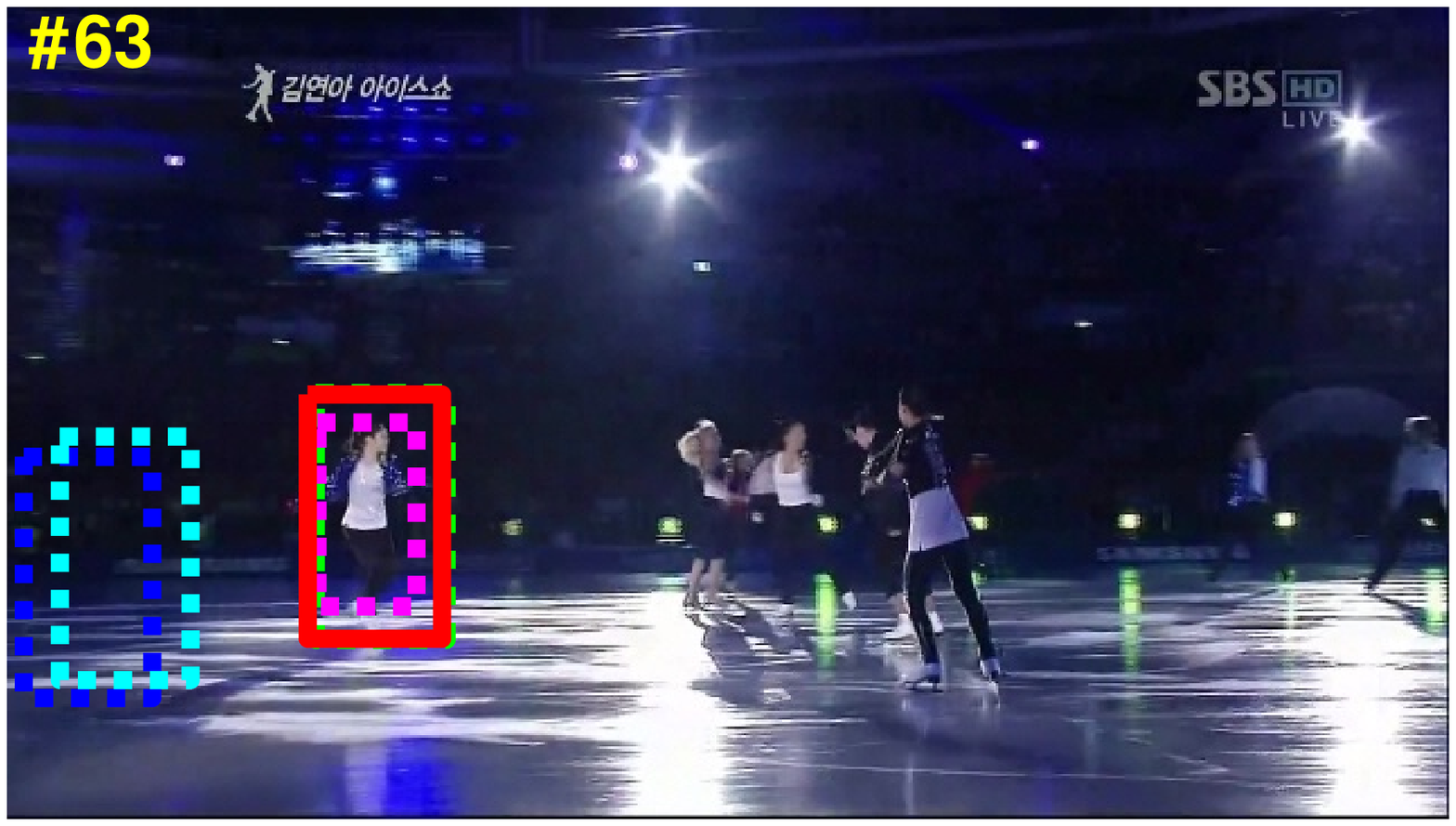,width=0.16\textwidth}
\epsfig{file=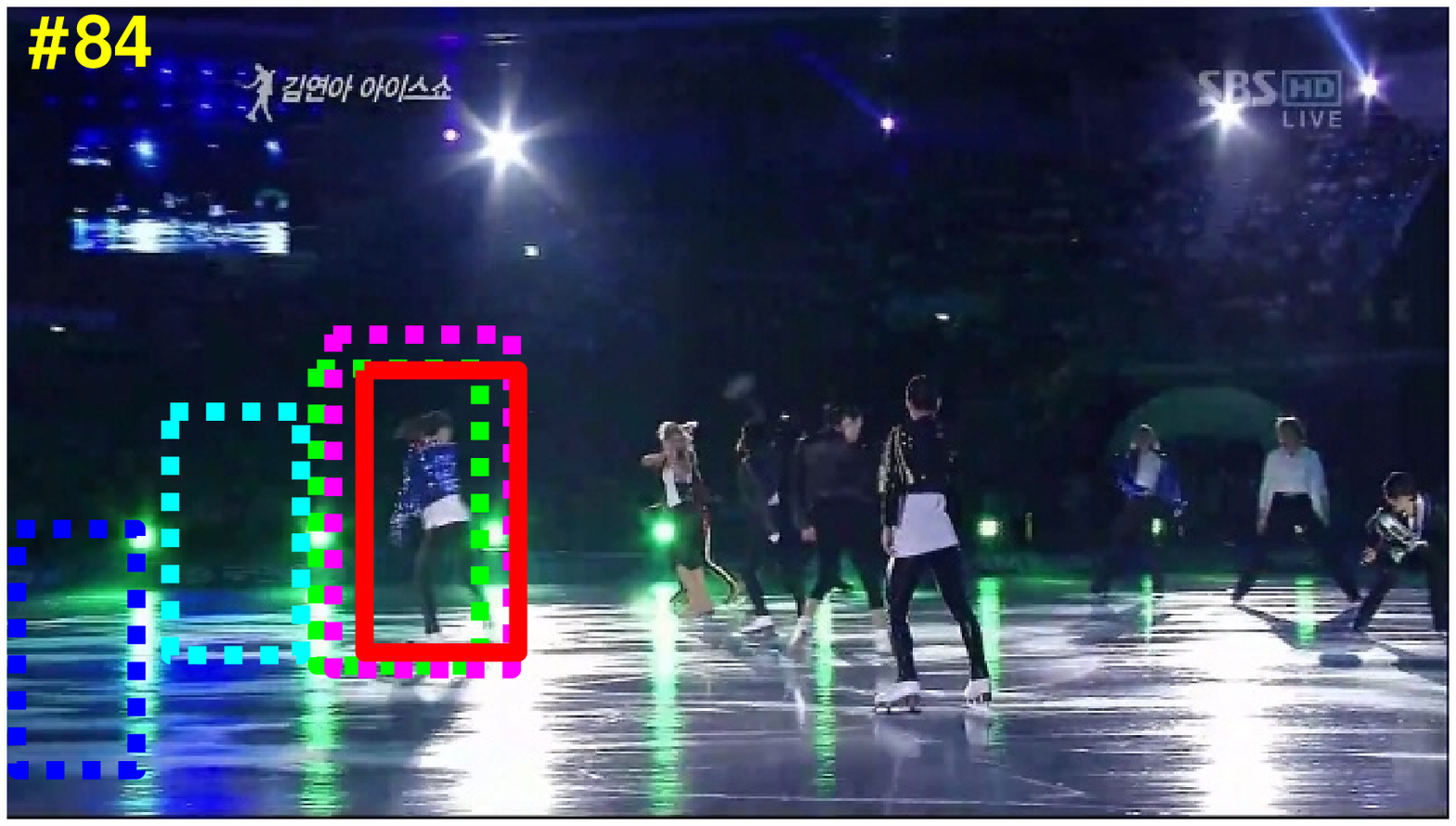,width=0.16\textwidth}
\epsfig{file=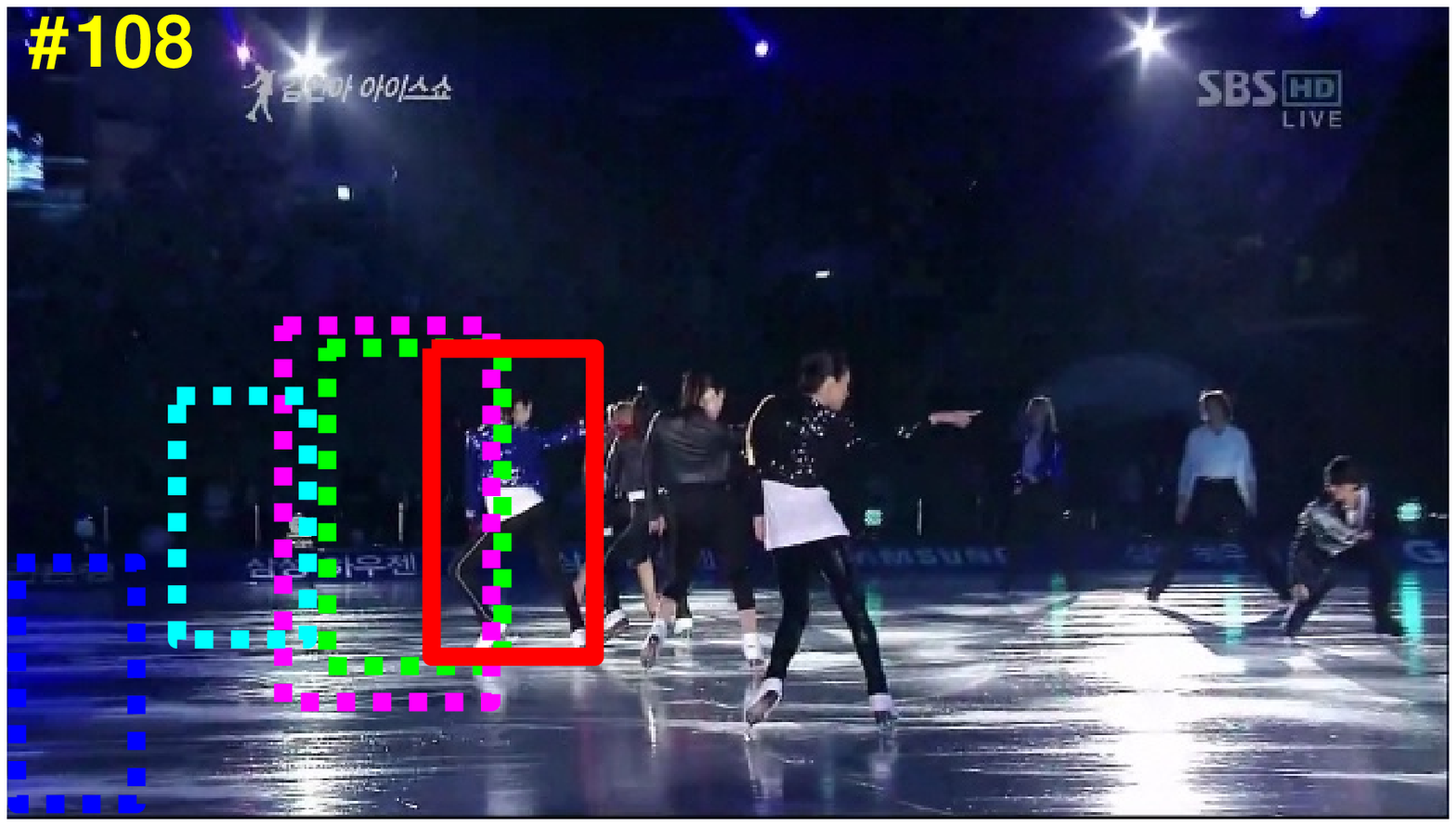,width=0.16\textwidth}}

\caption{Qualitative results on sequences with non-rigid object deformation. The purple, green, cyan, blue and red bounding boxes refer to ASLA\cite{DBLP:conf/cvpr/JiaLY12}\_RAW, ASLA\cite{DBLP:conf/cvpr/JiaLY12}\_HOG, $\ell_1$\_APG \cite{DBLP:conf/cvpr/BaoWLJ12}, CT\_DIF \cite{DBLP:conf/eccv/Zhang0Y12} and our tracker respectively. This figure is better viewed in color.} \label{fig:quali_deformation}
\end{figure*}

We present both quantitative and qualitative results on $15$ challenging sequences, in which target objects have complicated motion transformations. \eg in-plane rotation, out-of-plane rotation and non-rigid object deformation. To demonstrate our learned feature's robustness to complicated motion transformations, we compare our tracker with the other $4$ state-of-the-art trackers using different feature representations such as the raw pixel value (ASLA\cite{DBLP:conf/cvpr/JiaLY12}\_RAW), the hand-crafted feature of Histogram of Oriented Gradients (HOG) \cite{DBLP:conf/cvpr/DalalT05} (ASLA\cite{DBLP:conf/cvpr/JiaLY12}\_HOG), the sparse representation ($\ell_1$\_APG \cite{DBLP:conf/cvpr/BaoWLJ12}) and the data-independent feature (CT\_DIF \cite{DBLP:conf/eccv/Zhang0Y12}). It is necessary to mention that ASLA\_HOG and our tracker use the same tracking framework as in ASLA\_RAW \cite{DBLP:conf/cvpr/JiaLY12}. The difference is that ASLA\_HOG and our tracker integrate the HOG feature and our learned hierarchical features into the baseline ASLA tracker respectively. However, the other $2$ trackers, $\ell_1$\_APG and CT\_DIF, use their own tracking frameworks which are different from ASLA\_RAW. The hand-crafted HOG feature and the sparse feature are employed here because of their superior performances in object detection and recognition. Additionally, the data-independent feature is used here because it also aims to solve the problem of insufficient training data in object tracking.

\begin{figure*}
\centering \subfloat[David2]{
\epsfig{file=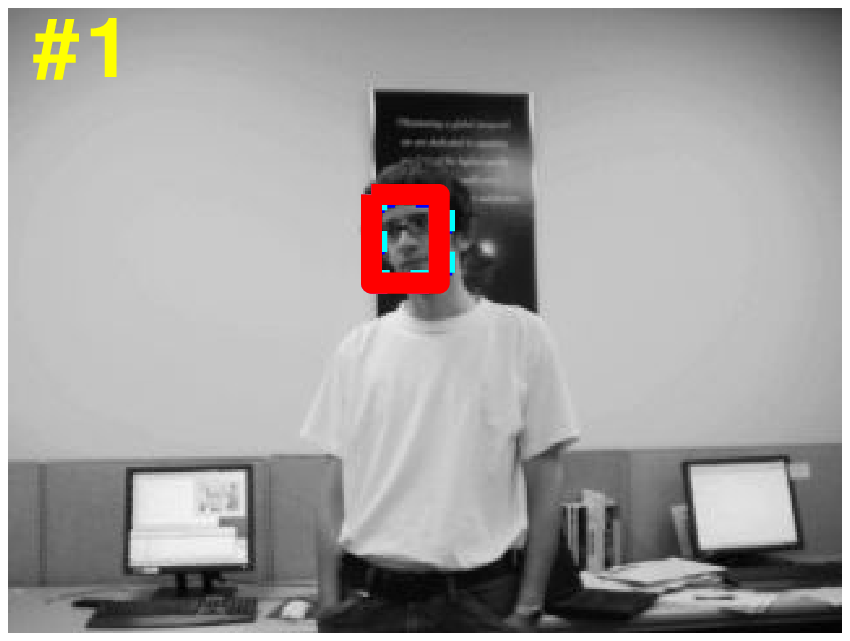,width=0.16\textwidth}
\epsfig{file=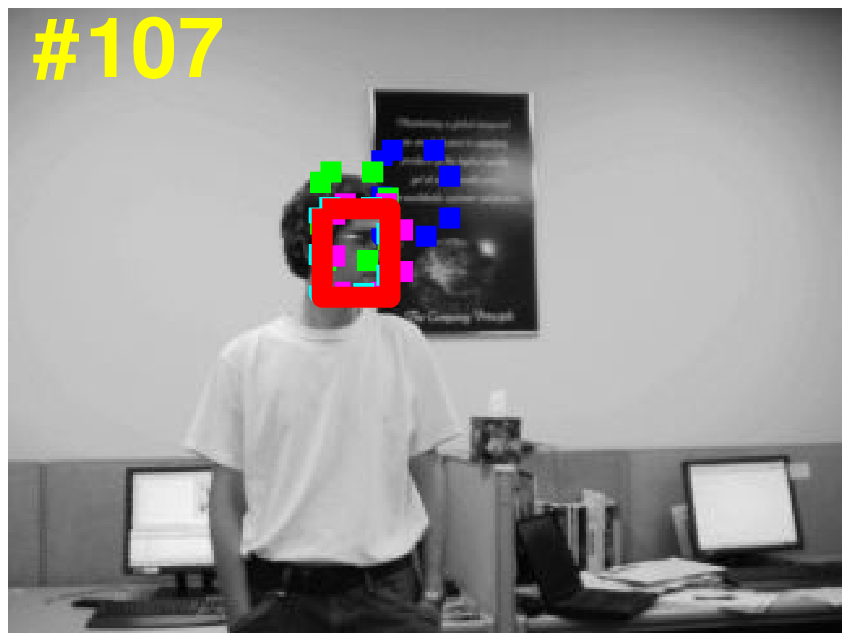,width=0.16\textwidth}
\epsfig{file=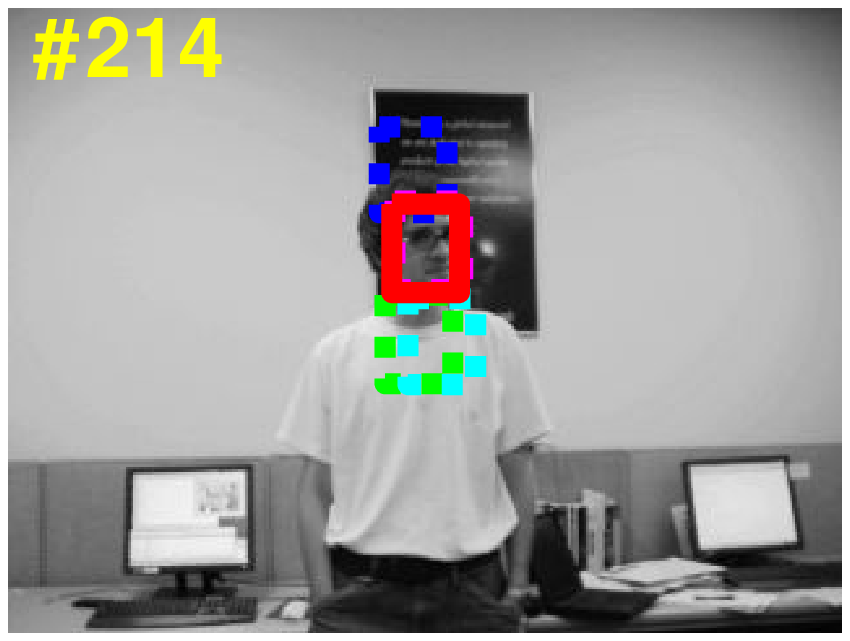,width=0.16\textwidth}
\epsfig{file=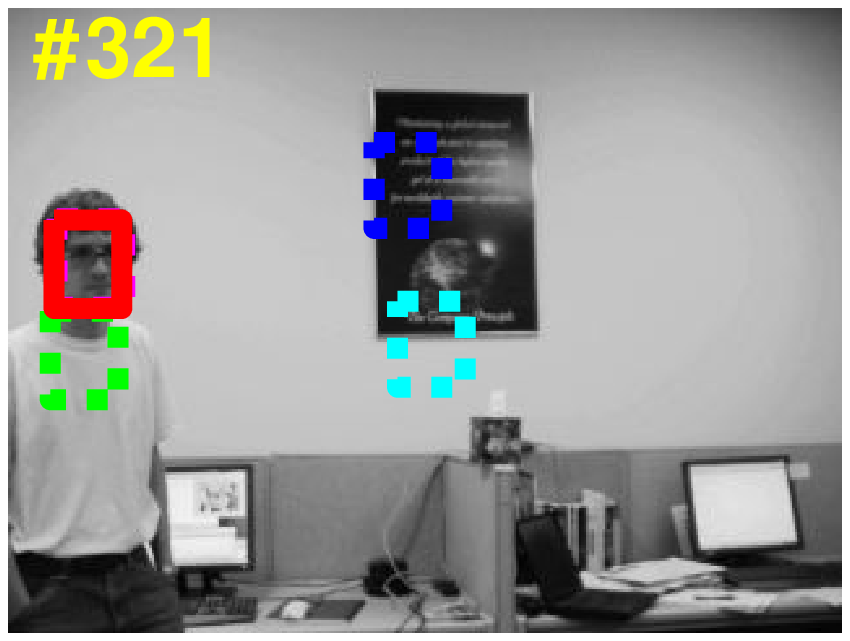,width=0.16\textwidth}
\epsfig{file=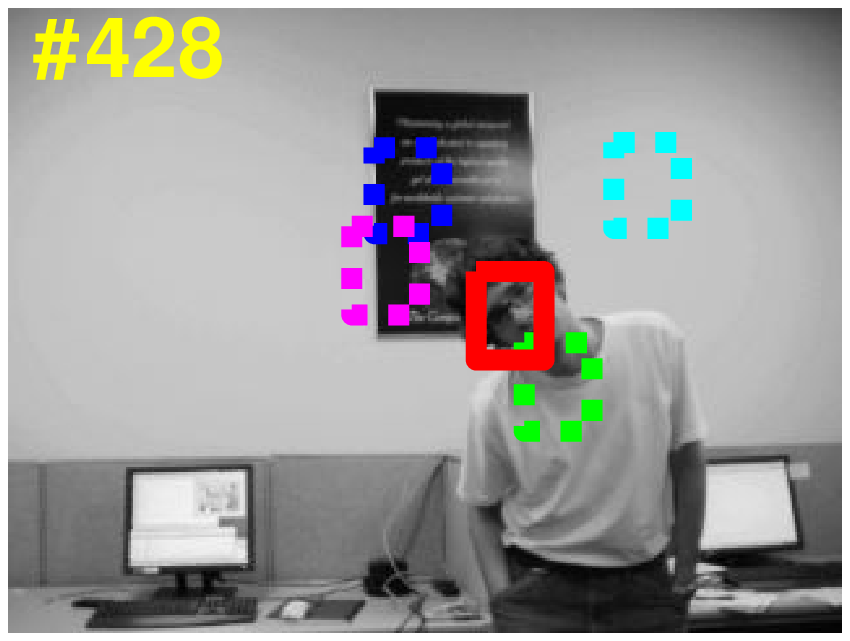,width=0.16\textwidth}
\epsfig{file=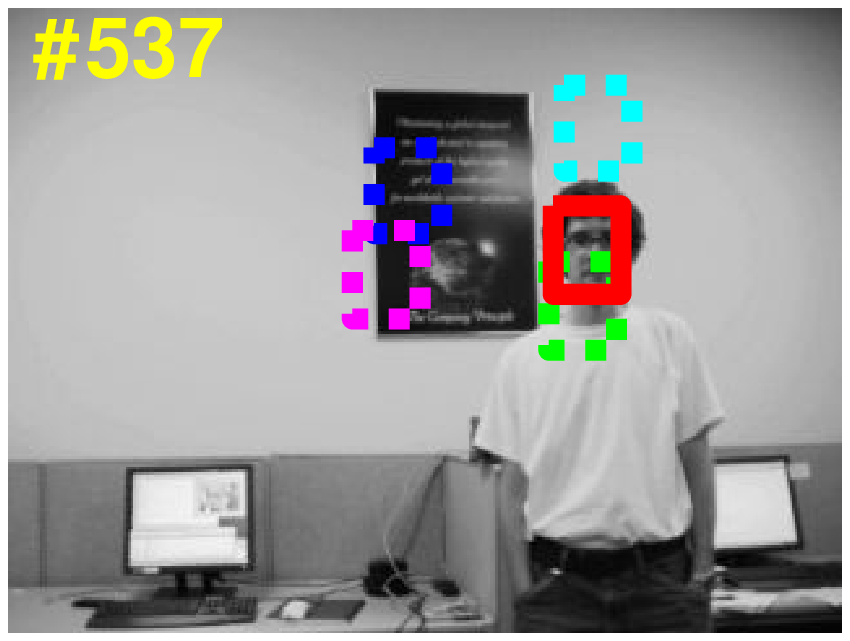,width=0.16\textwidth}}
\\ \vspace{-0.1in}

\centering \subfloat[Mountainbike]{
\epsfig{file=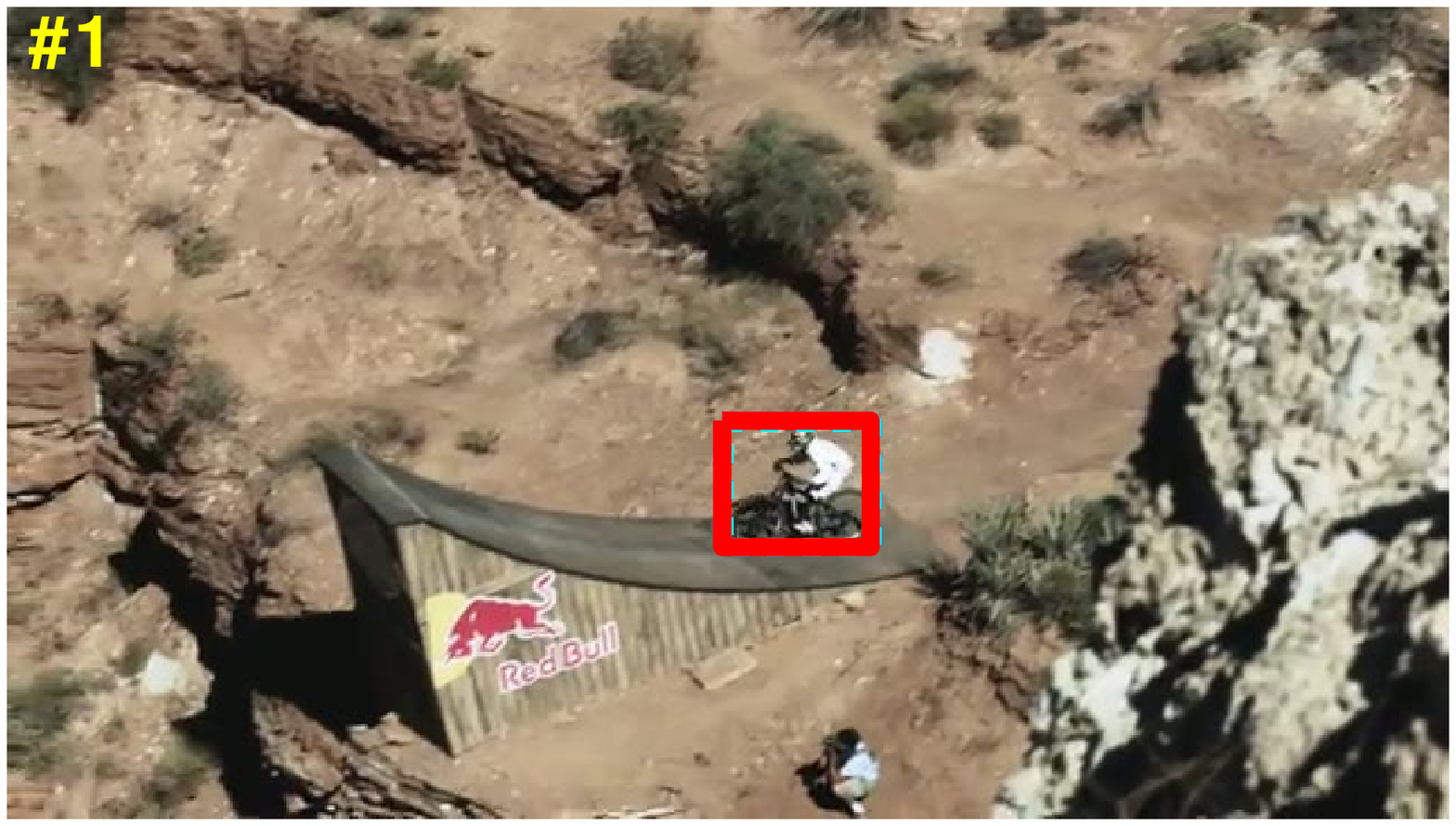,width=0.16\textwidth}
\epsfig{file=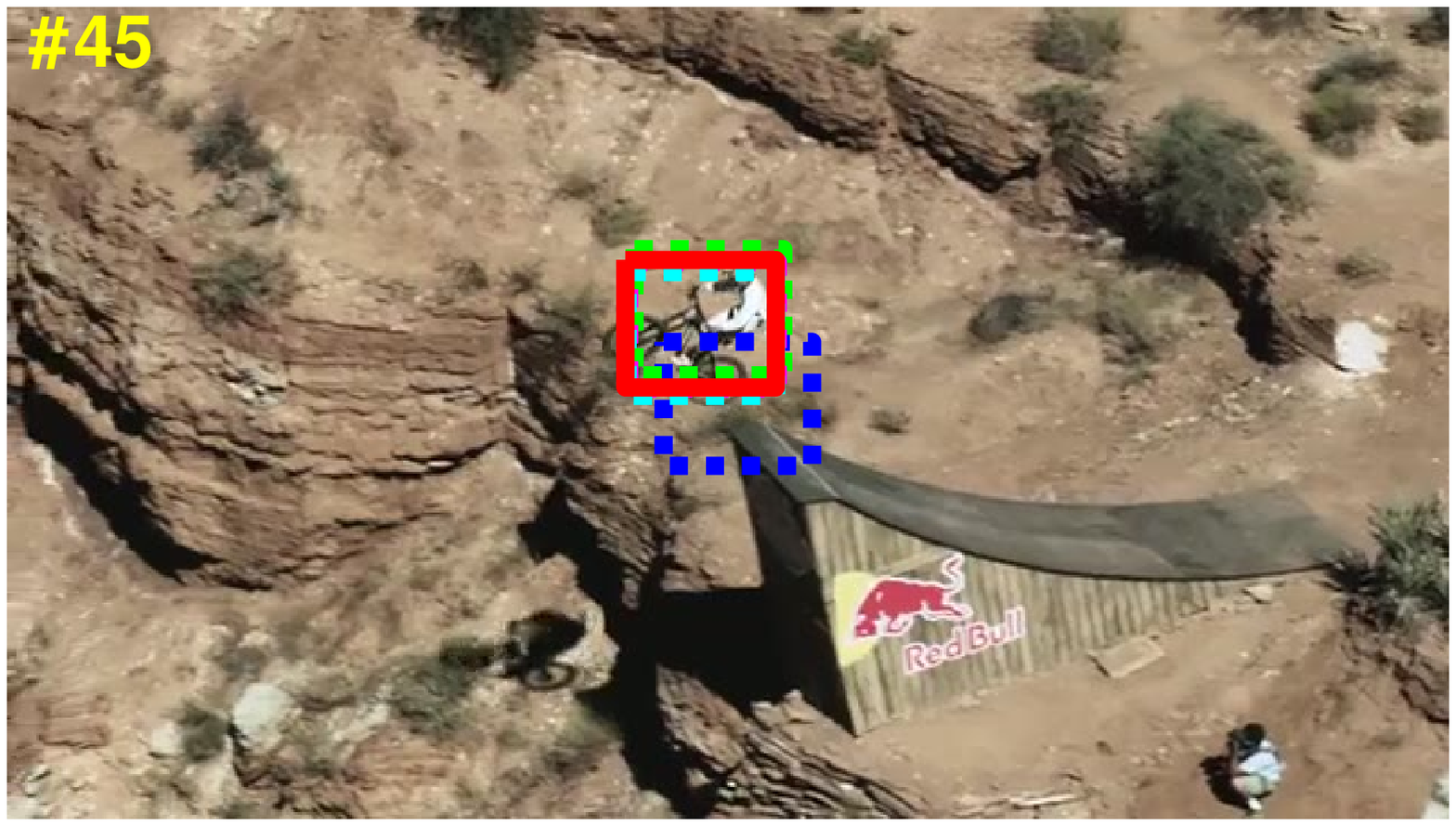,width=0.16\textwidth}
\epsfig{file=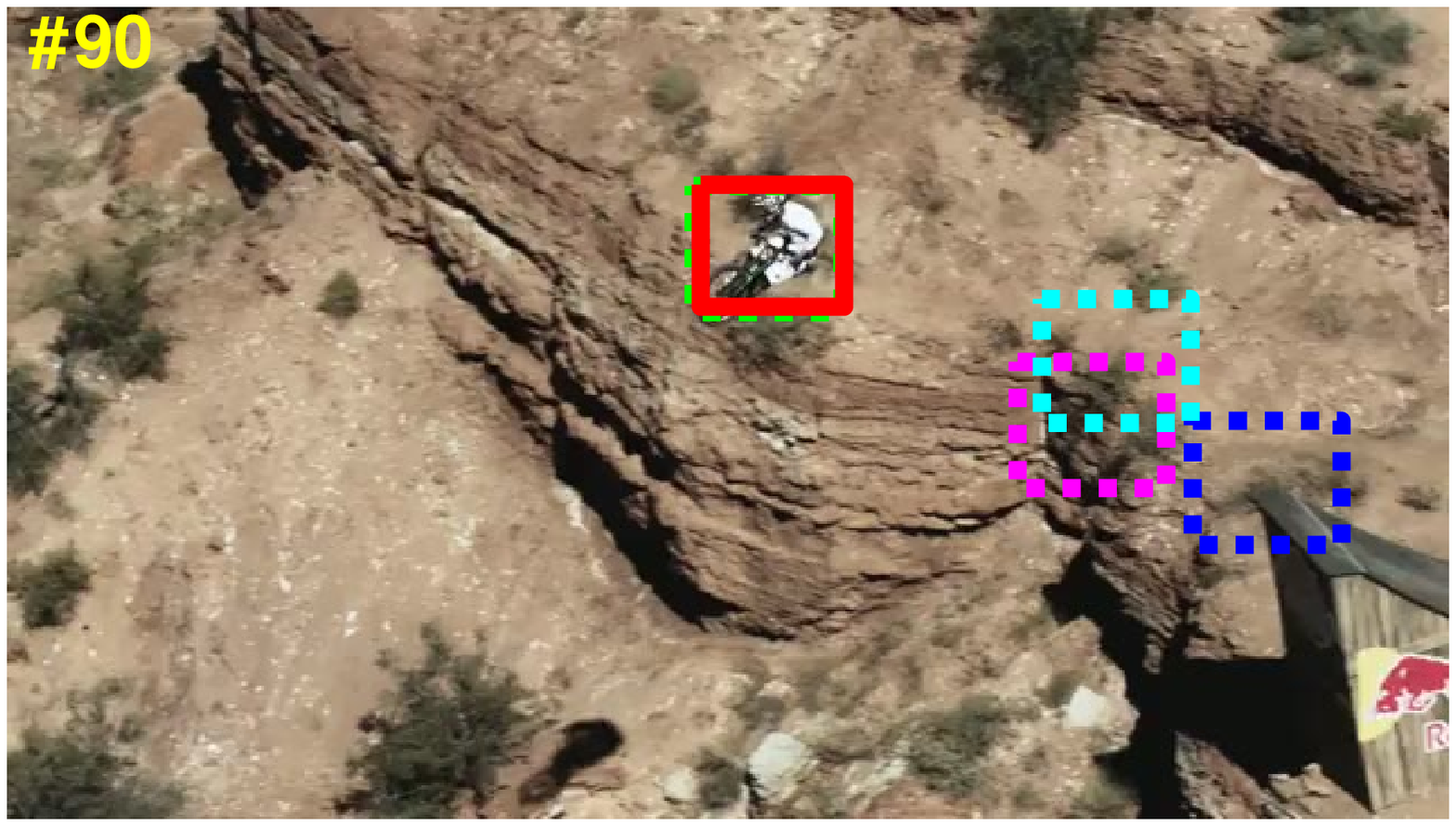,width=0.16\textwidth}
\epsfig{file=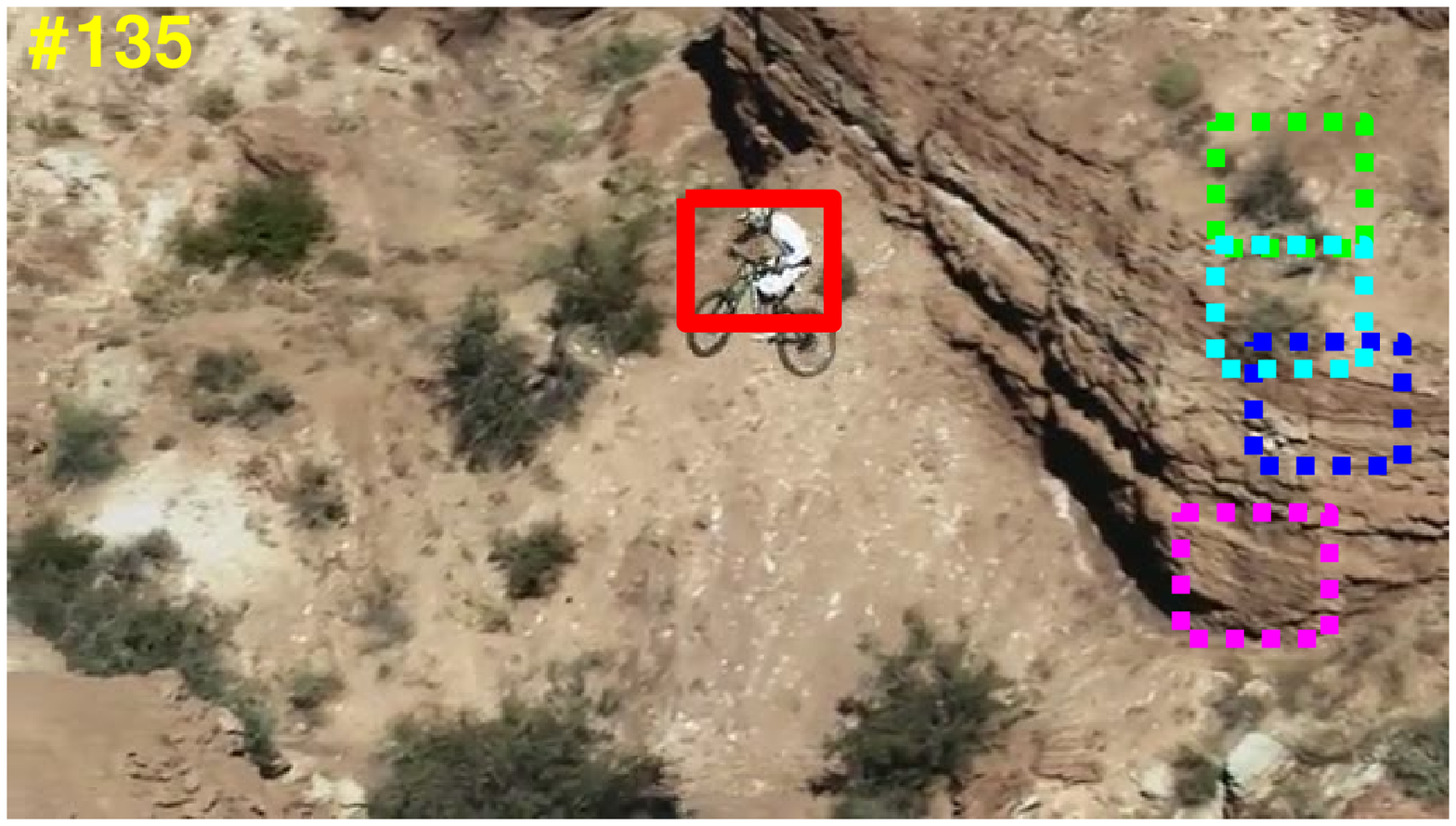,width=0.16\textwidth}
\epsfig{file=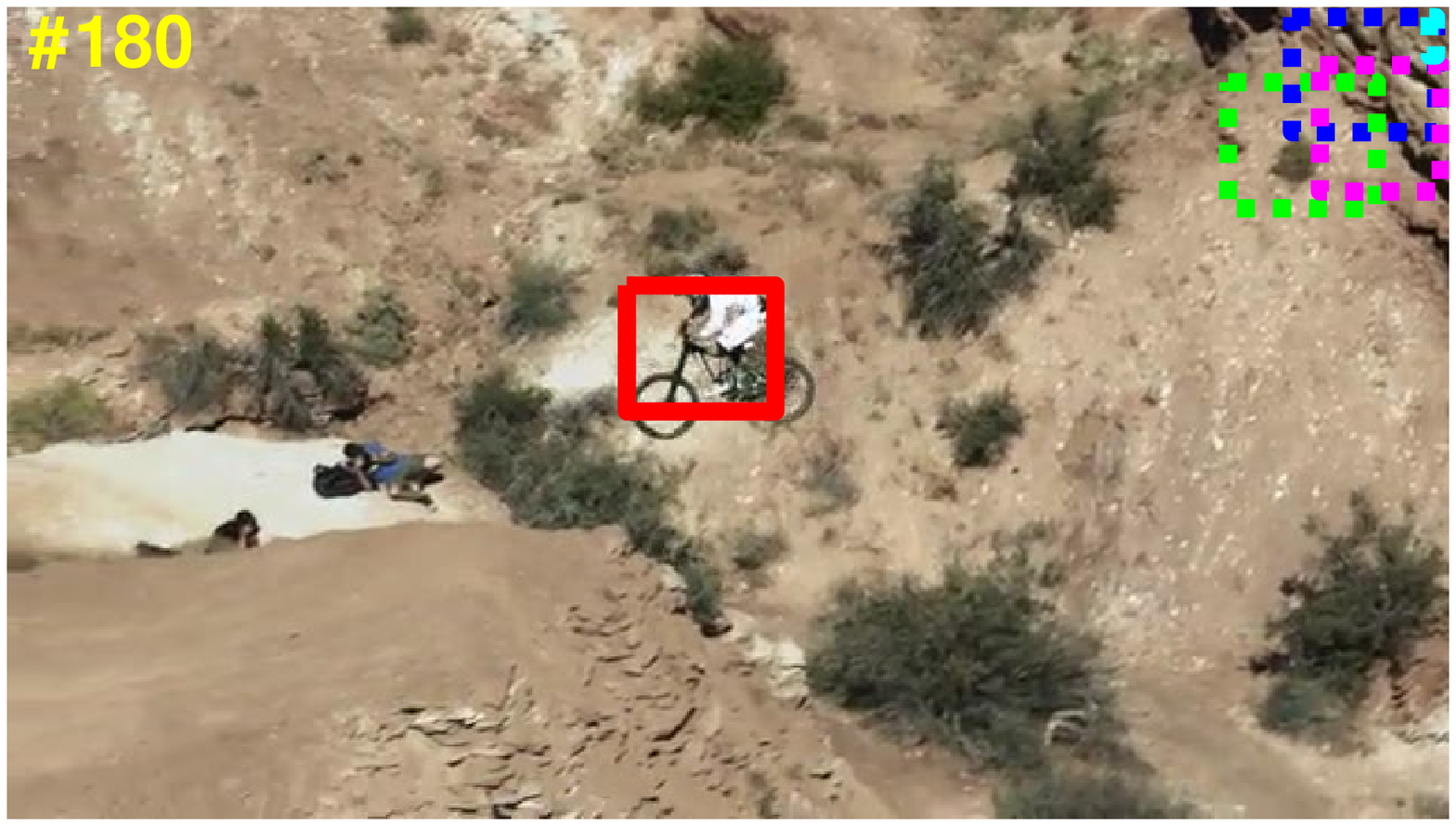,width=0.16\textwidth}
\epsfig{file=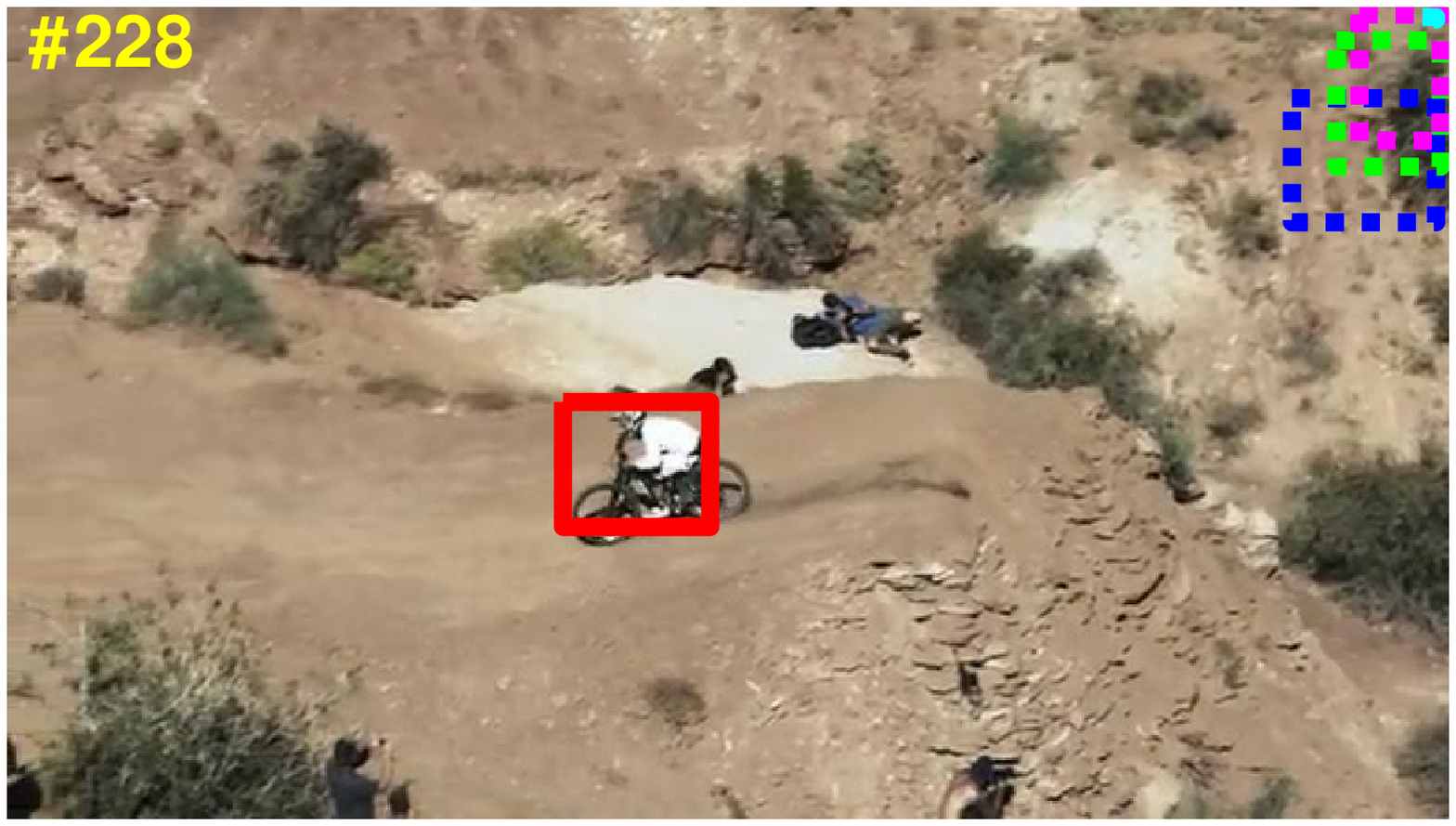,width=0.16\textwidth}}
\\ \vspace{-0.1in}

\centering \subfloat[Sylvester]{
\epsfig{file=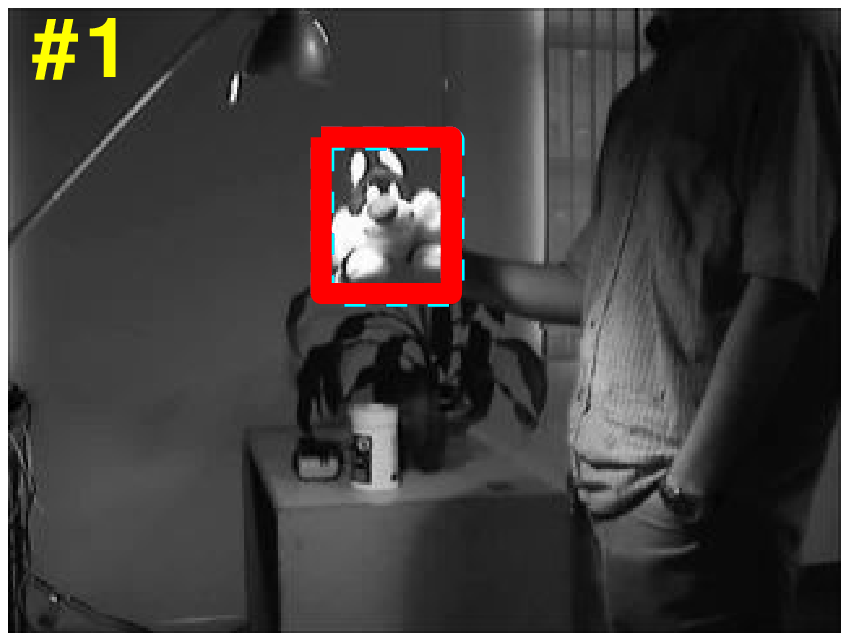,width=0.16\textwidth}
\epsfig{file=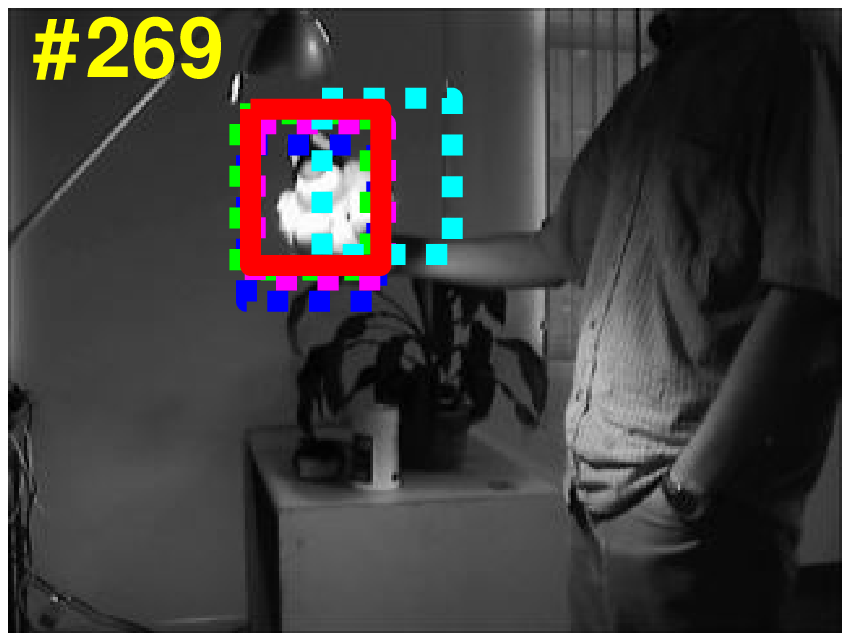,width=0.16\textwidth}
\epsfig{file=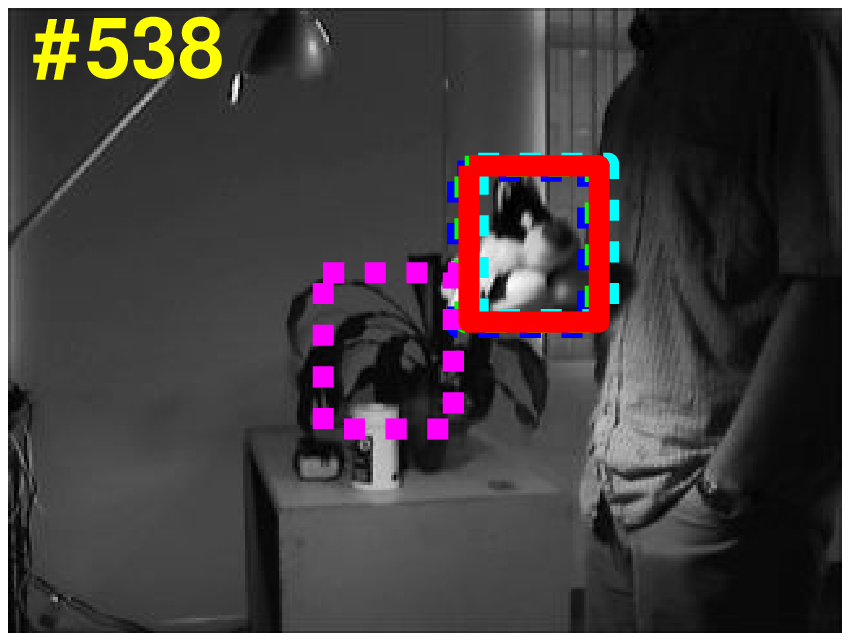,width=0.16\textwidth}
\epsfig{file=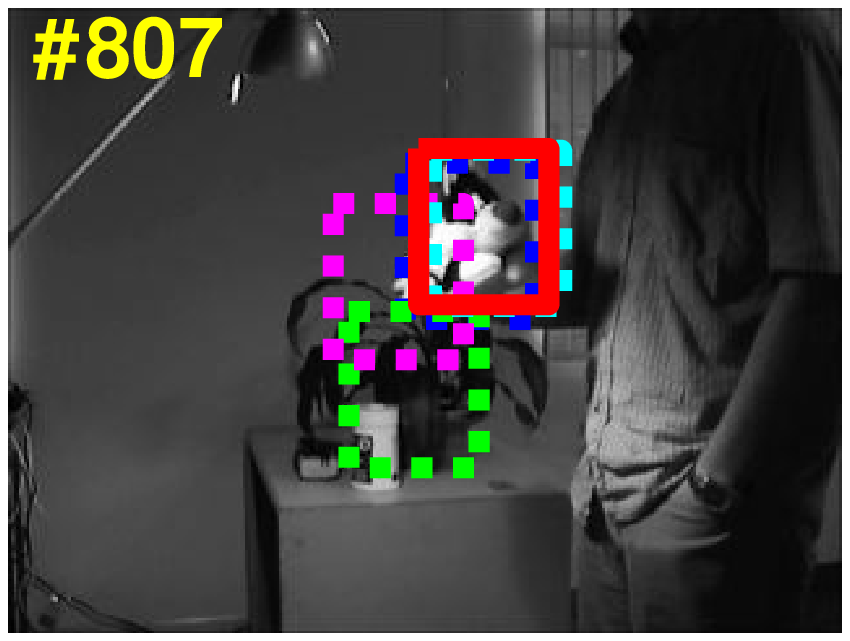,width=0.16\textwidth}
\epsfig{file=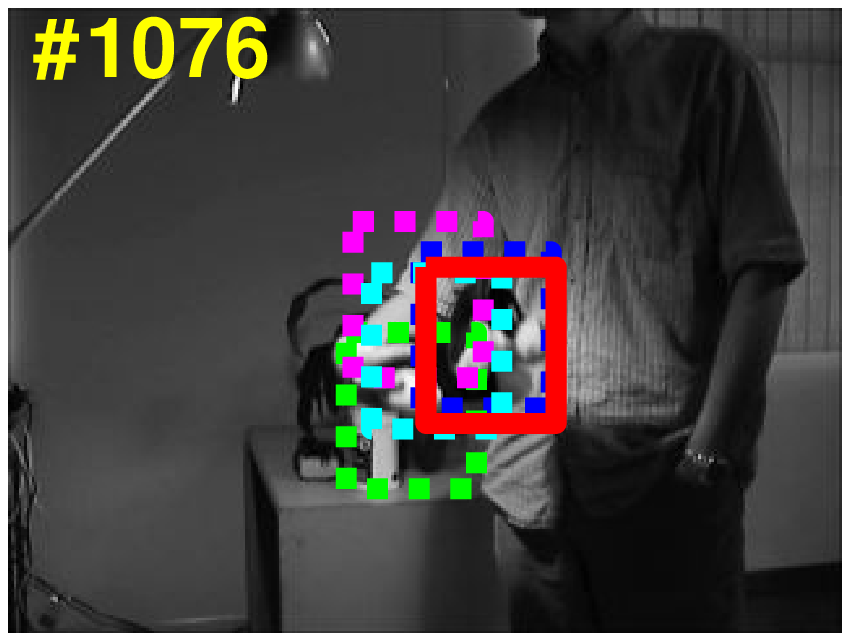,width=0.16\textwidth}
\epsfig{file=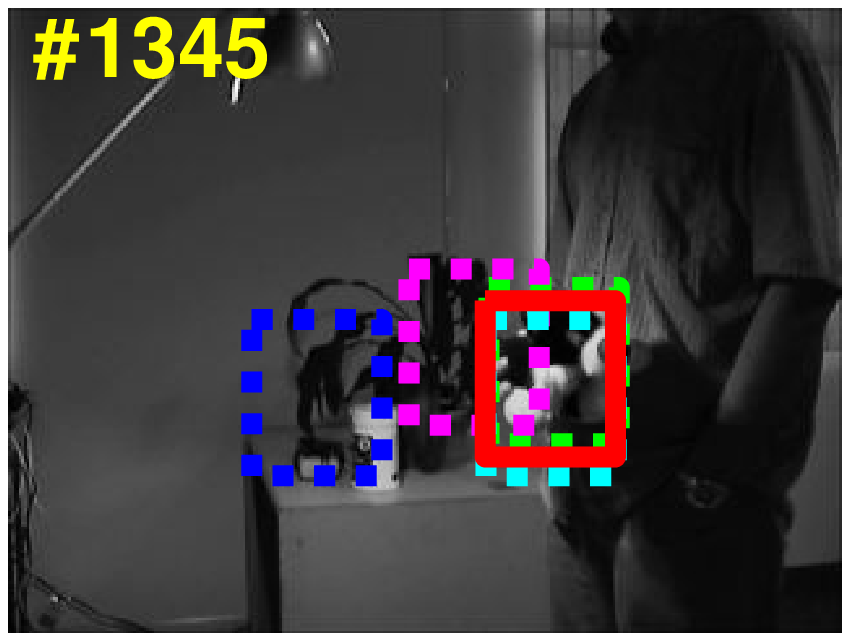,width=0.16\textwidth}}
\\ \vspace{-0.1in}

\centering \subfloat[Tiger1]{
\epsfig{file=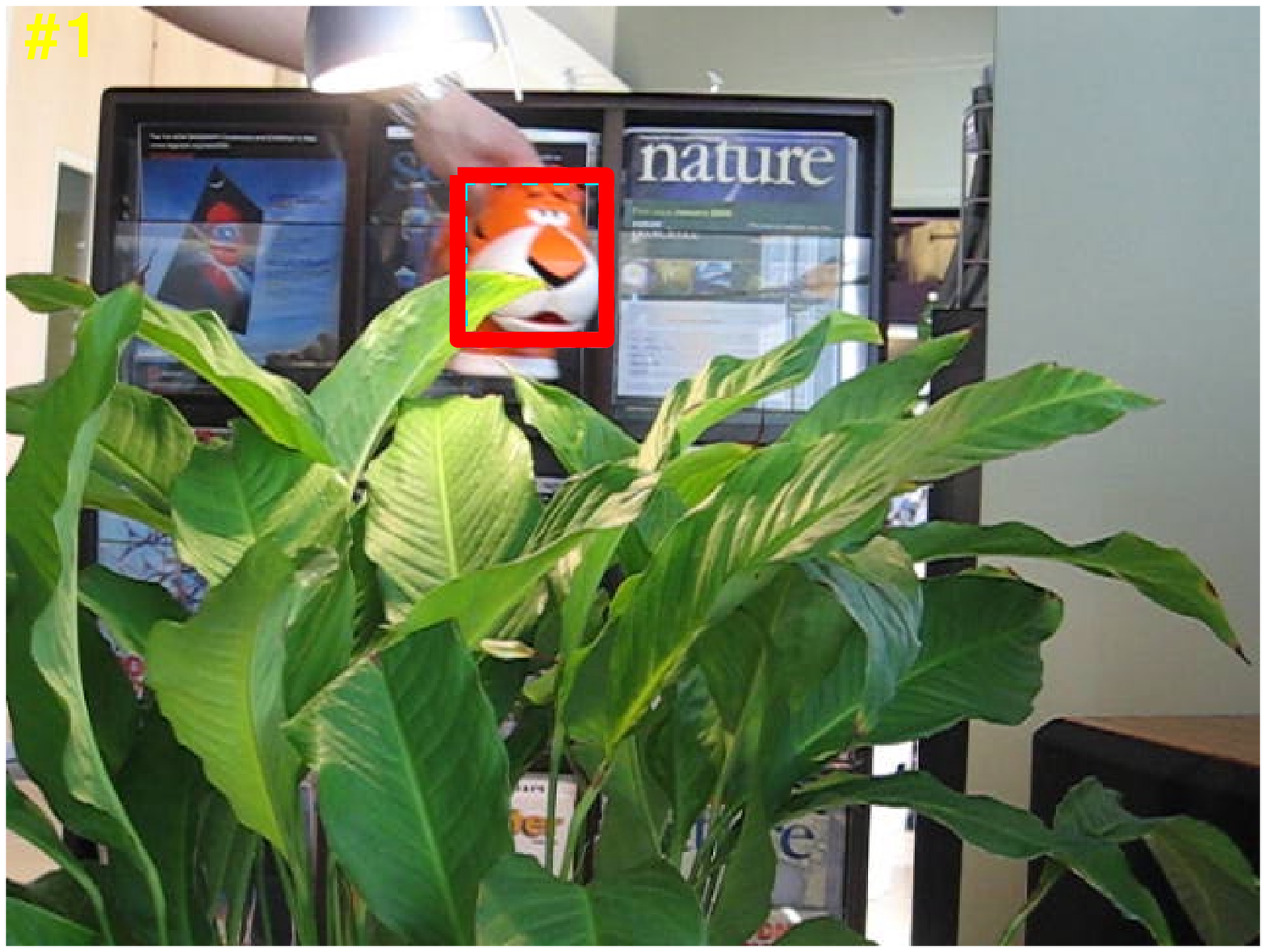,width=0.16\textwidth}
\epsfig{file=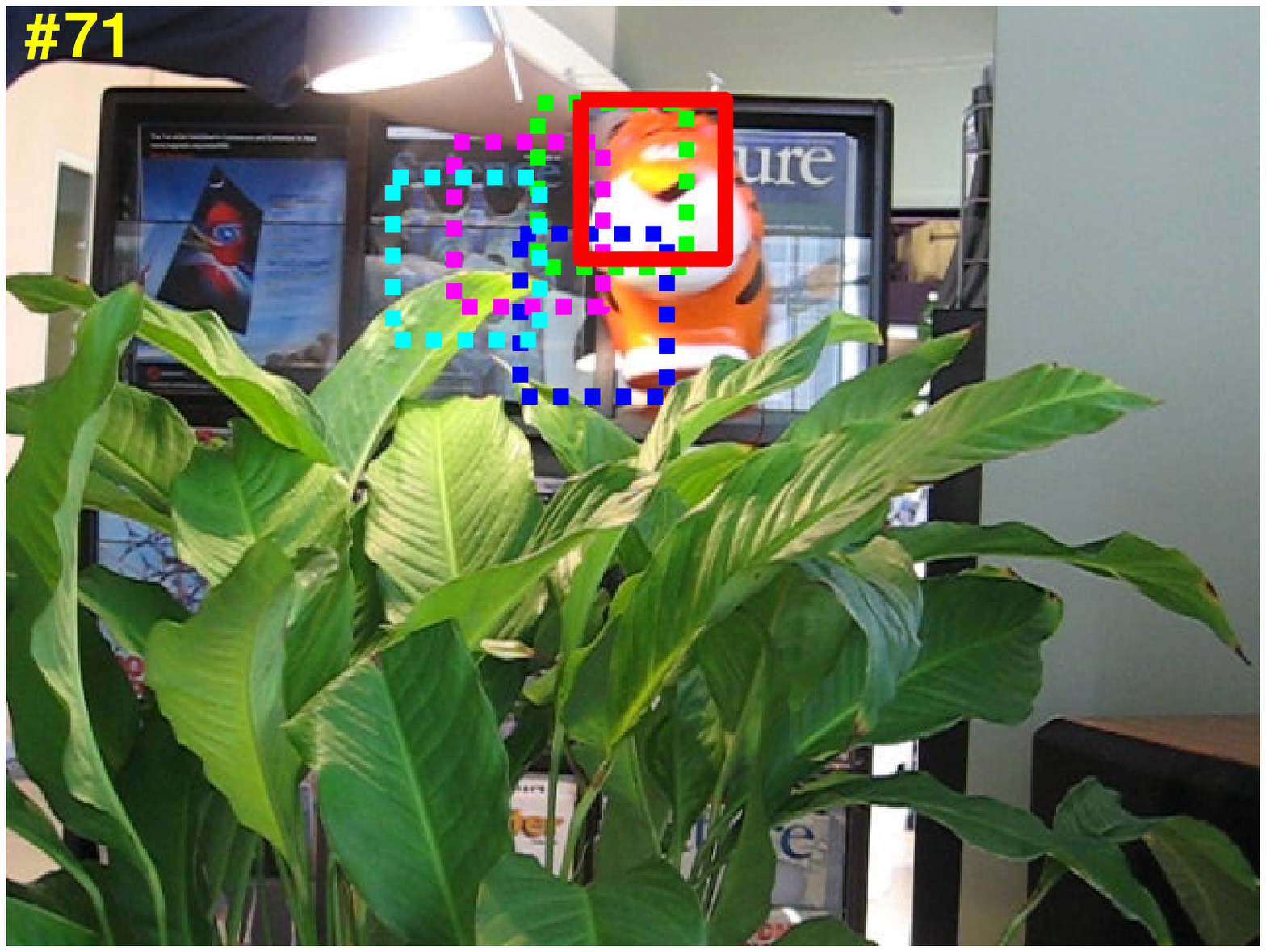,width=0.16\textwidth}
\epsfig{file=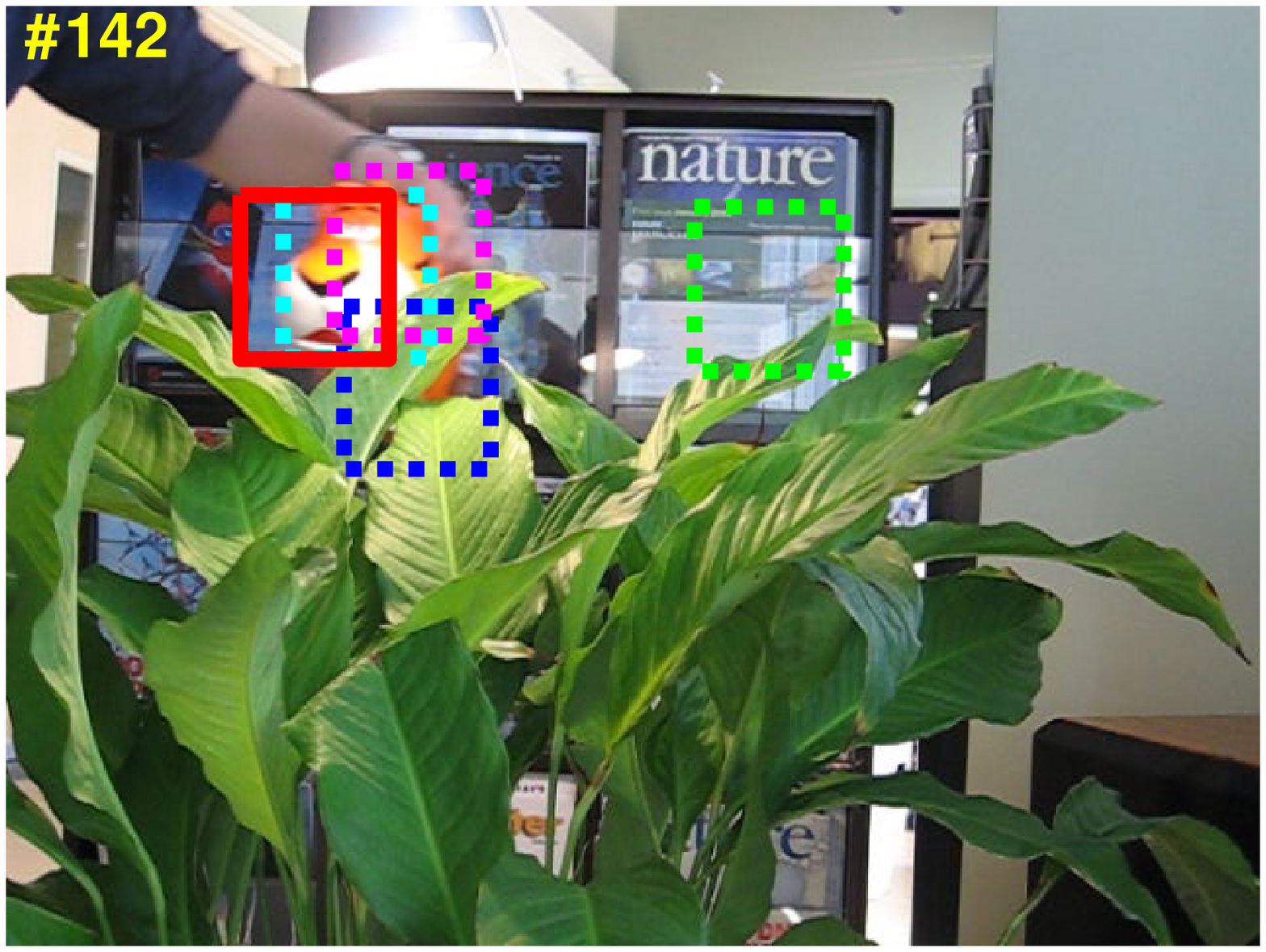,width=0.16\textwidth}
\epsfig{file=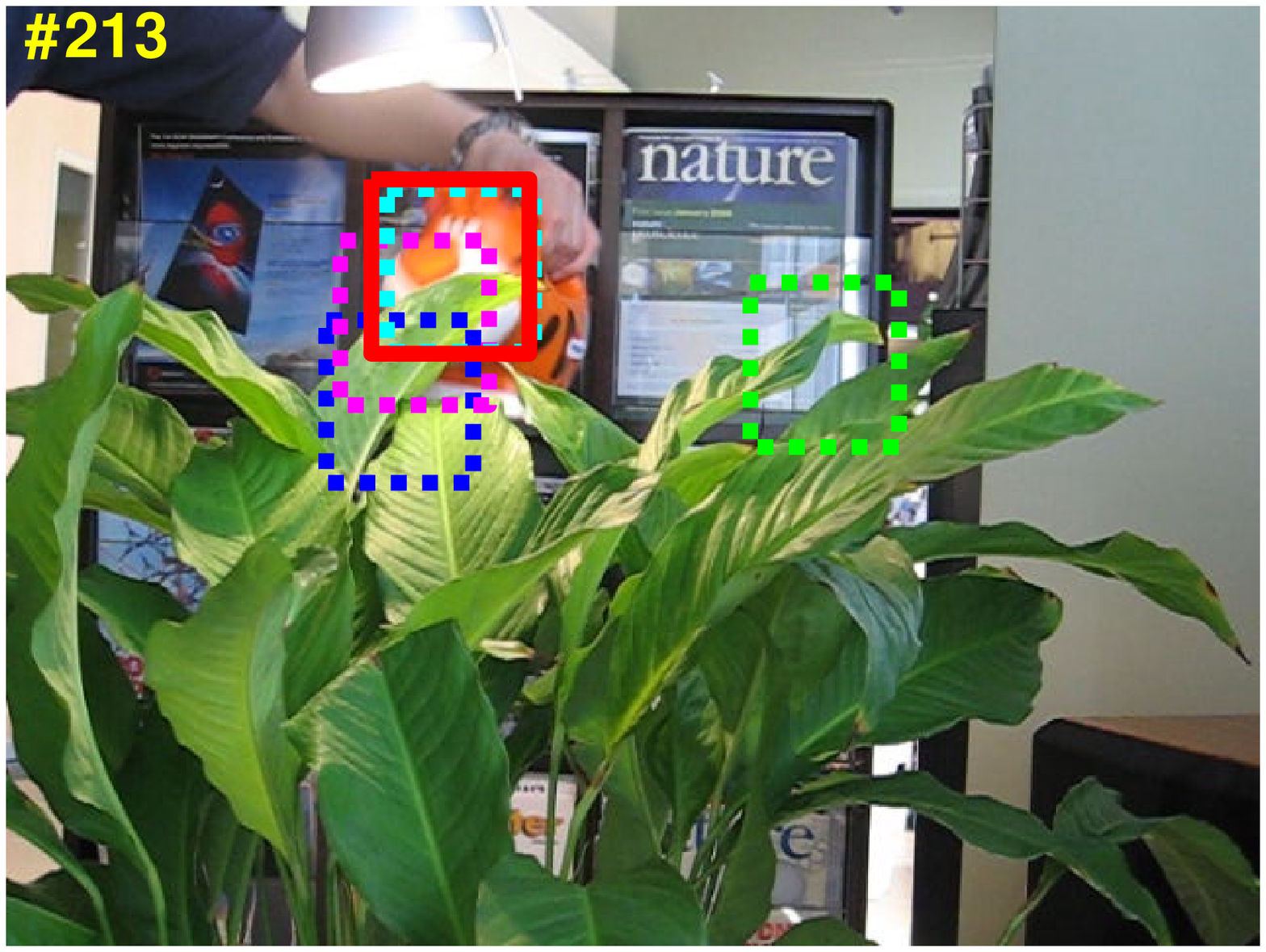,width=0.16\textwidth}
\epsfig{file=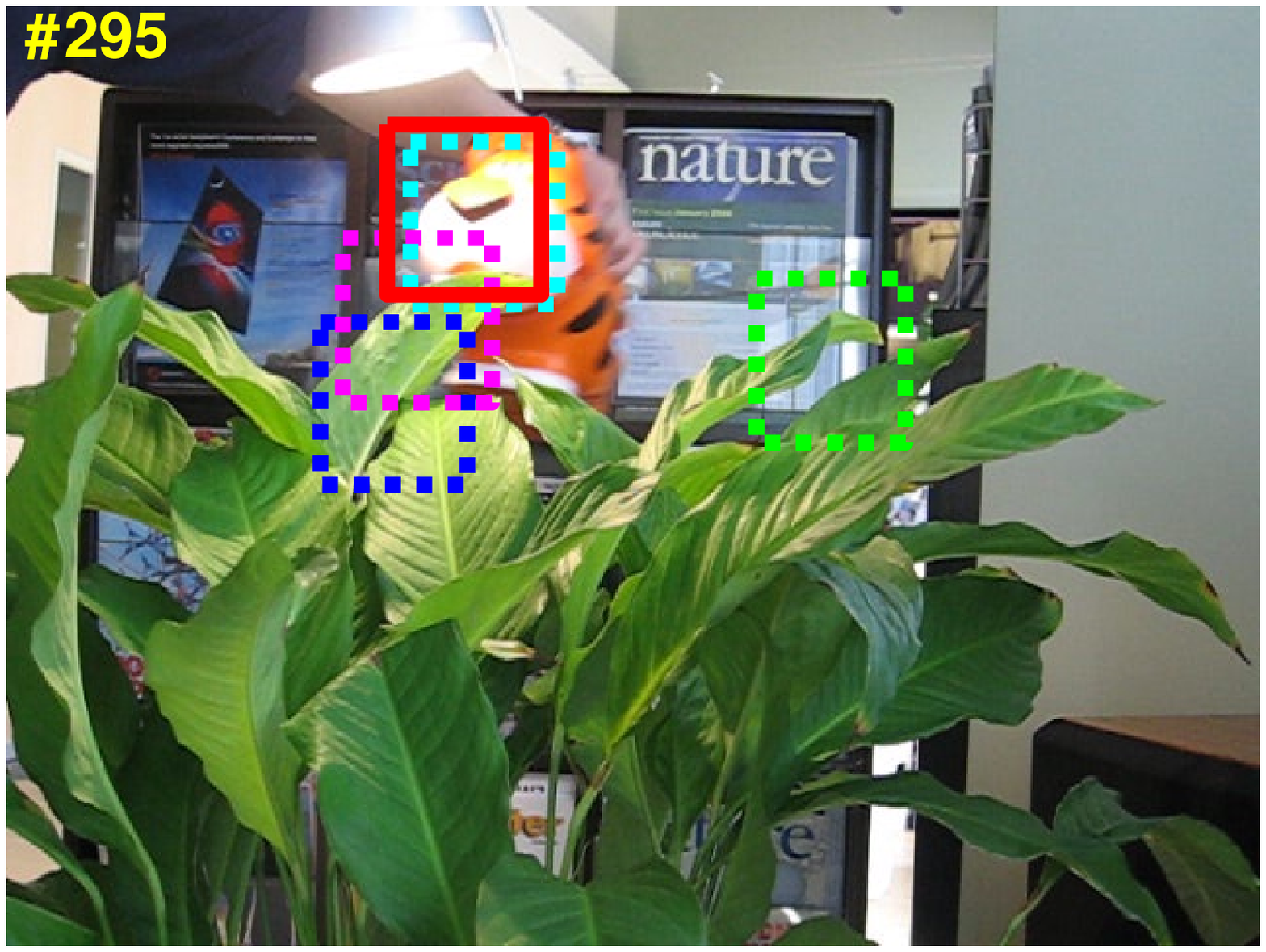,width=0.16\textwidth}
\epsfig{file=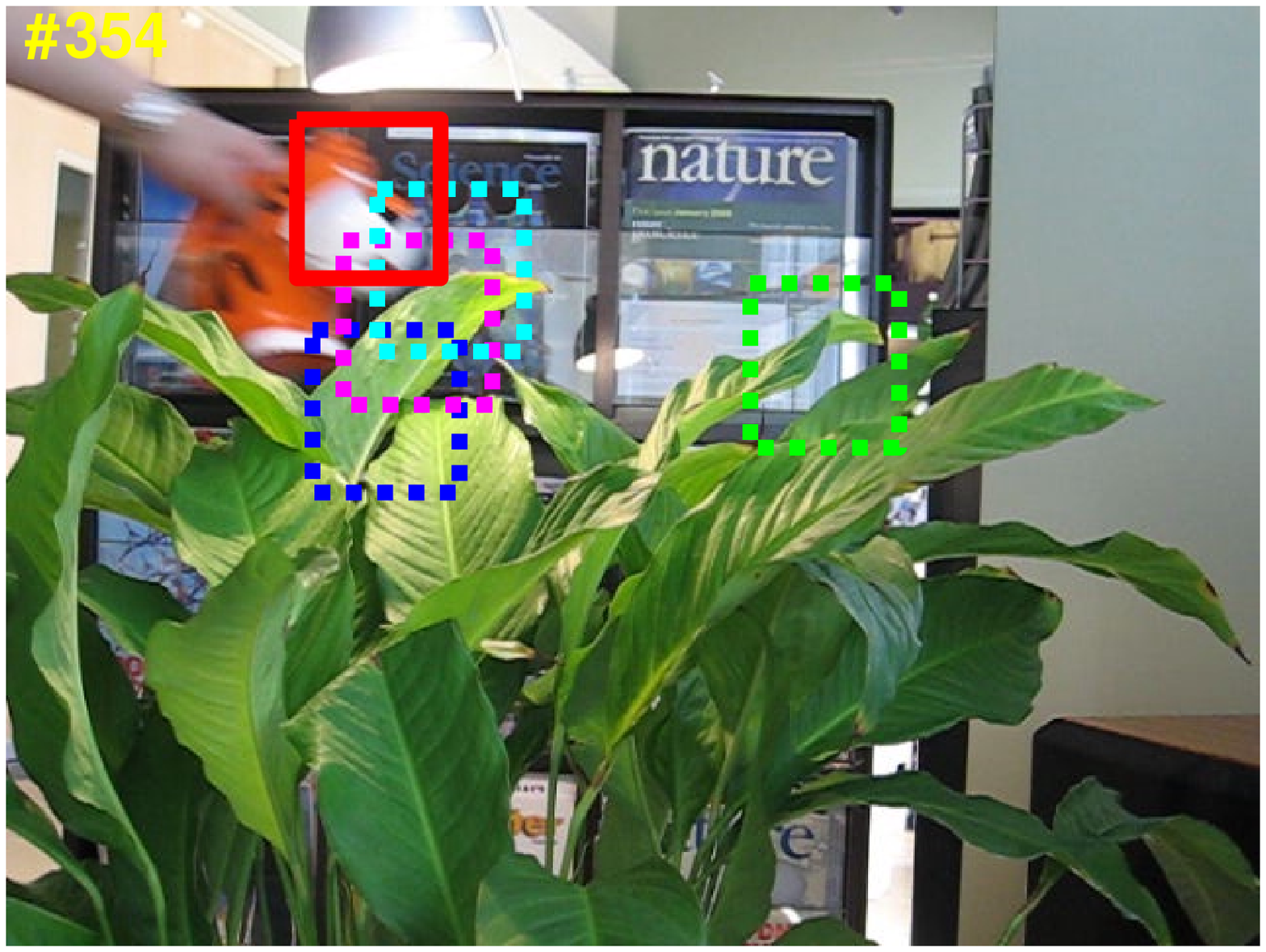,width=0.16\textwidth}}
\\ \vspace{-0.1in}

\centering \subfloat[Tiger2]{
\epsfig{file=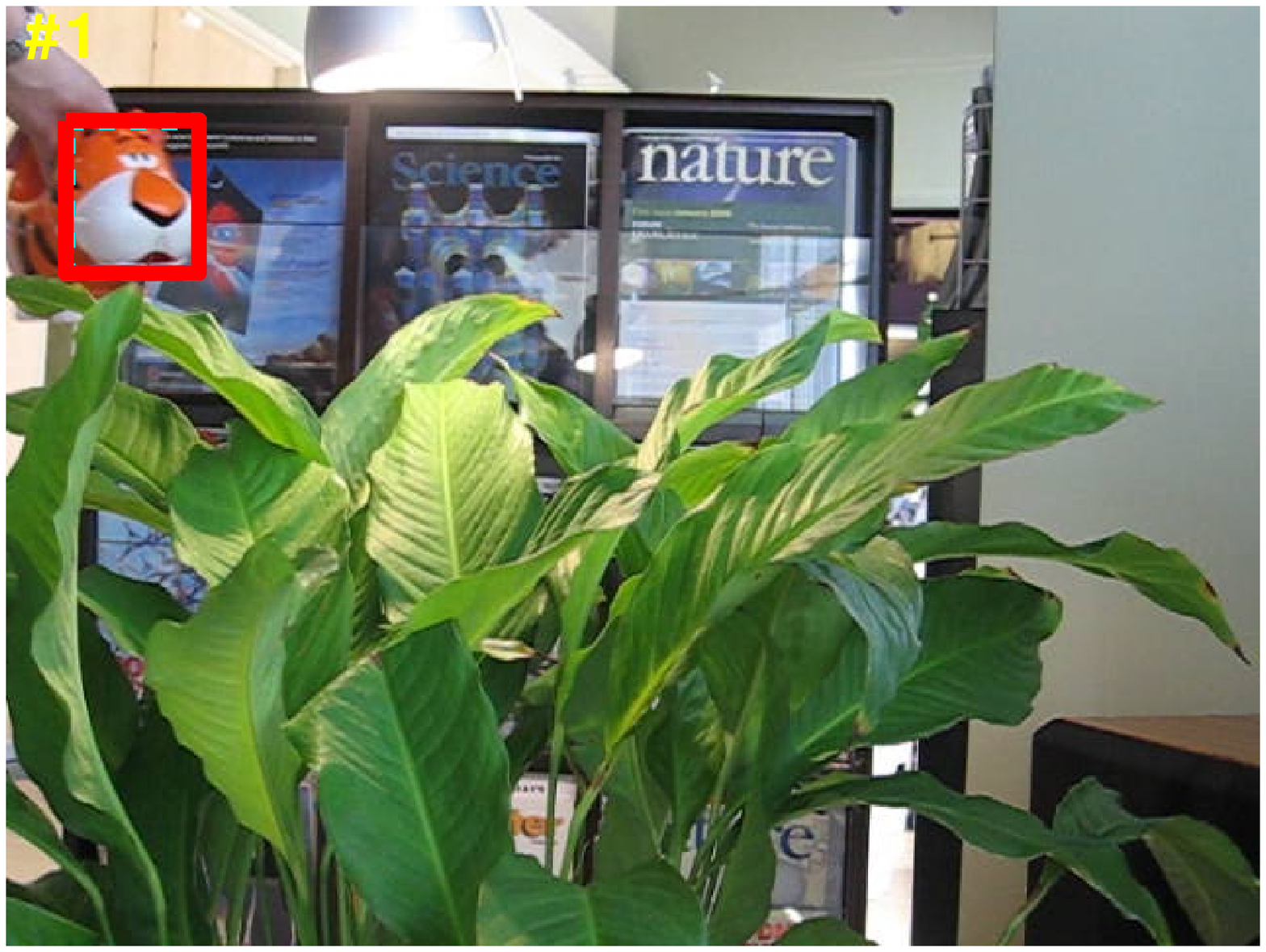,width=0.16\textwidth}
\epsfig{file=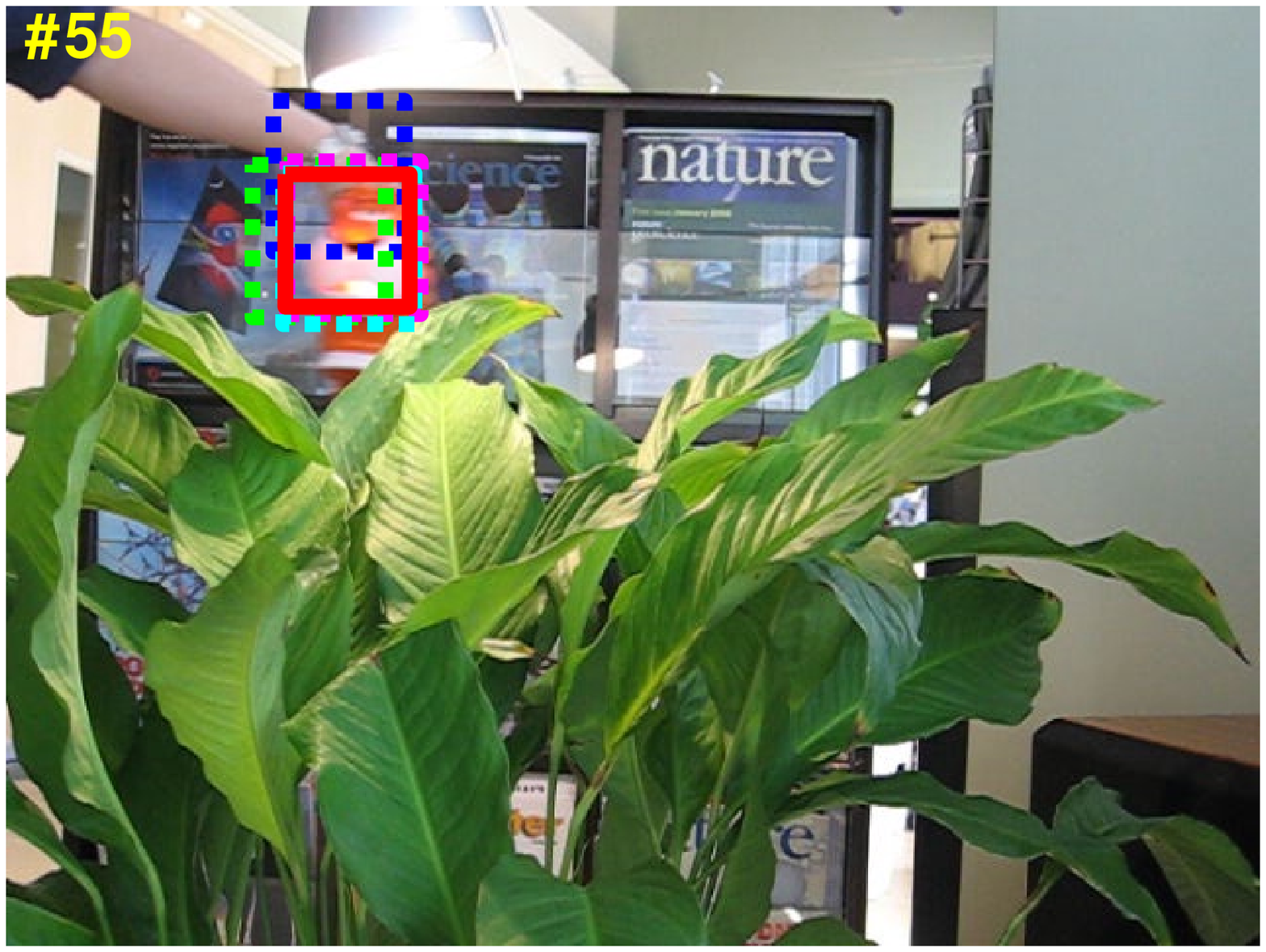,width=0.16\textwidth}
\epsfig{file=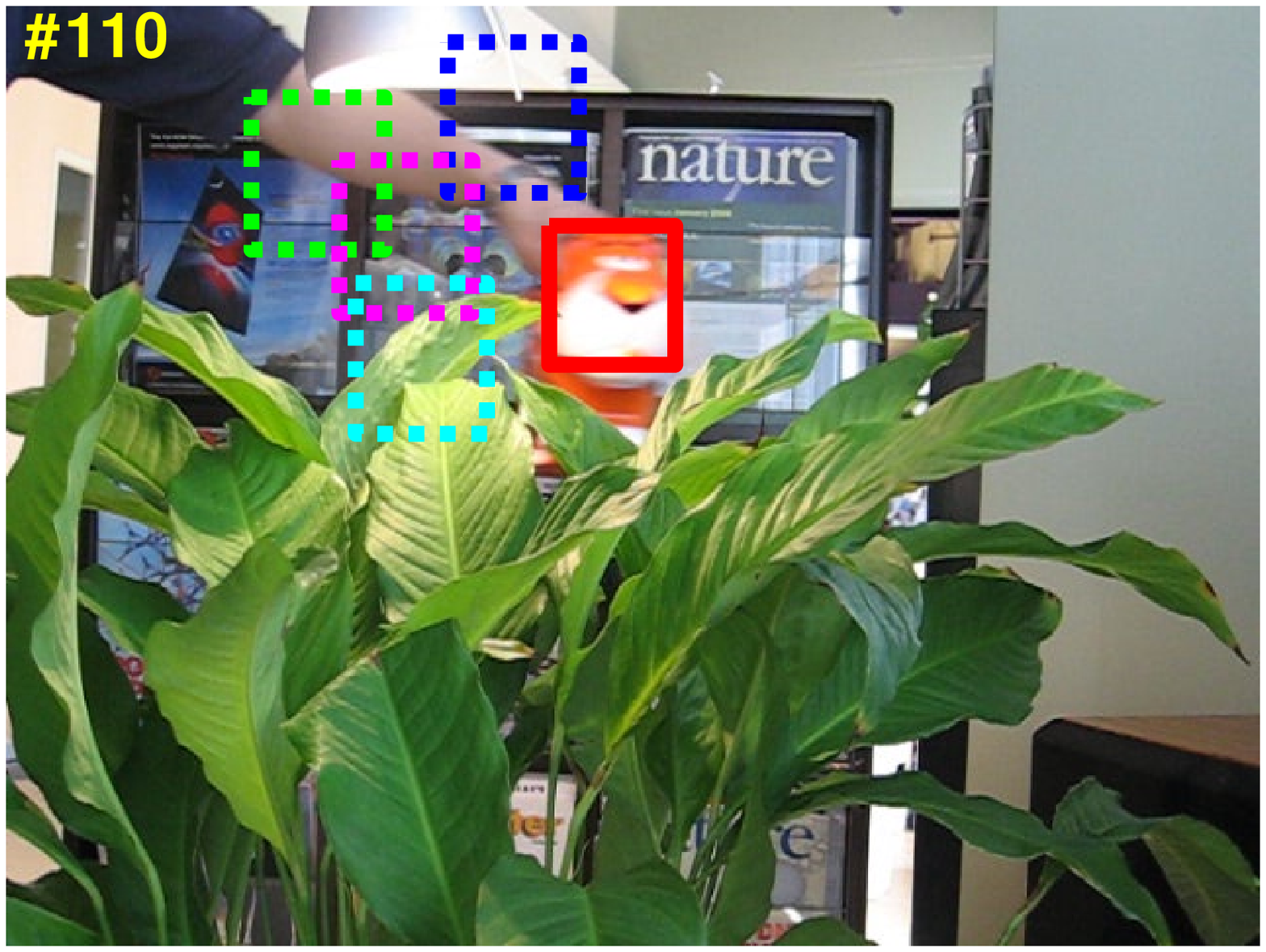,width=0.16\textwidth}
\epsfig{file=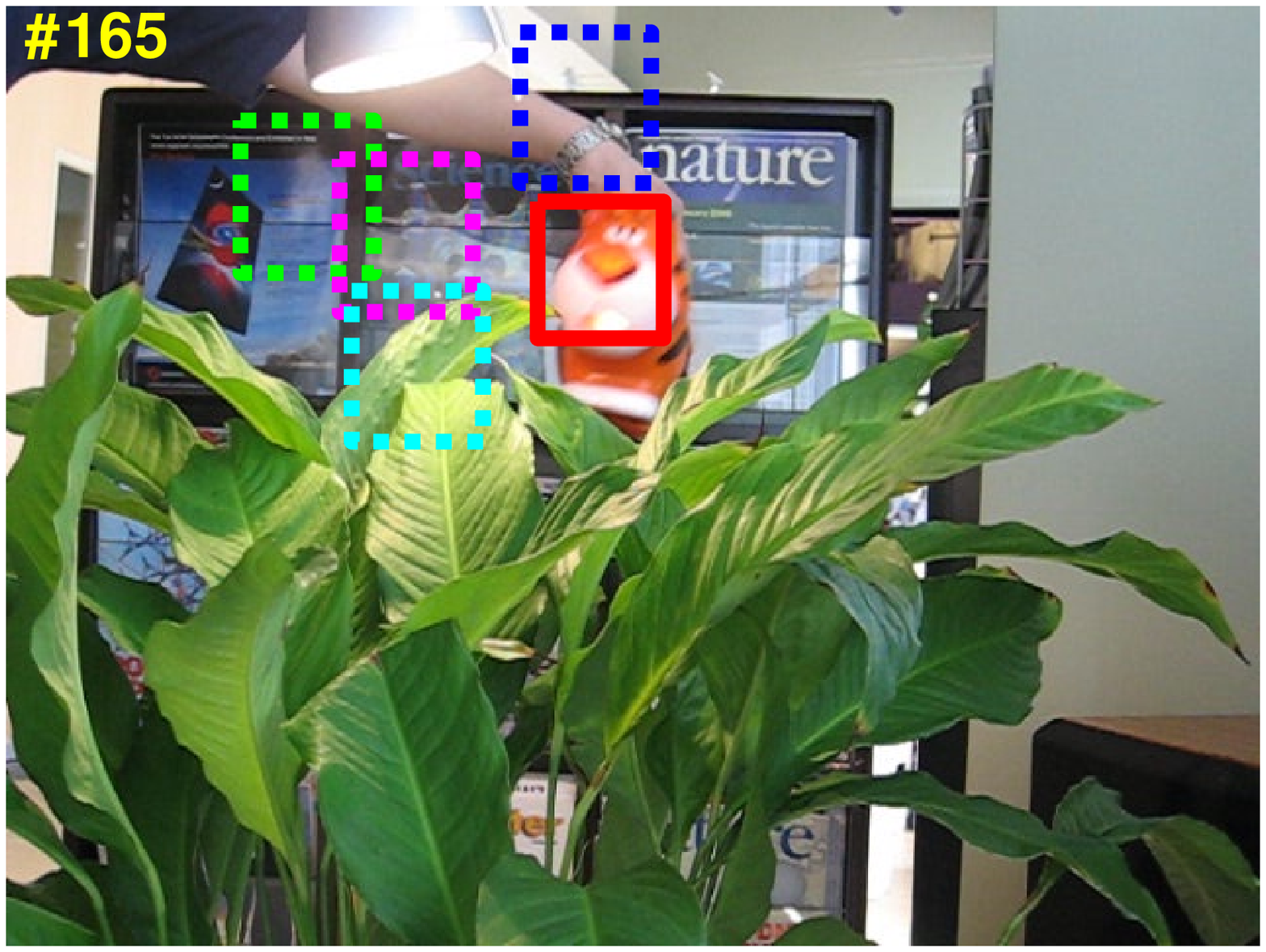,width=0.16\textwidth}
\epsfig{file=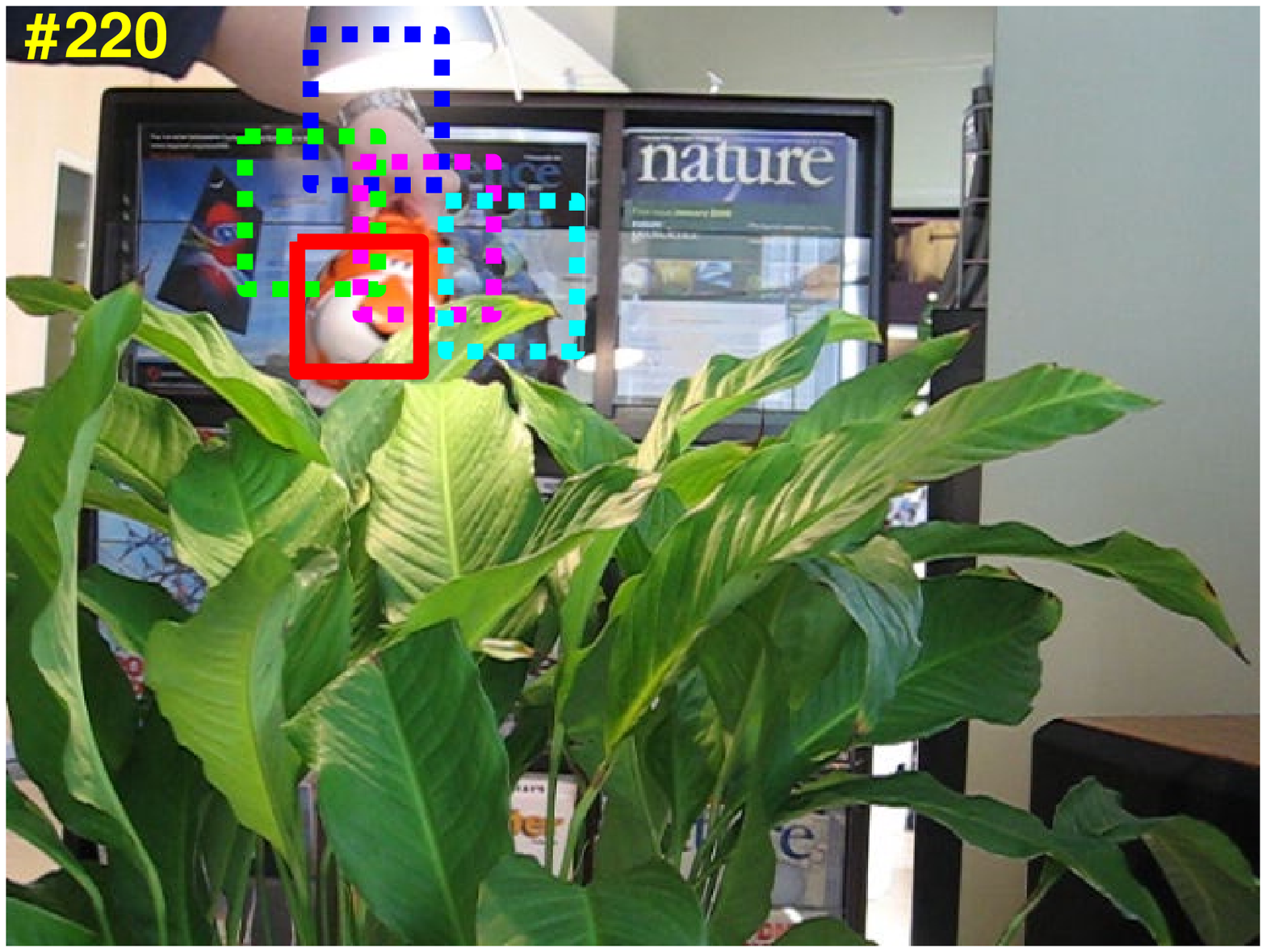,width=0.16\textwidth}
\epsfig{file=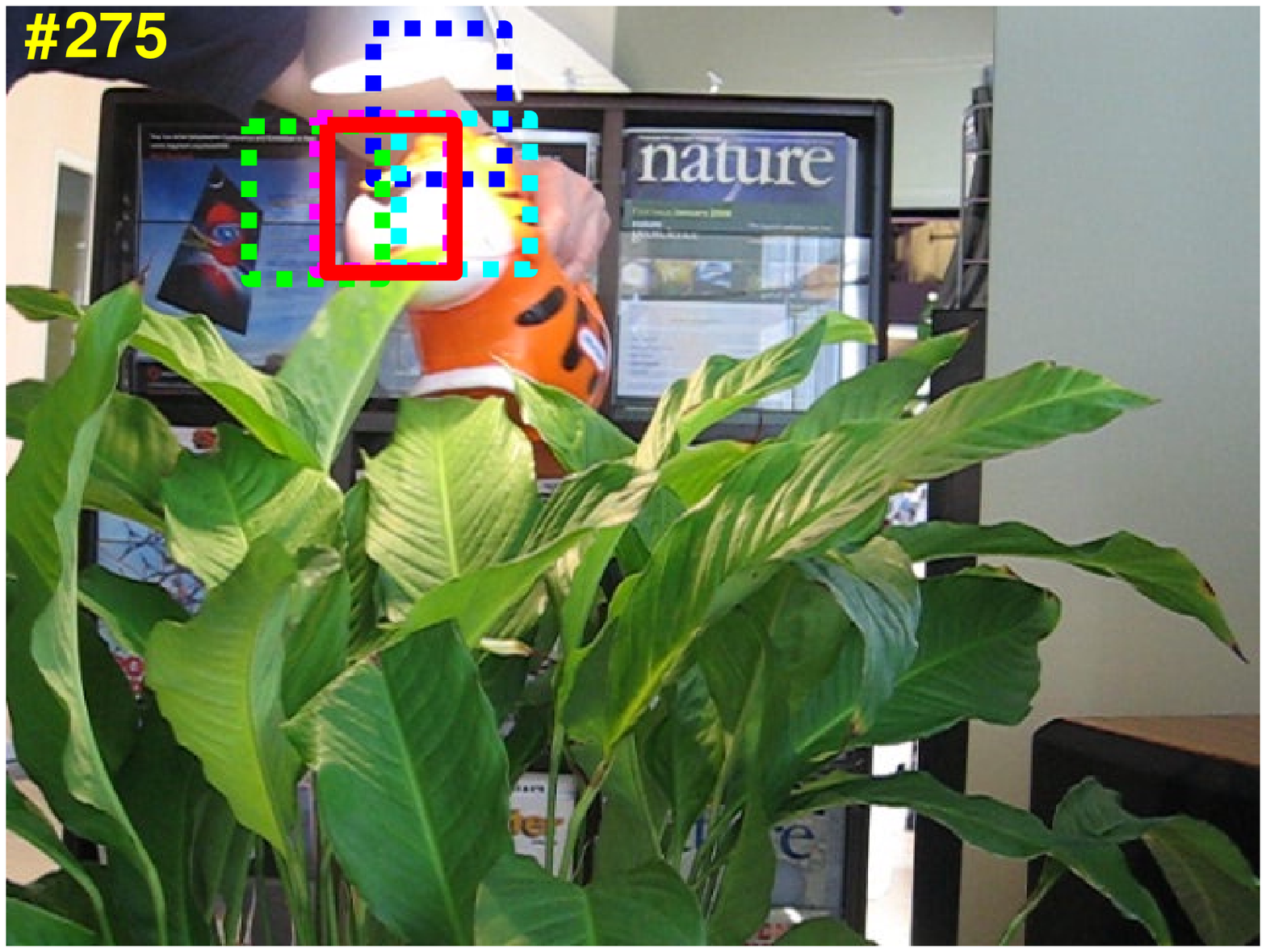,width=0.16\textwidth}}

\caption{Qualitative results on sequences with in-plane rotations. The purple, green, cyan, blue and red bounding boxes refer to ASLA\cite{DBLP:conf/cvpr/JiaLY12}\_RAW, ASLA\cite{DBLP:conf/cvpr/JiaLY12}\_HOG, $\ell_1$\_APG \cite{DBLP:conf/cvpr/BaoWLJ12}, CT\_DIF \cite{DBLP:conf/eccv/Zhang0Y12} and our tracker respectively. This figure is better viewed in color.} \label{fig:quali_inPlaneRotation}
\end{figure*}

\begin{figure*}
\centering \subfloat[Freeman1]{
\epsfig{file=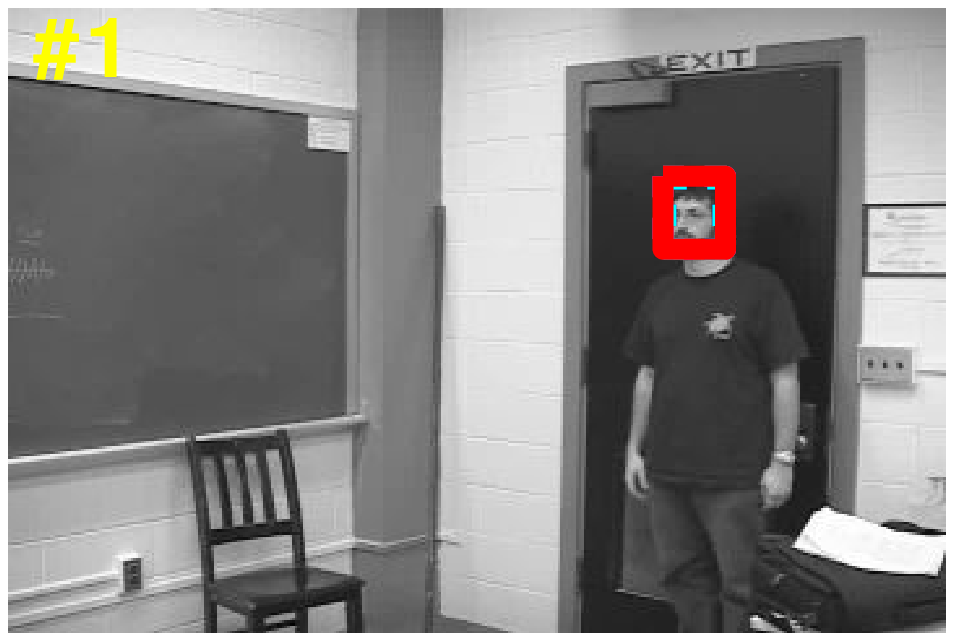,width=0.16\textwidth}
\epsfig{file=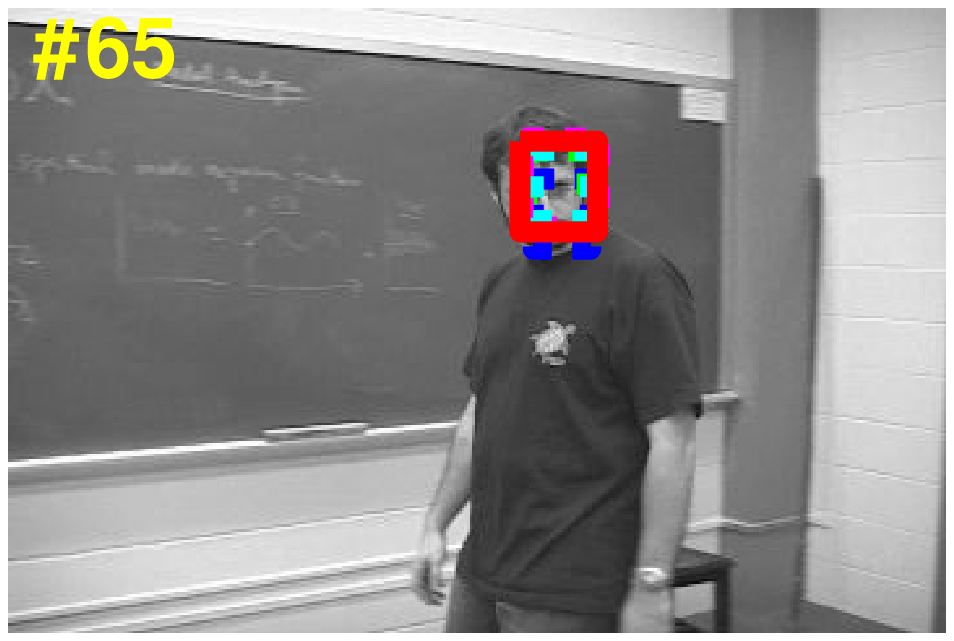,width=0.16\textwidth}
\epsfig{file=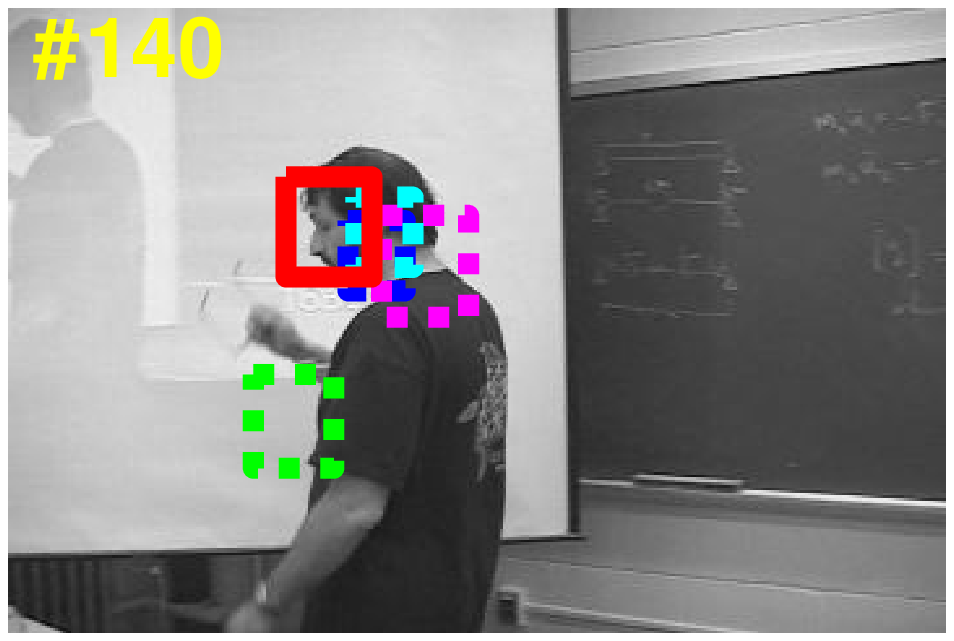,width=0.16\textwidth}
\epsfig{file=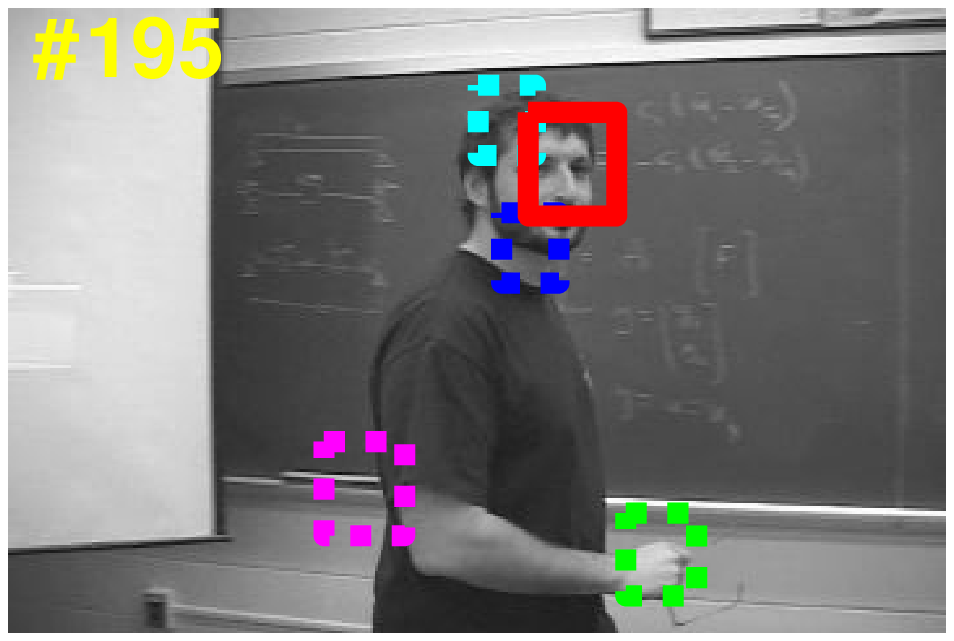,width=0.16\textwidth}
\epsfig{file=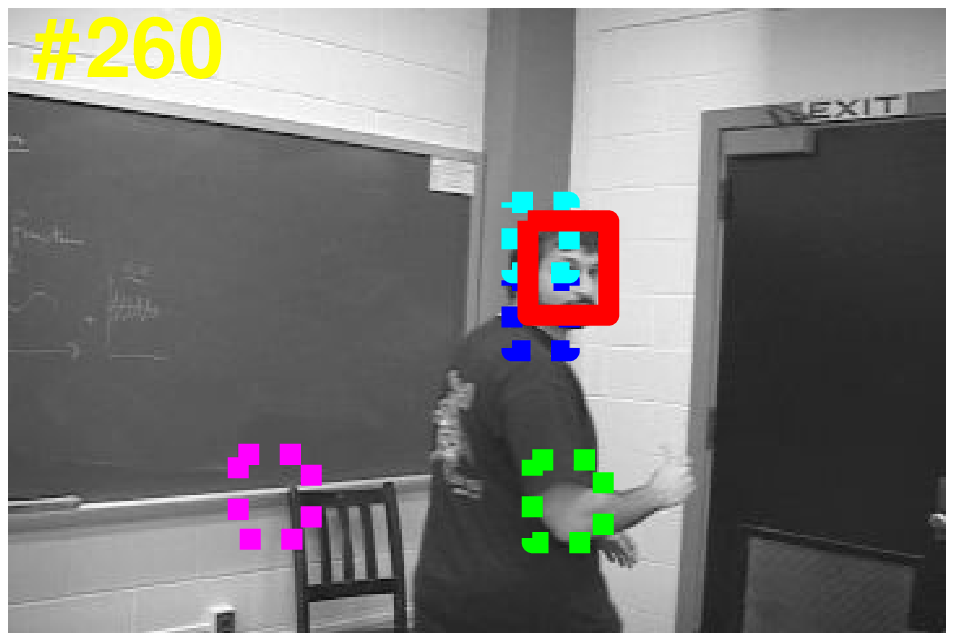,width=0.16\textwidth}
\epsfig{file=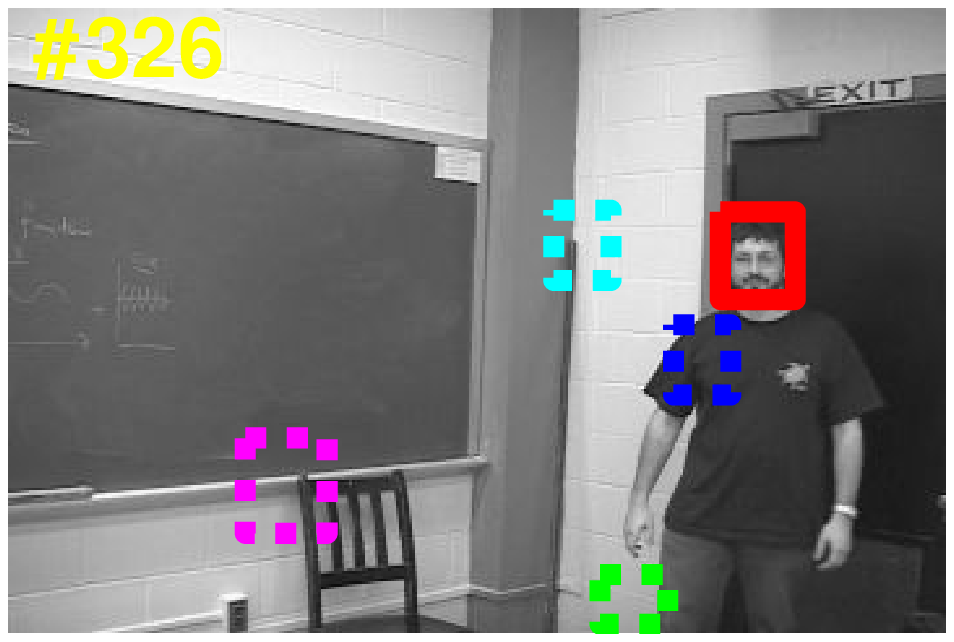,width=0.16\textwidth}}
\\ \vspace{-0.1in}

\centering \subfloat[Freeman3]{
\epsfig{file=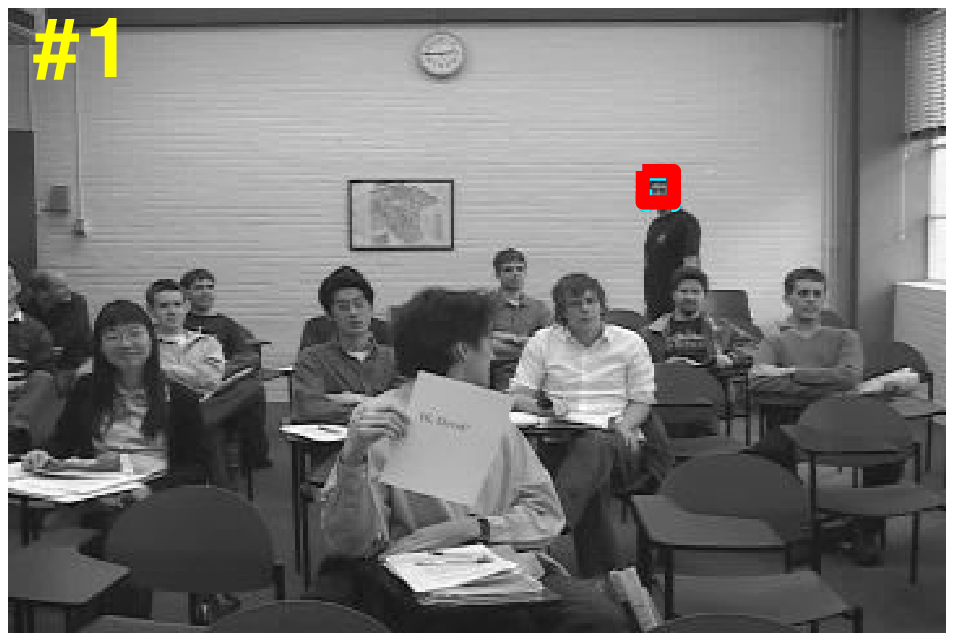,width=0.16\textwidth}
\epsfig{file=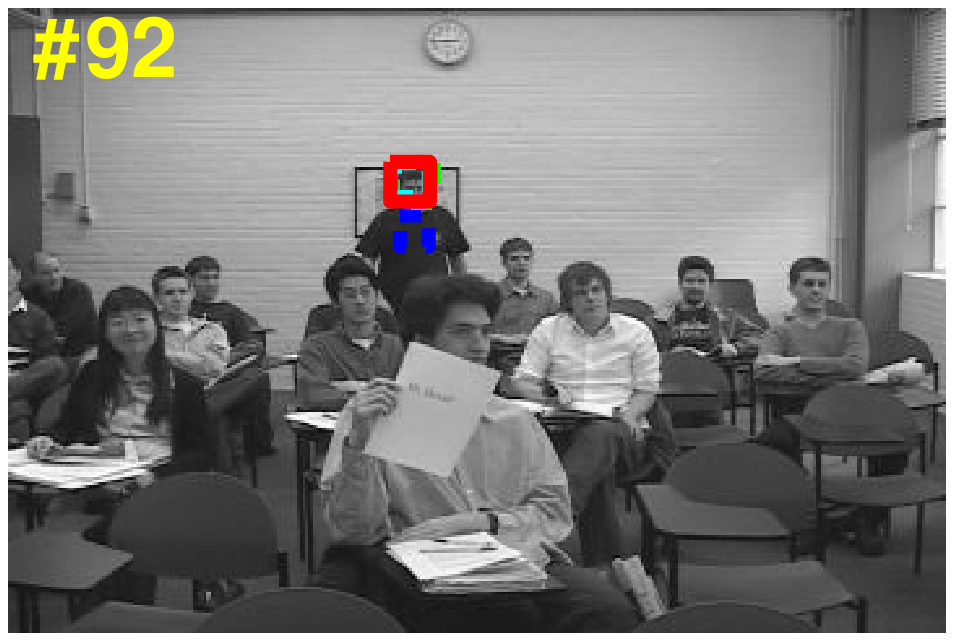,width=0.16\textwidth}
\epsfig{file=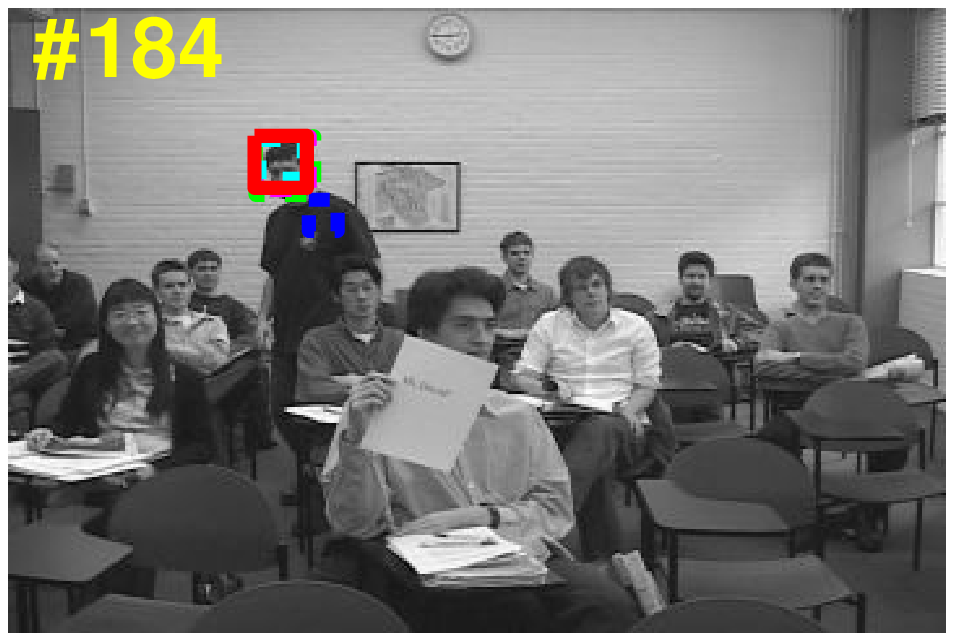,width=0.16\textwidth}
\epsfig{file=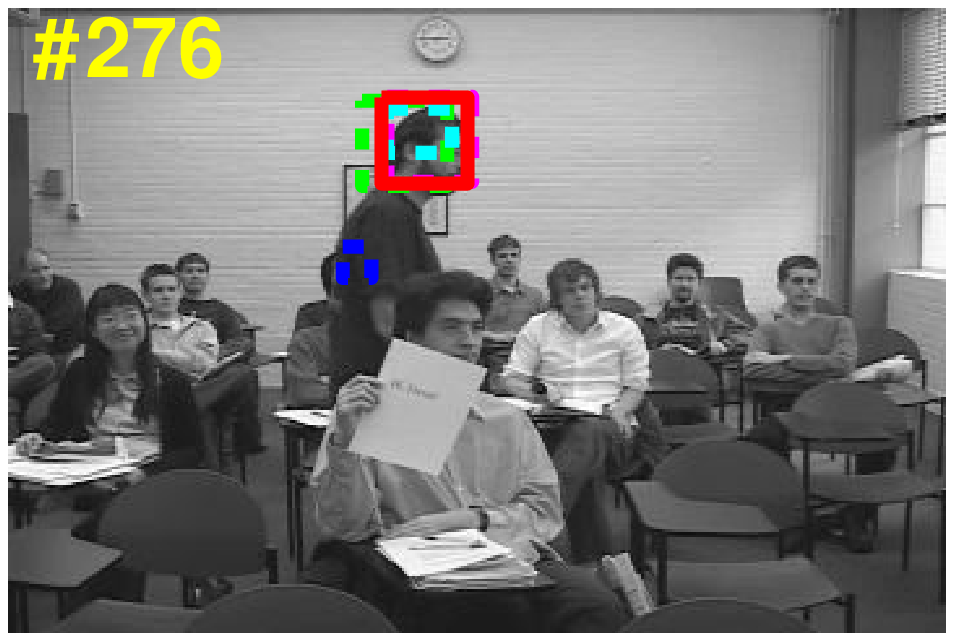,width=0.16\textwidth}
\epsfig{file=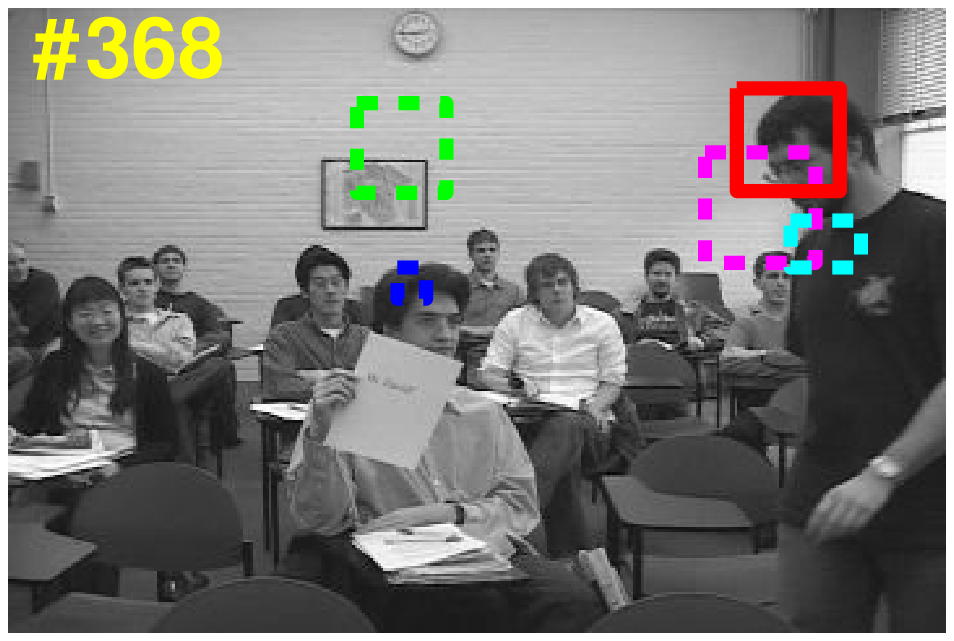,width=0.16\textwidth}
\epsfig{file=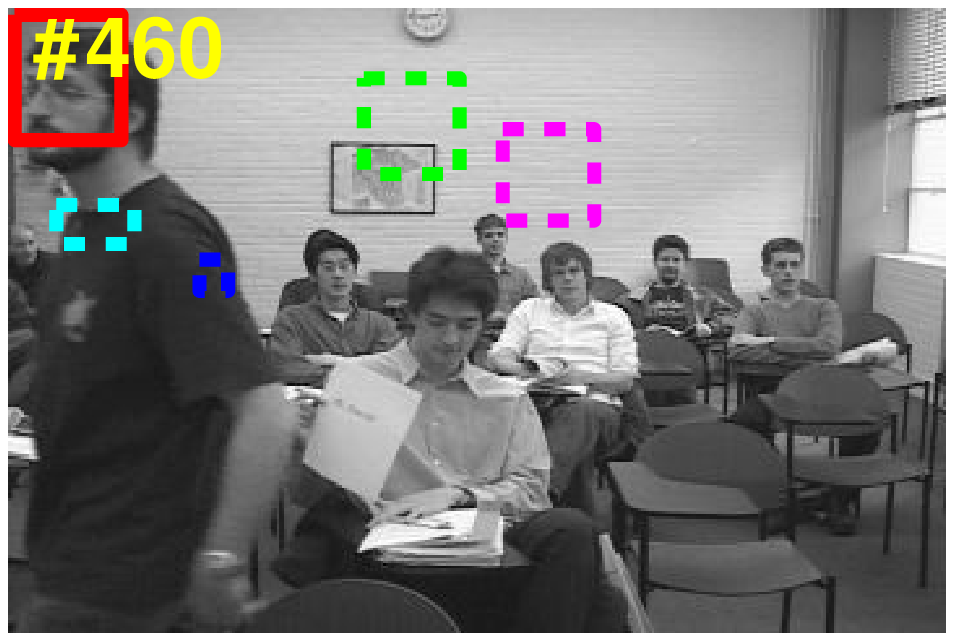,width=0.16\textwidth}}
\\ \vspace{-0.1in}

\centering \subfloat[Lemming]{
\epsfig{file=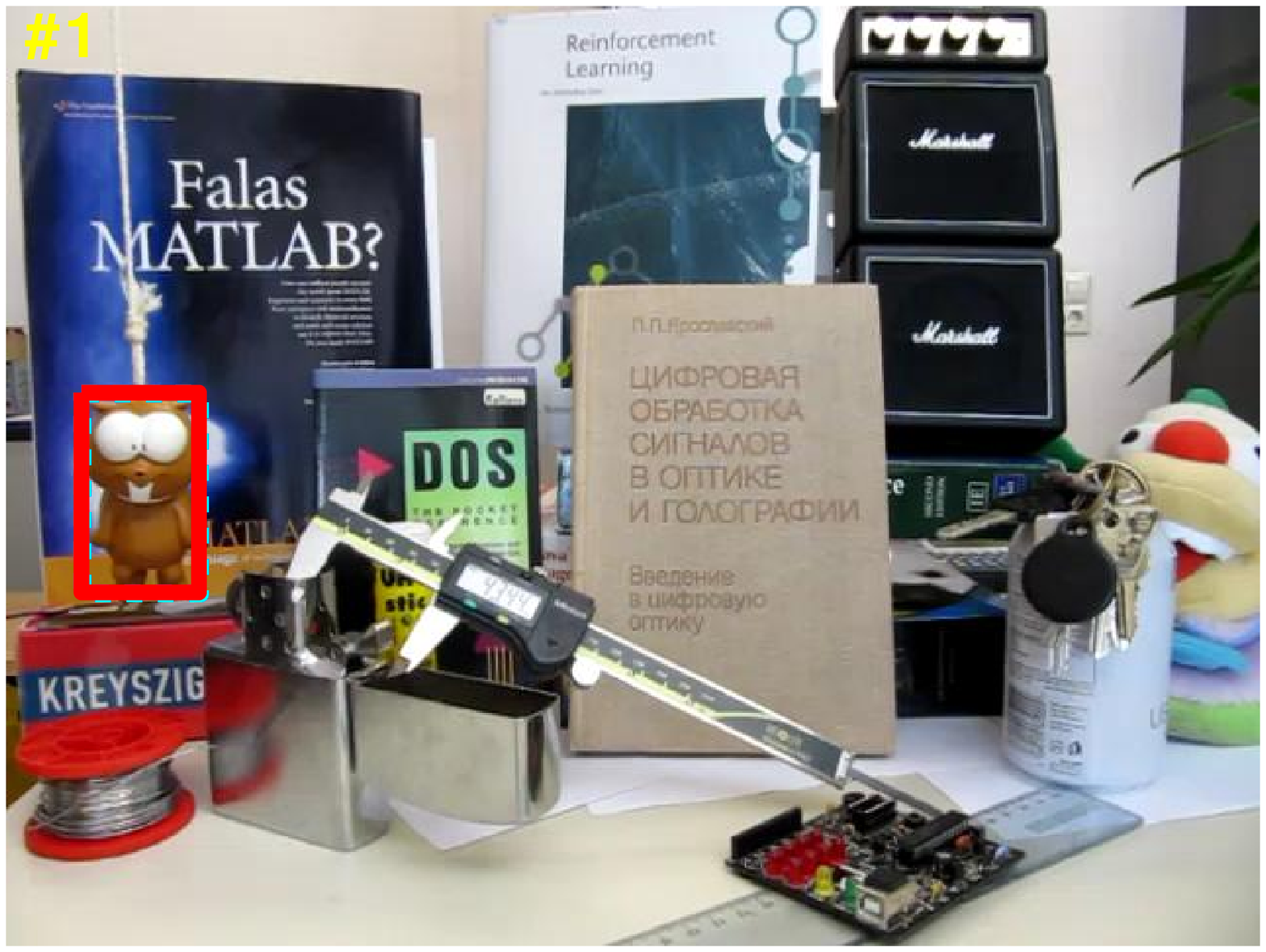,width=0.16\textwidth}
\epsfig{file=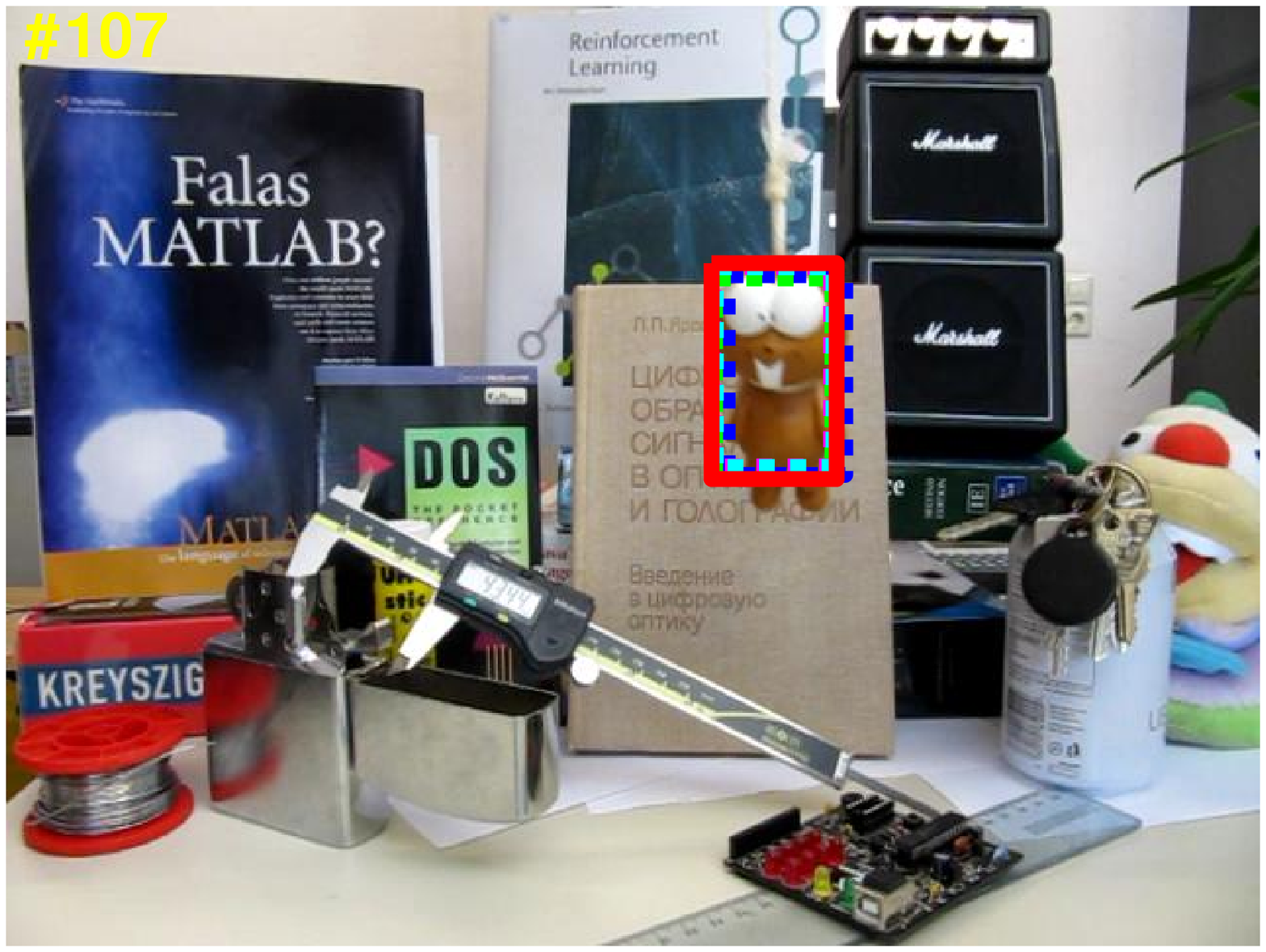,width=0.16\textwidth}
\epsfig{file=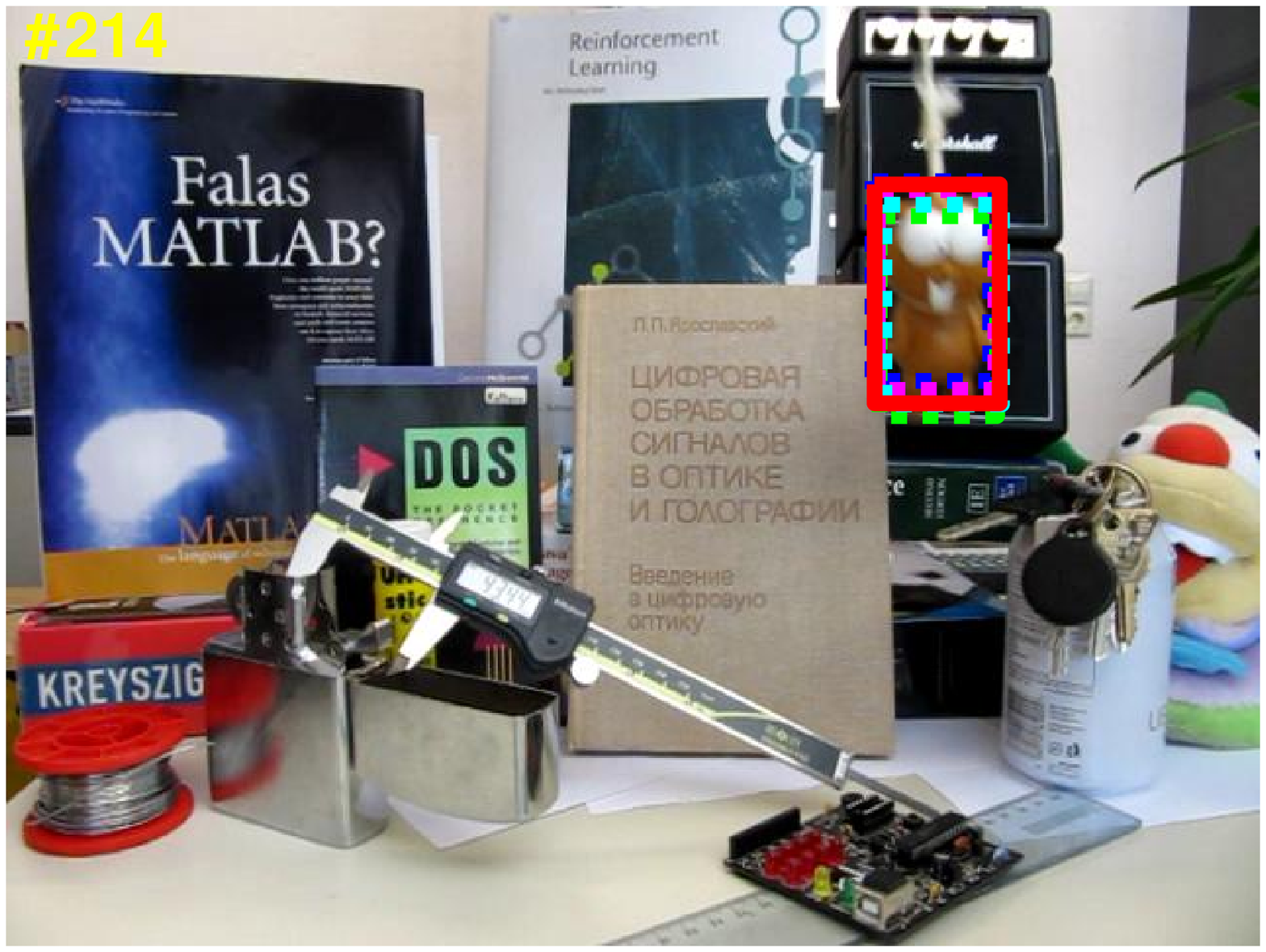,width=0.16\textwidth}
\epsfig{file=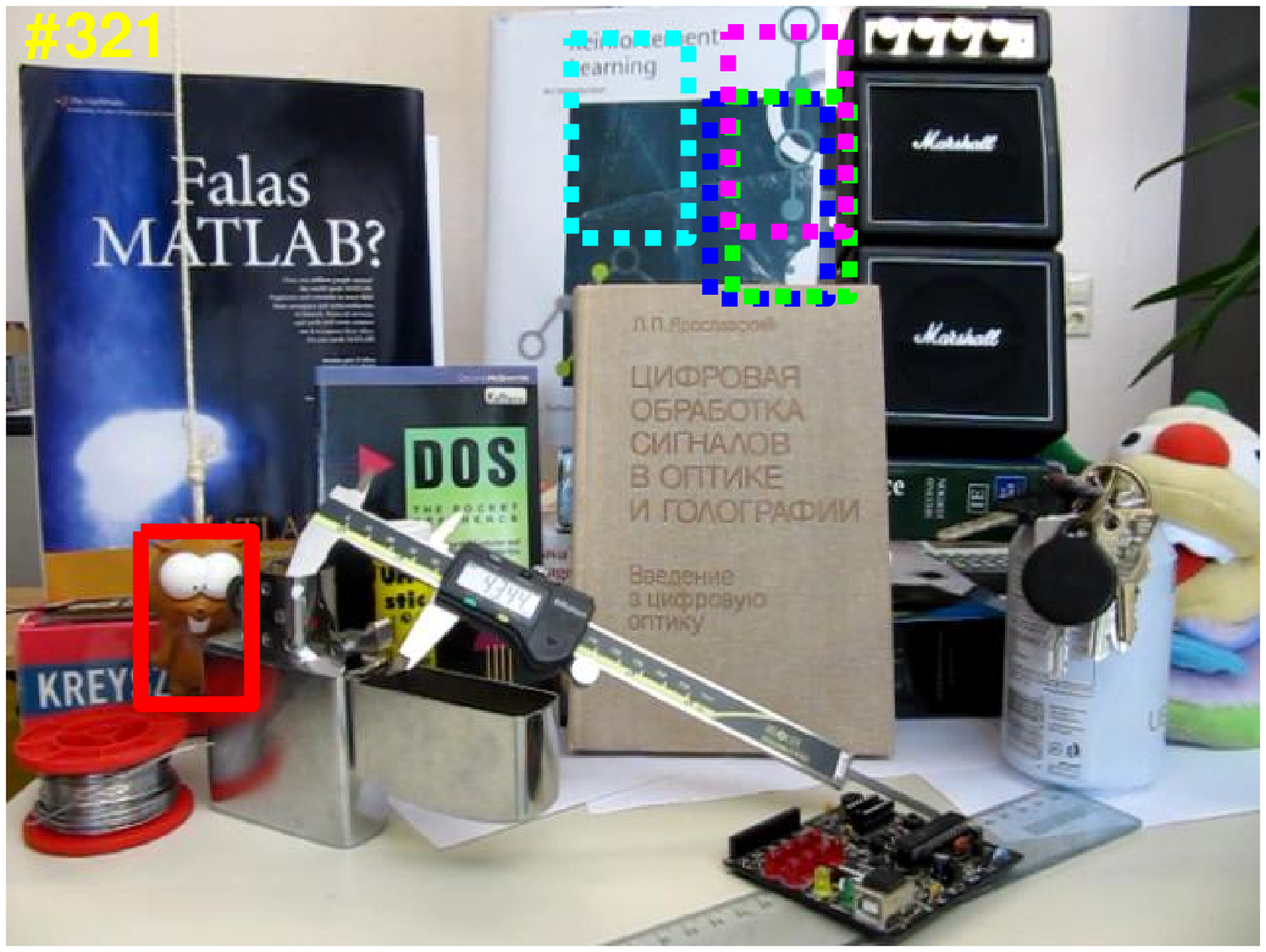,width=0.16\textwidth}
\epsfig{file=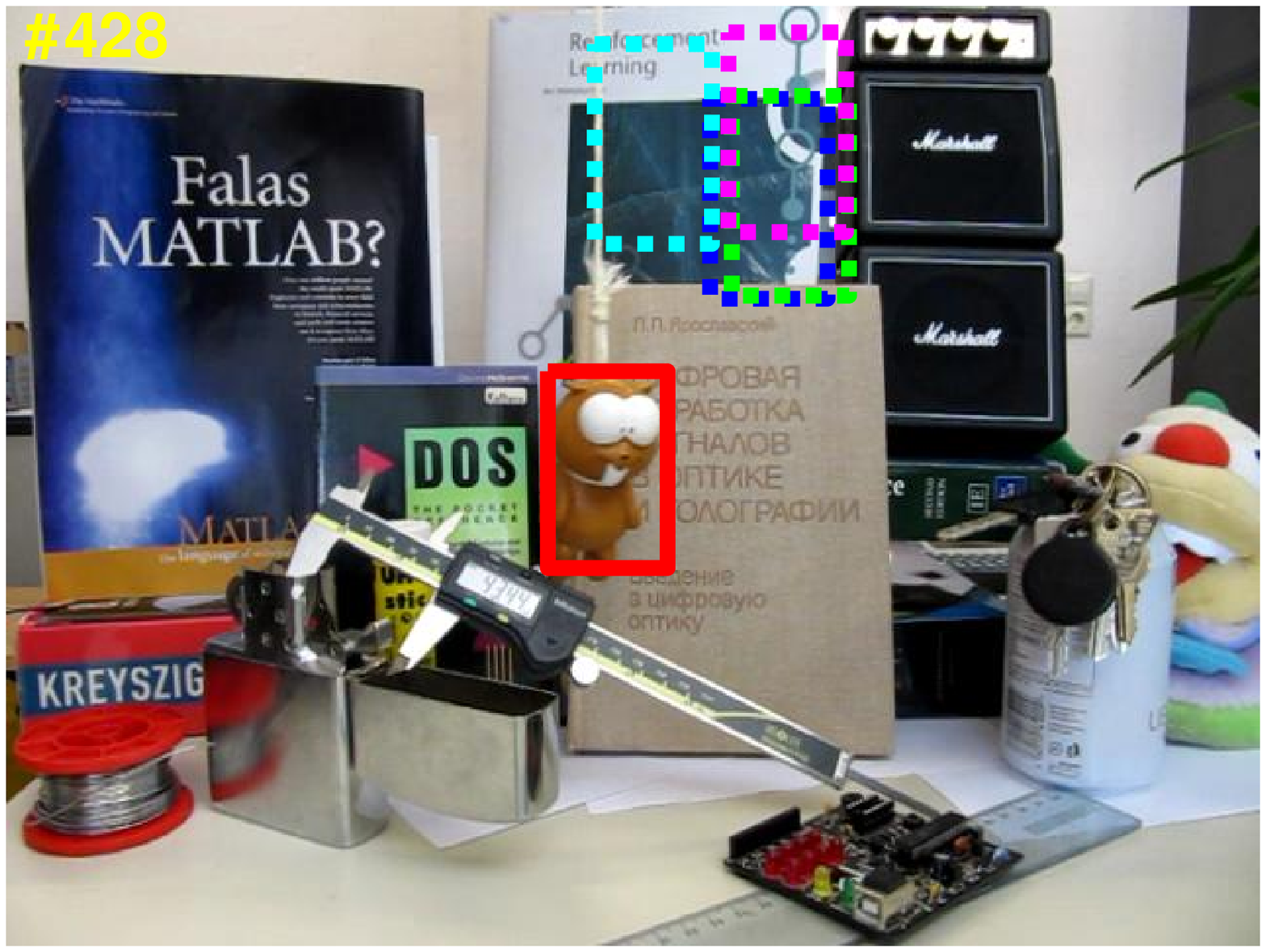,width=0.16\textwidth}
\epsfig{file=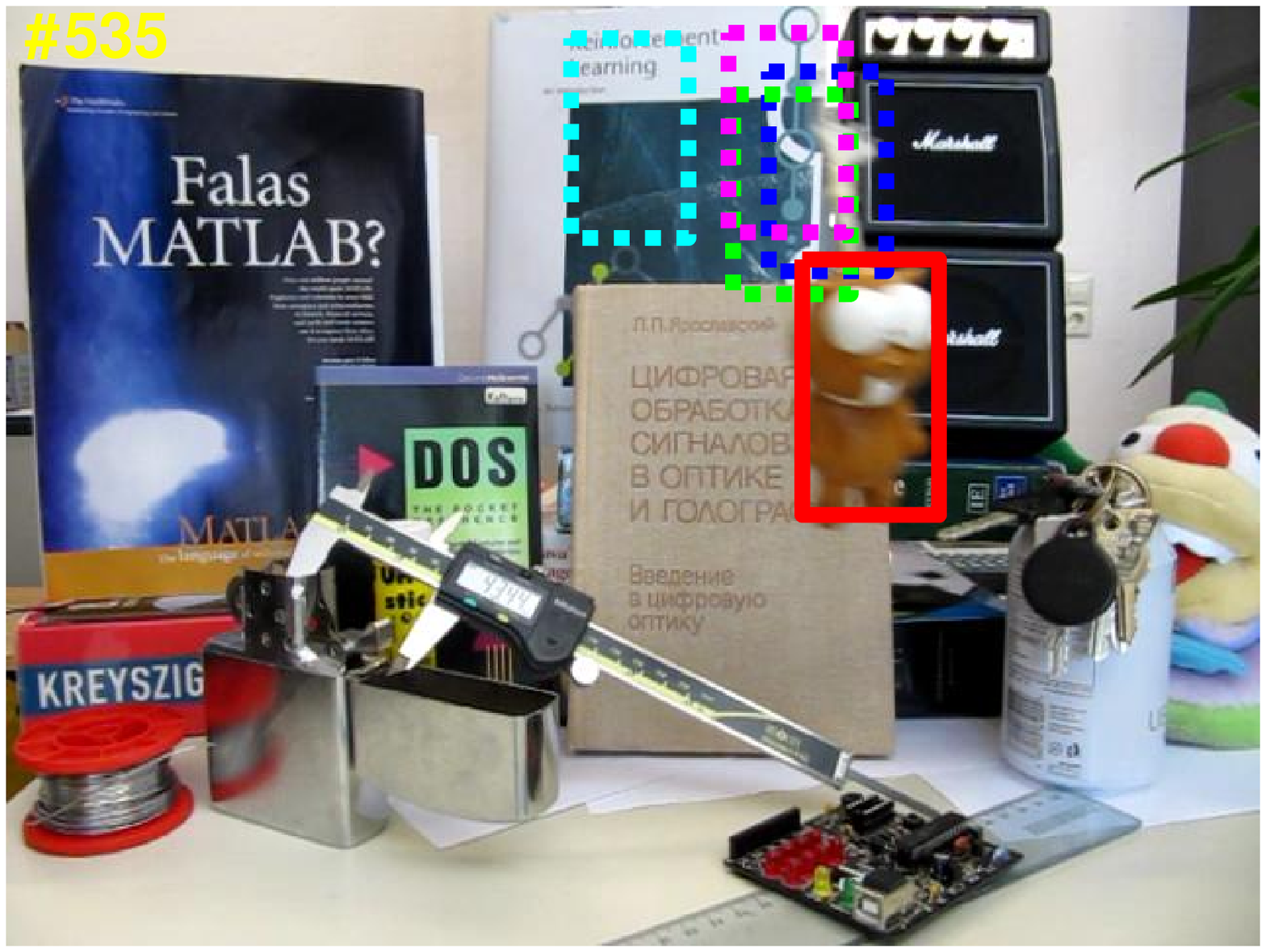,width=0.16\textwidth}}
\\ \vspace{-0.1in}

\centering \subfloat[Shaking]{
\epsfig{file=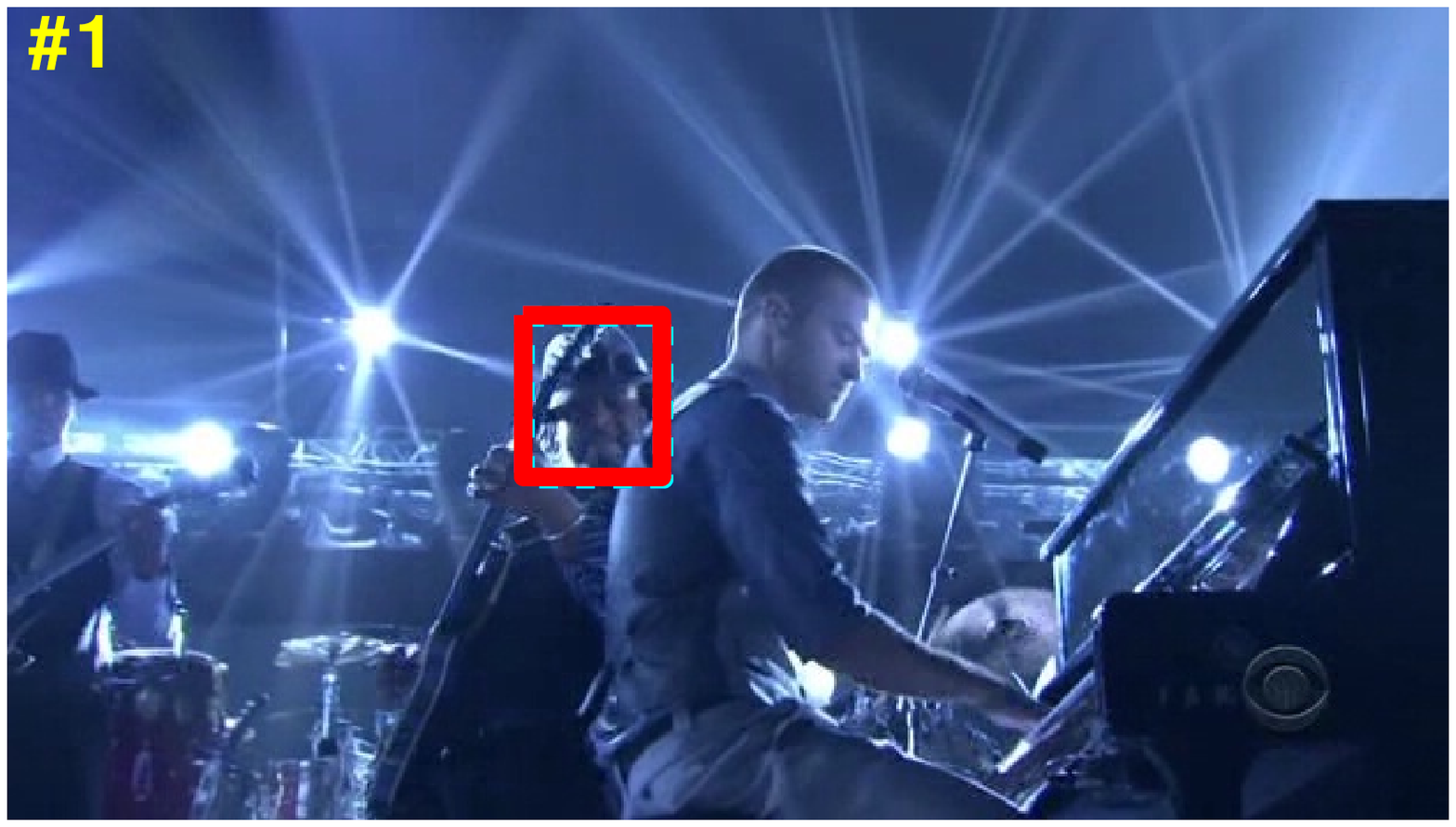,width=0.16\textwidth}
\epsfig{file=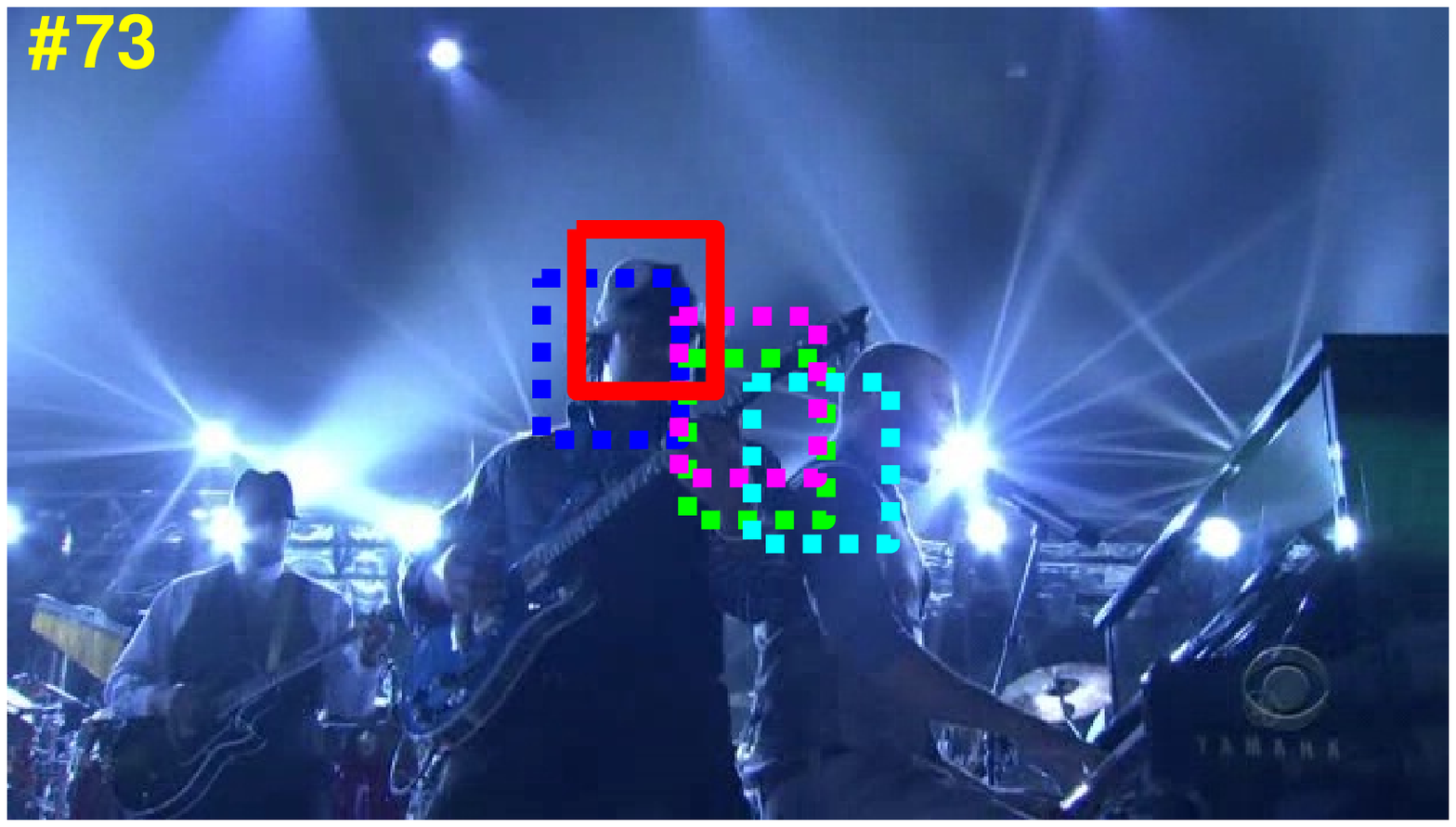,width=0.16\textwidth}
\epsfig{file=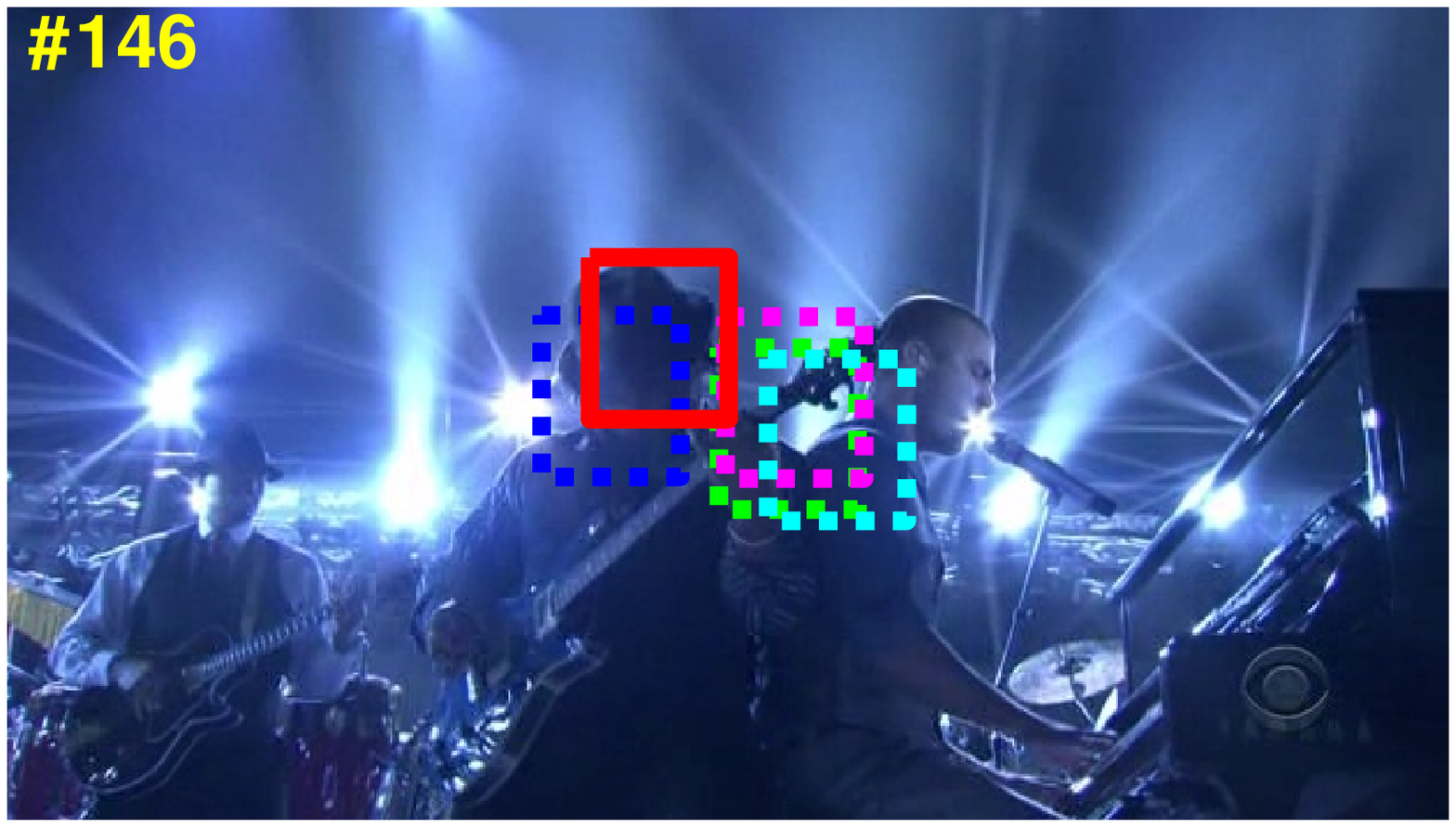,width=0.16\textwidth}
\epsfig{file=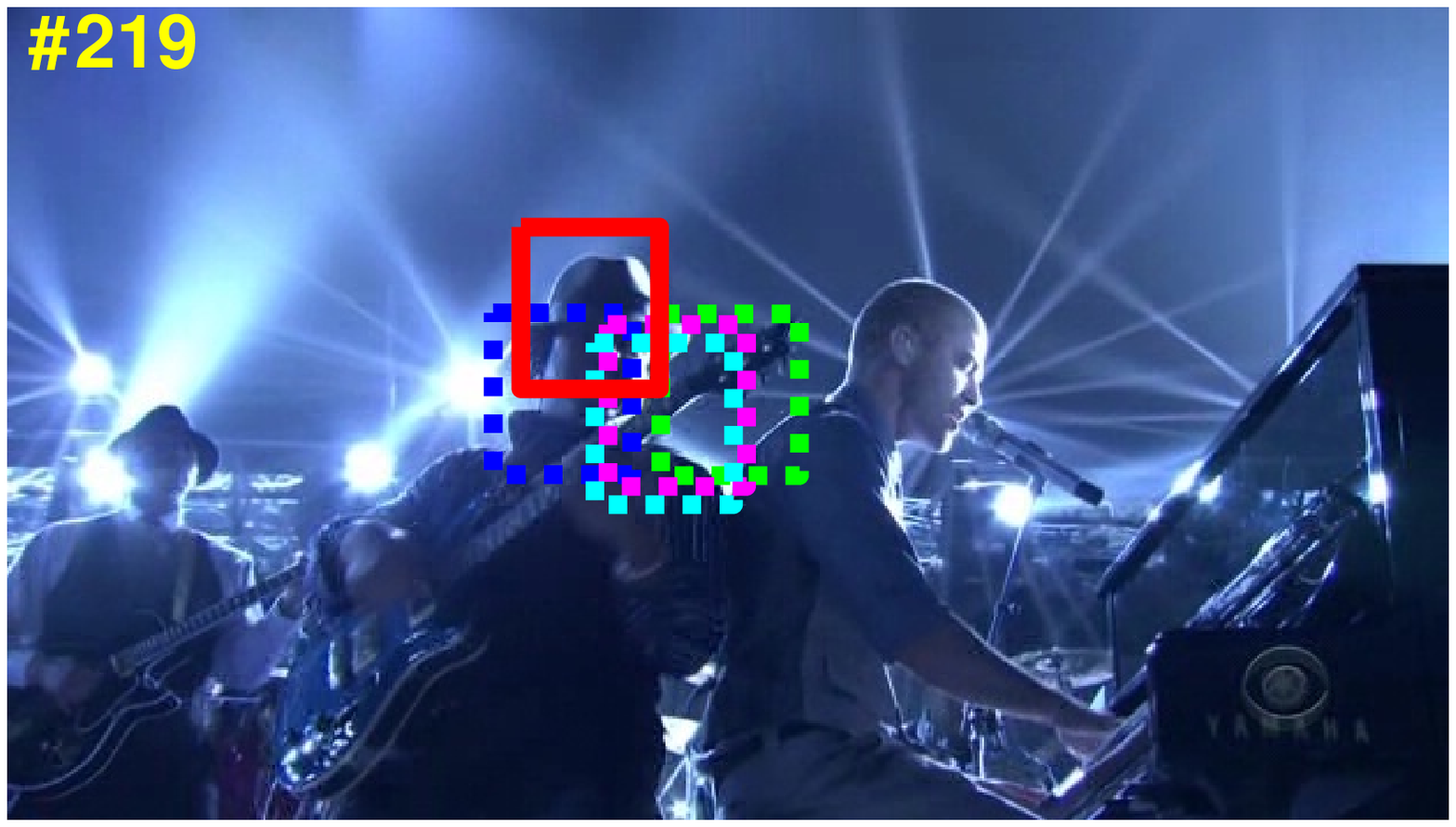,width=0.16\textwidth}
\epsfig{file=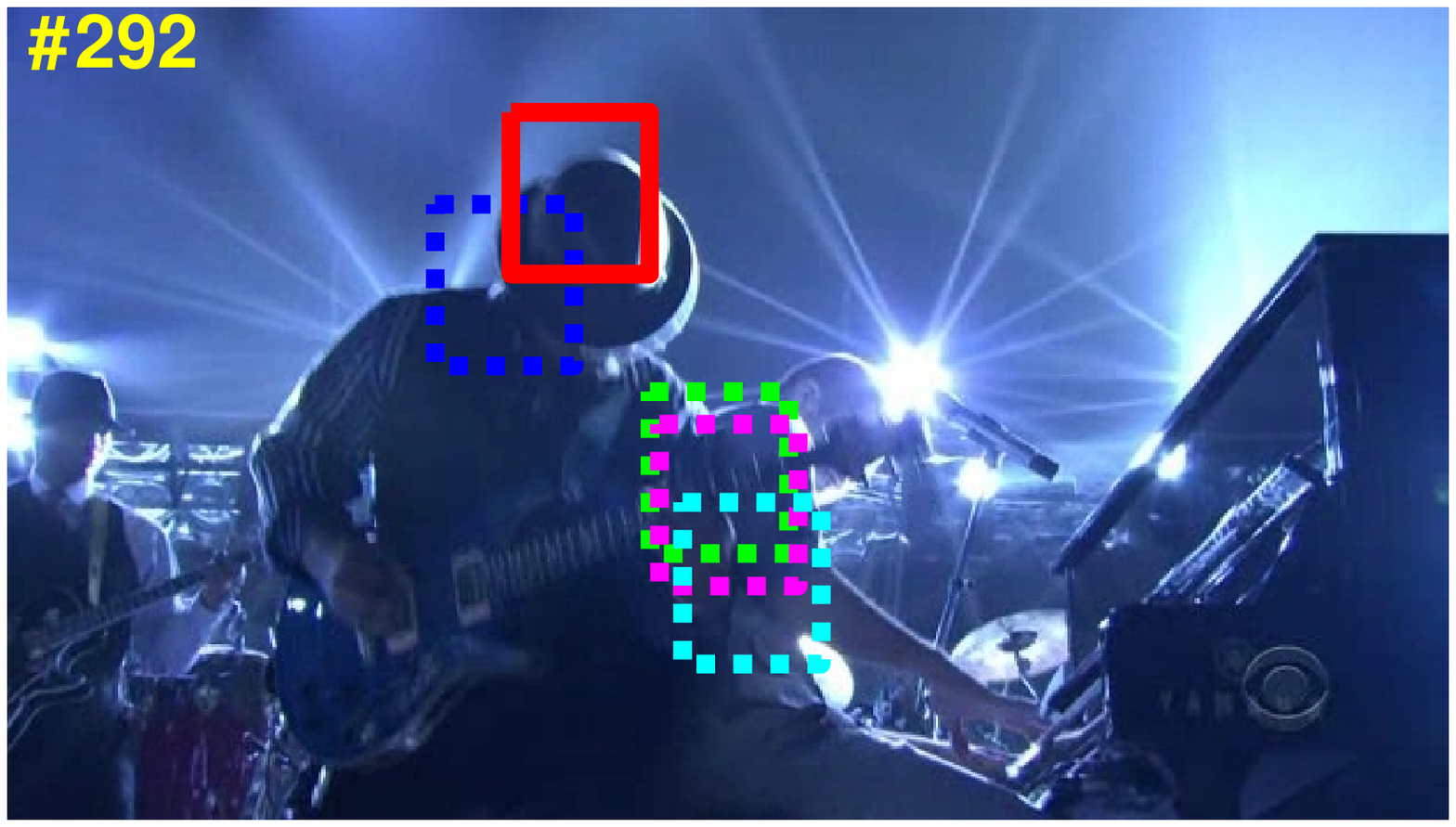,width=0.16\textwidth}
\epsfig{file=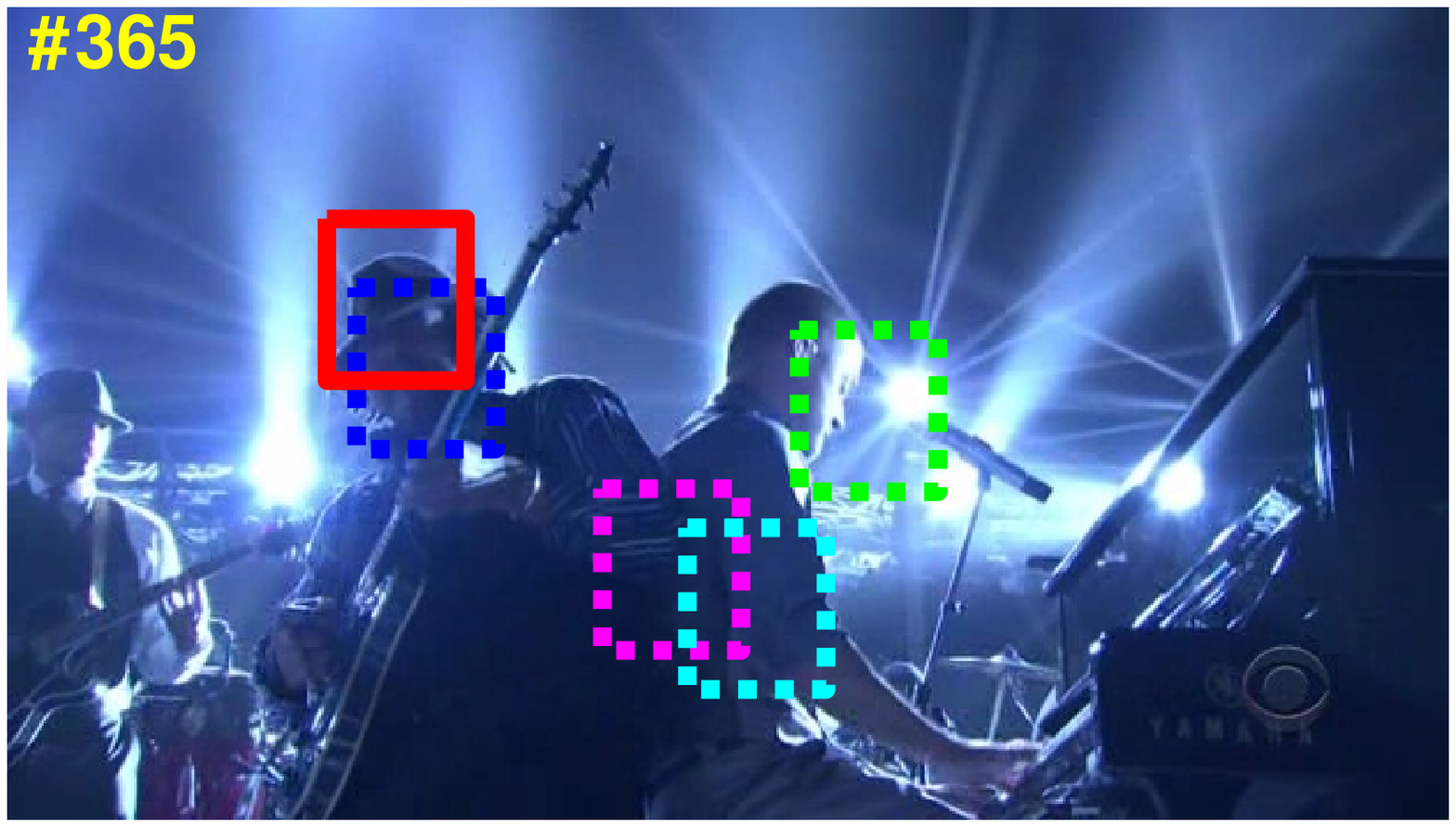,width=0.16\textwidth}}
\\ \vspace{-0.1in}

\centering \subfloat[Trellis]{
\epsfig{file=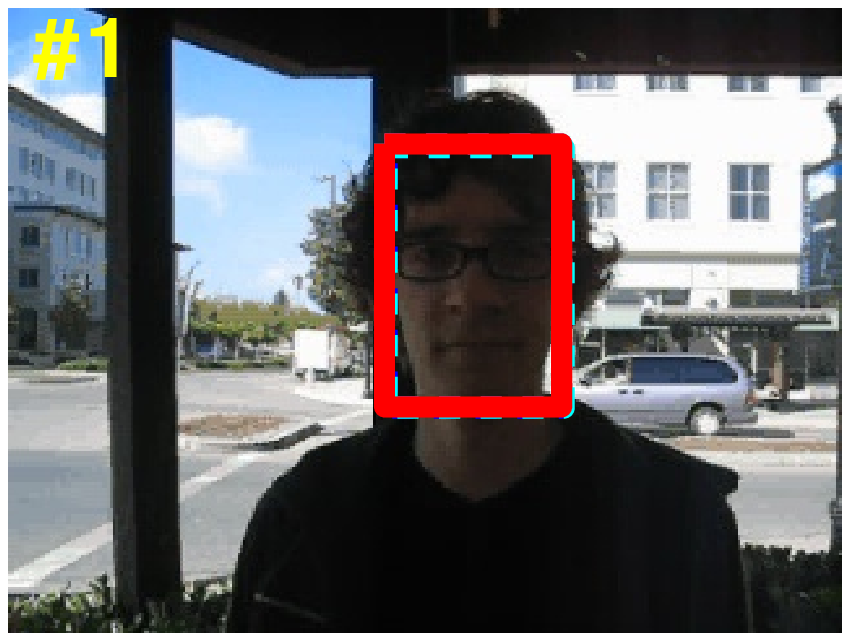,width=0.16\textwidth}
\epsfig{file=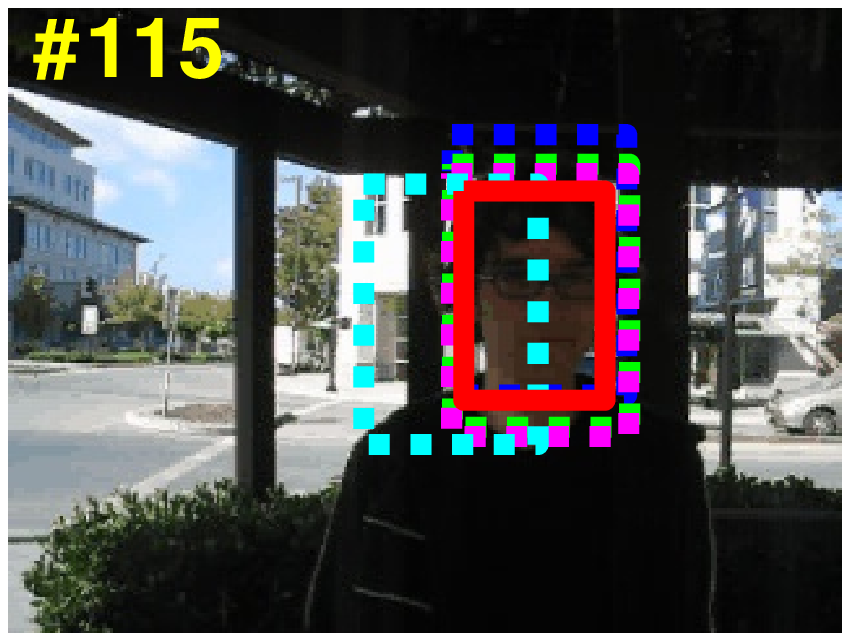,width=0.16\textwidth}
\epsfig{file=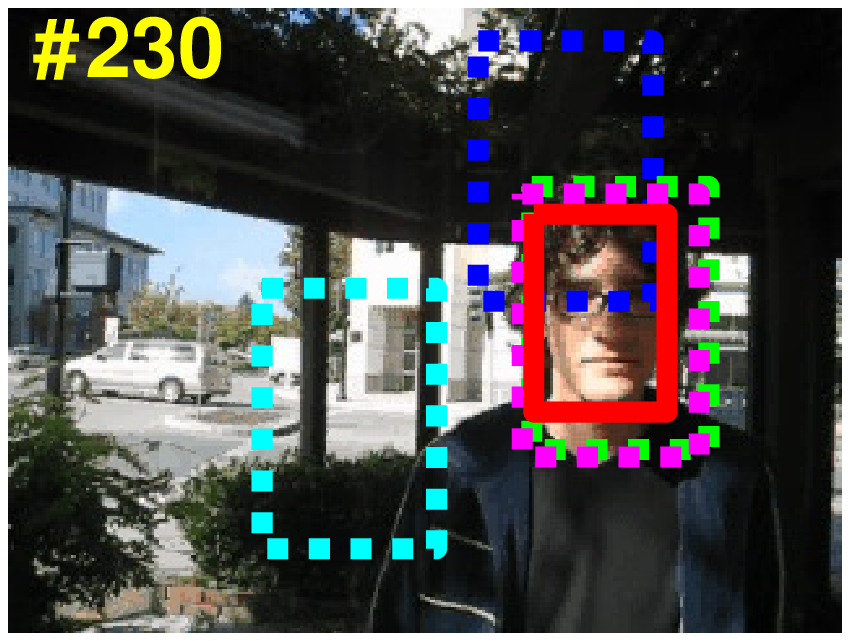,width=0.16\textwidth}
\epsfig{file=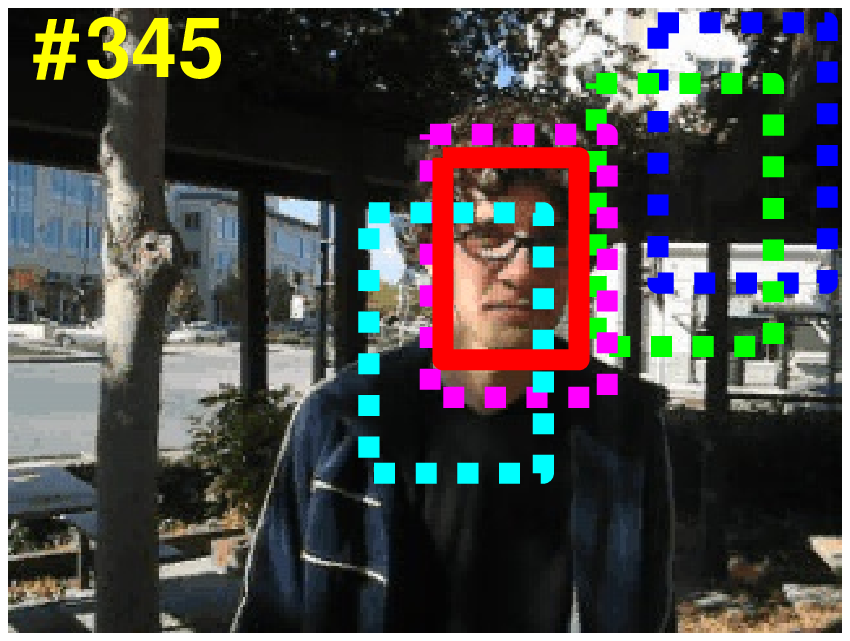,width=0.16\textwidth}
\epsfig{file=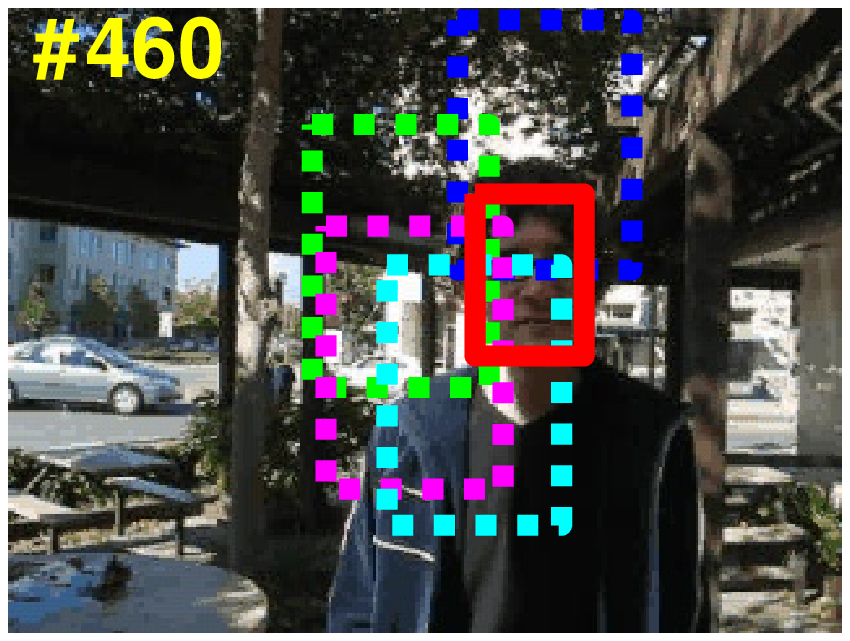,width=0.16\textwidth}
\epsfig{file=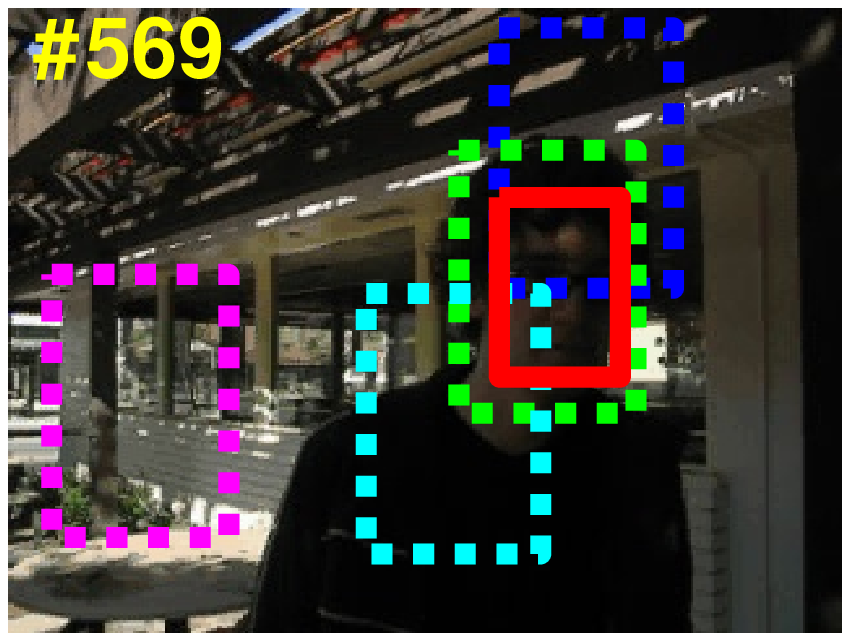,width=0.16\textwidth}}

\caption{Qualitative results on sequences with out-of-plane rotations. The purple, green, cyan, blue and red bounding boxes refer to ASLA\cite{DBLP:conf/cvpr/JiaLY12}\_RAW, ASLA\cite{DBLP:conf/cvpr/JiaLY12}\_HOG, $\ell_1$\_APG \cite{DBLP:conf/cvpr/BaoWLJ12}, CT\_DIF \cite{DBLP:conf/eccv/Zhang0Y12} and our tracker respectively. This figure is better viewed in color.} \label{fig:quali_outOfPlaneRotation}
\end{figure*}

In this evaluation, we test on $13$ sequences used in \cite{DBLP:conf/cvpr/WuLY13}. Also, we have two special sequences of ``biker" and ``kitesurf", in which the original video sequences are used, but new target objects are defined for tracking. Our sequences are challenging because the newly defined objects contain complicated motion transformations. For example, in the sequence of ``biker" (see Figure~\ref{fig:quali_deformation}), we track the biker's whole body which has non-rigid object deformation. Tables \ref{tab:ace_motion} and \ref{tab:aor_motion} present quantitative results which demonstrate that our learned features outperform the other state-of-the-art feature representations in terms of handling complicated motion transformations well. Figures \ref{fig:quali_deformation}, \ref{fig:quali_inPlaneRotation} and \ref{fig:quali_outOfPlaneRotation} show the qualitative results on sequences with non-rigid object deformation, in-plane rotations and out-of-plane rotations respectively. Then, we explain the qualitative results as follows.

\emph{Non-rigid object deformation} The sequences (Basketball, Biker, FleetFace, Kitesurf and Skating1) shown in Figure~\ref{fig:quali_deformation} are challenging because the target objects have non-rigid object deformations. For example, the basketball player in Figure~\ref{fig:quali_deformation} (a) has deformable changes due to his running and defending actions. The biker in Figure~\ref{fig:quali_deformation} (b) has dramatic body deformations during his acrobatic actions. The man in Figure~\ref{fig:quali_deformation} (c) has significant facial changes due to his laughing expression. The person in Figure~\ref{fig:quali_deformation} (d) has deformable pose changes because of his surfing actions. The girl in Figure~\ref{fig:quali_deformation} (e) has articulated deformations caused by her arm waving and body spinning. We can observe that the $4$ baseline trackers (ASLA\cite{DBLP:conf/cvpr/JiaLY12}\_RAW, ASLA\cite{DBLP:conf/cvpr/JiaLY12}\_HOG, $\ell_1$\_APG \cite{DBLP:conf/cvpr/BaoWLJ12} and CT\_DIF \cite{DBLP:conf/eccv/Zhang0Y12}) fail to track the target objects in these challenging sequences. In contrast, our tracker succeeds to capture the target objects because our features are learned to be invariant to non-rigid object deformations.

\emph{In-plane rotations} The target objects in the sequences (David2, MountainBike, Sylvester, Tiger1 and Tiger2) have significant in-plane rotations which are difficult for trackers to capture. In Figure~\ref{fig:quali_inPlaneRotation} (a), the man's face not only has translations but also in-plane rotations which occur when the face is slanted. In Figure~\ref{fig:quali_inPlaneRotation} (b), the mountain bike has the in-plane rotations due to its acrobatic actions in the sky. In Figure~\ref{fig:quali_inPlaneRotation} (c), (d) and (e), the toys have a lot of in-plane rotations. We can see that all the baseline trackers have drifted away from the target objects in these sequences because of in-plane rotations, whereas our tracker can handle this kind of motion transformations effectively by using learned features.

\emph{Out-of-plane rotations} The sequences (Freeman1, Freeman3, Lemming, Shaking and Trellis) are difficult because the target objects have out-of-plane rotations which change object appearances significantly and hence yield tracking failures. For instance, in Figure~\ref{fig:quali_outOfPlaneRotation} (a), (b) and (e), the men' faces have significant out-of-plane rotations because the poses of their heads change a lot during walking. The toy in Figure~\ref{fig:quali_outOfPlaneRotation} (c) has out-of-plane rotations because it rotates along its vertical axis. The singer's head shown in Figure~\ref{fig:quali_outOfPlaneRotation} (d) has out-of-plane rotations because the head shakes up and down. We can observe that our tracker can successfully capture the target objects through these sequences. We owe this success to our learned feature's robustness to out-of-plane rotations. In contrast, the baseline trackers cannot handle this complicated motion transformation because their feature representations are not designed to capture motion invariance.

\subsection{Evaluation on the Temporal Slowness Constraint and the Adaptation Module in Our Feature Learning Algorithm}
First, we present the results of the variant of our tracker (Ours\_VAR) which does not use the temporal slowness constraint in feature learning in Tables \ref{tab:ace_motion} and \ref{tab:aor_motion}. We can observe that our tracker using the constraint has better performances on 15 challenging video sequences. It demonstrates that the temporal slowness constraint is beneficial for learning features robust to complicated motion transformations. Then, we evaluate the adaptation module in our feature learning method on $8$ video sequences reported in ASLA \cite{DBLP:conf/cvpr/JiaLY12}. Tables \ref{tab:ace_jia_hog_adp} and \ref{tab:asr_jia_hog_adp} respectively present the average center location errors and the average overlap rates of our tracker with (Ours\_adp) and without (Ours\_noadp) the adaptation module. From the quantitative comparison, we can find that the adaptation module enhances the performance of our tracker. It is due to the fact that the adaptation module not only preserves the pre-learned features' robustness to complicated motion transformations, but also includes appearance information of specific target objects.

\begin{table}
\caption{Average center error (in pixels). The best two results are shown in red and blue fonts. We present our tracker's performances with (Ours\_adp) and without (Ours\_noadp) the adaptation module. We also compare our tracker with $4$ baseline trackers using other features such as the raw pixel values (ASLA\cite{DBLP:conf/cvpr/JiaLY12}\_RAW), the hand-crafted HOG feature (ASLA\cite{DBLP:conf/cvpr/JiaLY12}\_HOG), the sparse feature ($\ell_1$\_APG \cite{DBLP:conf/cvpr/BaoWLJ12}) and the data-independent feature (CT\_DIF\cite{DBLP:conf/eccv/Zhang0Y12}).} \vspace{-0.1in}
\begin{center}
\tabcolsep 0.038in \scriptsize
\begin{tabular}{|c||c|c|c|c||c|c|}
\hline
\textbf{Sequence}&\tiny{\textbf{ASLA}\cite{DBLP:conf/cvpr/JiaLY12}\textbf{\_RAW}}&\tiny{\textbf{ASLA}\cite{DBLP:conf/cvpr/JiaLY12}\textbf{\_HOG}}&\tiny{\textbf{$\ell_1$\_APG}\cite{DBLP:conf/cvpr/BaoWLJ12}}&\tiny{\textbf{CT\_DIF}\cite{DBLP:conf/eccv/Zhang0Y12}}&\tiny{\textbf{Ours\_noadp}}&\tiny{\textbf{Ours\_adp}}\\
\hline \hline
Board     & \textcolor{blue}{7.3} &15.3 & 259.4&80.3&\textcolor{red}{7.1} &8.1\\
Car11     & 2.0 &2.7 & 22.2&78.0&\textcolor{blue}{1.9} &\textcolor{red}{1.4}\\
Caviar    & \textcolor{blue}{2.3} &66.8 & 95.6&65.5&\textcolor{red}{2.2} &\textcolor{red}{2.2}\\
David     & 3.6 &45.8 & 138.2&12.8&\textcolor{blue}{3.5} &\textcolor{red}{3.2}\\
Faceocc2  & 3.8 &32.4 & 17.7&12.8&\textcolor{blue}{3.7} &\textcolor{red}{3.1}\\
Singer1   & 4.8 &5.0 & 167.9&13.7&\textcolor{blue}{4.5} &\textcolor{red}{4.0}\\
Stone     & \textcolor{blue}{1.8} &2.8 & 136.8&32.4& 2.1 &\textcolor{red}{1.5}\\
Woman     & 2.8 &140.6 & 176.1&110.2&\textcolor{blue}{2.6} &\textcolor{red}{2.2}\\
\hline
\end{tabular}
\end{center}
\label{tab:ace_jia_hog_adp}
\end{table}

\begin{table}
\caption{Average overlap rates. (\%) The best two results are shown in red and blue fonts. We present our tracker's performances with (Ours\_adp) and without (Ours\_noadp) the adaptation module. We compare our tracker with $4$ baseline trackers using other features such as the raw pixel values (ASLA\cite{DBLP:conf/cvpr/JiaLY12}\_RAW), the hand-crafted HOG feature (ASLA\cite{DBLP:conf/cvpr/JiaLY12}\_HOG), the sparse feature ($\ell_1$\_APG \cite{DBLP:conf/cvpr/BaoWLJ12}) and the data-independent feature (CT\_DIF\cite{DBLP:conf/eccv/Zhang0Y12}).} \vspace{-0.1in}
\begin{center}
\tabcolsep 0.038in \scriptsize
\begin{tabular}{|c||c|c|c|c||c|c|}
\hline
\textbf{Sequence}&\tiny{\textbf{ASLA}\cite{DBLP:conf/cvpr/JiaLY12}\textbf{\_RAW}}&\tiny{\textbf{ASLA}\cite{DBLP:conf/cvpr/JiaLY12}\textbf{\_HOG}}&\tiny{\textbf{$\ell_1$\_APG}\cite{DBLP:conf/cvpr/BaoWLJ12}}&\tiny{\textbf{CT\_DIF}\cite{DBLP:conf/eccv/Zhang0Y12}}&\tiny{\textbf{Ours\_noadp}}&\tiny{\textbf{Ours\_adp}}\\
\hline \hline
Board     & \textcolor{blue}{0.74} &0.72 & 0.12&0.33& 0.73 &\textcolor{red}{0.83}\\
Car11     & \textcolor{blue}{0.81} &0.71 & 0.34&0.23& 0.80 &\textcolor{red}{0.85}\\
Caviar    & 0.84 &0.40 & 0.06&0.33&\textcolor{blue}{0.87} &\textcolor{red}{0.88}\\
David     & 0.79 &0.27 & 0.19&0.56&\textcolor{blue}{0.80} &\textcolor{red}{0.81}\\
Faceocc2  & 0.82 &0.52 & 0.61&0.68&\textcolor{blue}{0.87} &\textcolor{red}{0.88}\\
Singer1   & 0.81 &0.79 & 0.14&0.34&\textcolor{blue}{0.82} &\textcolor{red}{0.85}\\
Stone     & 0.56 &0.57 & 0.16&0.33&\textcolor{blue}{0.59} &\textcolor{red}{0.60}\\
Woman     & 0.78 &0.35 & 0.20&0.41&\textcolor{blue}{0.81} &\textcolor{red}{0.84}\\
\hline
\end{tabular}
\end{center}
\label{tab:asr_jia_hog_adp}
\end{table}

\subsection{Evaluation on Our Tracker's Capability of Handling Typical Problems in Visual Tracking}
We use the $8$ sequences in ASLA \cite{DBLP:conf/cvpr/JiaLY12} to evaluate our tracker's capability of handling typical problems in visual tracking, \eg illumination change, occlusion and cluttered background. We quantitatively compare our tracker with $4$ baseline trackers, ASLA\cite{DBLP:conf/cvpr/JiaLY12}\_RAW, ASLA\cite{DBLP:conf/cvpr/JiaLY12}\_HOG, $\ell_1$\_APG \cite{DBLP:conf/cvpr/BaoWLJ12} and CT\_DIF \cite{DBLP:conf/eccv/Zhang0Y12}, which use the raw pixel values, the hand-crafted HOG feature, the sparse representation and the data-independent feature respectively. From Tables \ref{tab:ace_jia_hog_adp} and \ref{tab:asr_jia_hog_adp}, we can find that our learned features are more competitive than the other $4$ feature representations for handling typical issues in visual tracking.

\begin{table*}
\caption{Average center error (in pixels). The best two results are shown in red and blue fonts. We compare the proposed tracker with FragT \cite{DBLP:conf/cvpr/AdamRS06}, IVT \cite{DBLP:journals/ijcv/RossLLY08}, $\ell_1$T \cite{DBLP:conf/iccv/MeiL09}, MIL \cite{DBLP:conf/cvpr/BabenkoYB09}, TLD \cite{DBLP:journals/pami/KalalMM12}, VTD \cite{DBLP:conf/cvpr/KwonL10}, LSK \cite{DBLP:conf/cvpr/LiuHYK11}, CT \cite{DBLP:conf/eccv/Zhang0Y12}, ASLA \cite{DBLP:conf/cvpr/JiaLY12} $\ell_1$APG \cite{DBLP:conf/cvpr/BaoWLJ12}, MTT \cite{DBLP:conf/cvpr/ZhangGLA12}, SCM \cite{DBLP:conf/cvpr/ZhongLY12}, OSPT \cite{DBLP:journals/tip/WangLY13} and LSST \cite{DBLP:conf/cvpr/WangLY13}. Our tracker outperforms the state-of-the-art tracking algorithms.} \vspace{-0.1in}
\begin{center}
\tabcolsep 0.08in
\scriptsize
\begin{tabular}{|c||c|c|c|c|c|c|c|c|c|c|c|c|c|c|c|}
\hline
\textbf{Sequence}&\tiny{\textbf{FragT}\cite{DBLP:conf/cvpr/AdamRS06}}&\tiny{\textbf{IVT}\cite{DBLP:journals/ijcv/RossLLY08}}&\tiny{\textbf{$\ell_1$T}\cite{DBLP:conf/iccv/MeiL09}}&\tiny{\textbf{MIL}\cite{DBLP:conf/cvpr/BabenkoYB09}}&\tiny{\textbf{TLD}\cite{DBLP:journals/pami/KalalMM12}}&\tiny{\textbf{VTD}\cite{DBLP:conf/cvpr/KwonL10}}&\tiny{\textbf{LSK}\cite{DBLP:conf/cvpr/LiuHYK11}}&\tiny{\textbf{CT}\cite{DBLP:conf/eccv/Zhang0Y12}}&\tiny{\textbf{ASLA}\cite{DBLP:conf/cvpr/JiaLY12}}&\tiny{\textbf{$\ell_1$APG}\cite{DBLP:conf/cvpr/BaoWLJ12}}&\tiny{\textbf{MTT}\cite{DBLP:conf/cvpr/ZhangGLA12}}&\tiny{\textbf{SCM}\cite{DBLP:conf/cvpr/ZhongLY12}}&\tiny{\textbf{OSPT}\cite{DBLP:journals/tip/WangLY13}}&\tiny{\textbf{LSST}\cite{DBLP:conf/cvpr/WangLY13}}&\tiny{\textbf{Ours}}\\
\hline \hline
Car4&179.8&\textcolor{red}{2.9}&9.0&60.1&18.8&12.3&3.3&229.7&4.3&16.4&37.2&3.5&\textcolor{blue}{3.0}&\textcolor{red}{2.9}&\textcolor{blue}{3.0}\\
Car11&63.9&2.1&33.3&43.5&25.1&27.1&4.1&78.0&2.0&1.7&1.8&1.8&2.2&\textcolor{blue}{1.6}&\textcolor{red}{1.4}\\
Caviar&94.2&66.2&65.9&83.9&53.0&60.9&55.3&65.5&\textcolor{blue}{2.3}&68.6&67.5&\textcolor{red}{2.2}&45.7&3.1&\textcolor{red}{2.2}\\
David&76.7&3.6&7.6&16.1&9.7&13.6&6.3&12.8&3.6&10.8&13.4&\textcolor{blue}{3.4}&\textcolor{red}{3.2}&4.3&\textcolor{red}{3.2}\\
Deer&50.4&127.5&140.5&55.6&25.7&11.9&69.8&10.5&\textcolor{blue}{8.0}&38.4&9.2&36.8&8.5&10.0&\textcolor{red}{5.5}\\
Faceooc1& 4.6 & 16.3 & 6.3 & 19.2 & 17.6 & 11.1 & 5.3 & 19.9 & 10.8 & 6.8 & 14.1 &\textcolor{red}{3.2}& 4.7 & 5.3 &\textcolor{blue}{3.9}\\
Faceocc2&15.5&10.2&11.1&14.1&18.6&10.4&58.6&12.8&\textcolor{blue}{3.8}&6.3&9.2&4.8&4.0&\textcolor{red}{3.1}&\textcolor{red}{3.1}\\
Football&16.9&42.5&48.6&6.6&11.8&\textcolor{red}{4.1}&14.1&11.6&18.0&12.4&\textcolor{blue}{6.5}&10.4&33.7&7.6&\textcolor{blue}{6.5}\\
Jumping& 58.4 & 36.8 & 12.5 & 9.9 &\textcolor{red}{3.6}& 63.0 & 55.2 &53.0& 39.1 & 8.8 & 19.2 &\textcolor{blue}{3.9}& 5.0 & 4.8 &4.9\\
Singer1&22.0 &8.5 &4.6 &15.2 &32.7 &4.1& 14.5 & 13.7 & 4.8 &\textcolor{red}{3.1}& 41.2 & 3.8 & 4.7 &\textcolor{blue}{3.5}&4.0\\
\hline
Average&58.2&31.7&33.9&32.4&21.7&21.9&28.7&50.8&9.7&17.3&21.9&7.4&11.5&\textcolor{blue}{4.6}&\textcolor{red}{3.8}\\
\hline
\end{tabular}
\end{center}
\label{tab:ace_all}
\end{table*}

\begin{table*}
\caption{Average overlap rate (\%). The best two results are shown in red and blue fonts. We compare the proposed tracker with FragT \cite{DBLP:conf/cvpr/AdamRS06}, IVT \cite{DBLP:journals/ijcv/RossLLY08}, $\ell_1$T \cite{DBLP:conf/iccv/MeiL09}, MIL \cite{DBLP:conf/cvpr/BabenkoYB09}, TLD \cite{DBLP:journals/pami/KalalMM12}, VTD \cite{DBLP:conf/cvpr/KwonL10}, LSK \cite{DBLP:conf/cvpr/LiuHYK11}, CT \cite{DBLP:conf/eccv/Zhang0Y12}, ASLA \cite{DBLP:conf/cvpr/JiaLY12} $\ell_1$APG \cite{DBLP:conf/cvpr/BaoWLJ12}, MTT \cite{DBLP:conf/cvpr/ZhangGLA12}, SCM \cite{DBLP:conf/cvpr/ZhongLY12}, OSPT \cite{DBLP:journals/tip/WangLY13} and LSST \cite{DBLP:conf/cvpr/WangLY13}. Our tracker outperforms the state-of-the-art tracking algorithms.} \vspace{-0.1in}
\begin{center}
\tabcolsep 0.08in
\scriptsize
\begin{tabular}{|c||c|c|c|c|c|c|c|c|c|c|c|c|c|c|c|}
\hline
\textbf{Sequence}&\tiny{\textbf{FragT}\cite{DBLP:conf/cvpr/AdamRS06}}&\tiny{\textbf{IVT}\cite{DBLP:journals/ijcv/RossLLY08}}&\tiny{\textbf{$\ell_1$T}\cite{DBLP:conf/iccv/MeiL09}}&\tiny{\textbf{MIL}\cite{DBLP:conf/cvpr/BabenkoYB09}}&\tiny{\textbf{TLD}\cite{DBLP:journals/pami/KalalMM12}}&\tiny{\textbf{VTD}\cite{DBLP:conf/cvpr/KwonL10}}&\tiny{\textbf{LSK}\cite{DBLP:conf/cvpr/LiuHYK11}}&\tiny{\textbf{CT}\cite{DBLP:conf/eccv/Zhang0Y12}}&\tiny{\textbf{ASLA}\cite{DBLP:conf/cvpr/JiaLY12}}&\tiny{\textbf{$\ell_1$APG}\cite{DBLP:conf/cvpr/BaoWLJ12}}&\tiny{\textbf{MTT}\cite{DBLP:conf/cvpr/ZhangGLA12}}&\tiny{\textbf{SCM}\cite{DBLP:conf/cvpr/ZhongLY12}}&\tiny{\textbf{OSPT}\cite{DBLP:journals/tip/WangLY13}}&\tiny{\textbf{LSST}\cite{DBLP:conf/cvpr/WangLY13}}&\tiny{\textbf{Ours}}\\
\hline \hline
Car4&0.22&\textcolor{red}{0.92}&0.78&0.34&0.64&0.73&\textcolor{blue}{0.91}&0.28&0.89&0.70&0.53&0.89&\textcolor{red}{0.92}&\textcolor{red}{0.92}&\textcolor{blue}{0.91}\\
Car11&0.09&0.81&0.44&0.17&0.38&0.43&0.49&0.23&0.81&0.83&0.58&0.79&0.81&\textcolor{blue}{0.84}&\textcolor{red}{0.85}\\
Caviar&0.19&0.21&0.20&0.19&0.21&0.19&0.58&0.33&0.84&0.13&0.14&\textcolor{blue}{0.87}&0.25&0.85&\textcolor{red}{0.88}\\
David&0.19&0.72&0.63&0.45&0.60&0.53&0.72&0.56&\textcolor{blue}{0.79}&0.63&0.53&0.75&0.76&0.75&\textcolor{red}{0.81}\\
Deer&0.08&0.22&0.07&0.21&0.41&0.58&0.35&0.60&\textcolor{blue}{0.62}&0.45&0.60&0.46&0.61&0.58&\textcolor{red}{0.68}\\
Faceooc1&0.90&0.85&0.89&0.59&0.65&0.77&0.90&0.74&0.83&0.87&0.79&\textcolor{red}{0.93}&\textcolor{blue}{0.91}&0.89&\textcolor{blue}{0.91}\\
Faceocc2&0.60&0.59&0.67&0.61&0.49&0.59&0.33&0.68&0.82&0.70&0.72&0.82&0.84&\textcolor{blue}{0.86}&\textcolor{red}{0.88}\\
Football&0.57&0.55&0.11&0.55&0.56&\textcolor{red}{0.81}&0.63&0.46&0.57&0.68&0.71&0.69&0.62&0.69&\textcolor{blue}{0.72}\\
Jumping&0.14&0.28&0.55&0.53&0.69&0.08&0.09&0.07&0.24&0.59&0.30&\textcolor{red}{0.73}&0.69&0.65&\textcolor{blue}{0.70}\\
Singer1&0.34&0.66&0.70&0.33&0.41&0.79&0.52&0.34&0.81&\textcolor{blue}{0.83}&0.32&\textcolor{red}{0.85}&0.82&0.80&\textcolor{red}{0.85}\\
\hline
Average&0.33&0.58&0.50&0.40&0.50&0.55&0.55&0.45&0.72&0.64&0.52&\textcolor{blue}{0.78}&0.72&\textcolor{blue}{0.78}&\textcolor{red}{0.82}\\
\hline
\end{tabular}
\end{center}
\label{tab:asr_all}
\end{table*}

\subsection{Comparison with the State-of-the-art Trackers}
We compare our tracker against $14$ state-of-the-art algorithms on $10$ video sequences used in previous works \cite{DBLP:journals/ijcv/RossLLY08} \cite{DBLP:conf/cvpr/BabenkoYB09} \cite{DBLP:conf/cvpr/KwonL10} \cite{DBLP:conf/cvpr/LiSDH13} \cite{DBLP:conf/cvpr/HeYLWY13}. Tables \ref{tab:ace_all} and \ref{tab:asr_all} respectively show the average center location errors and the average overlap rates of different tracking methods. Our tracker outperforms other state-of-the-art tracking algorithms in most cases and especially improves the baseline ASLA \cite{DBLP:conf/cvpr/JiaLY12}. We owe this success to our learned hierarchical features.

\subsection{Comparison between DLT and Our Tracker}
We present the comparison results in terms of average center error (in pixels) between DLT \cite{DBLP:conf/nips/WangY13} and our tracker in Table~\ref{tab:comp_dlt}. We can observe that our tracker outperforms DLT on 5 of 8 sequences.

\subsection{Evaluation on Our Learned Feature's Generalizability}
\begin{figure}
\centering \epsfig{file=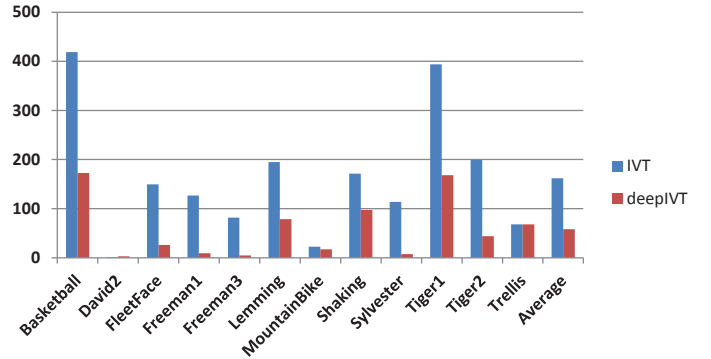,width=0.5\textwidth}
\caption{Average center error (in pixels). We compare performances between the original IVT \cite{DBLP:journals/ijcv/RossLLY08} and our tracker (deepIVT) using the learned features.}
\label{fig:ivt_deepivt_ace}
\end{figure}

\begin{figure}
\centering \epsfig{file=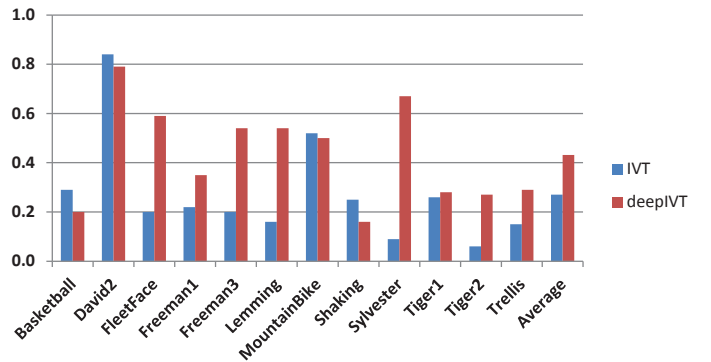,width=0.5\textwidth}
\caption{Average overlap rates (\%). We compare performances between the original IVT \cite{DBLP:journals/ijcv/RossLLY08} and our tracker (deepIVT) using the learned features.} \label{fig:ivt_deepivt_aor}
\end{figure}

To demonstrate the generalizability of our learned features, we integrate our feature learning algorithm into another baseline tracker which is called the incremental learning tracker (IVT) \cite{DBLP:journals/ijcv/RossLLY08}. We present the performances on both the original IVT and our tracker (deepIVT) in terms of average center errors and average overlap rates in Figures \ref{fig:ivt_deepivt_ace} and \ref{fig:ivt_deepivt_aor} respectively. We can observe that our tracker (deepIVT) outperforms the original IVT in most of $12$ test sequences. Due to IVT's limited performance, our tracker also misses objects in some sequences. However, the figures presented here aim to show that our learned features can boost performances of the baseline tracker. In addition, we verify our learned feature's generalizability by using $\ell_1$\_APG tracker \cite{DBLP:conf/cvpr/BaoWLJ12} and evaluating performances on the same $12$ sequences as used for IVT. As shown in Tables \ref{tab:ace_motion} and \ref{tab:aor_motion}, $\ell_1$\_APG can hardly handle these challenging sequences with complicated motion transformations. In contrast, integrating our learned features into $\ell_1$\_APG can succeed to track objects in $6$ (David2, FleetFace, Freeman1, Freeman3, MountainBike and Sylvester) of $12$ sequences. Therefore, we can conclude that our learned features are not only beneficial to ASLA \cite{DBLP:conf/cvpr/JiaLY12}, but also generally helpful to other trackers.

\begin{table}
\caption{Comparison between DLT \cite{DBLP:conf/nips/WangY13} and our tracker on $8$ sequences. The better results are shown in red fonts. ``BetterCount" means the number of sequences on which the performance of the current tracker is better than the other one.} \vspace{-0.1in}
\begin{center}
\tabcolsep 0.065in
\scriptsize
\begin{tabular}{|c|c|c|c|c|c|c|c|c|c|}
\hline
&\tiny{\textbf{Car4}}&\tiny{\textbf{Car11}}&\tiny{\textbf{David}}&\tiny{\textbf{Deer}}&\tiny{\textbf{Shaking}}&\tiny{\textbf{Singer1}}&\tiny{\textbf{Trellis}}&\tiny{\textbf{Woman}}&\tiny{\textbf{BetterCount}}\\
\hline
DLT&6.0&\textcolor{red}{1.2}&7.1&10.2&\textcolor{red}{11.5}&\textcolor{red}{3.3}&3.3&9.4&3\\
\hline
Ours&\textcolor{red}{3.0}&1.4&\textcolor{red}{3.2}&\textcolor{red}{5.5}&15.2&4.0&\textcolor{red}{3.0}&\textcolor{red}{2.2}&\textcolor{red}{5}\\
\hline
\end{tabular}
\end{center}
\label{tab:comp_dlt}
\end{table}

\section{Conclusion}
In this paper, we propose a hierarchical feature learning algorithm for visual object tracking. We learn the generic features from auxiliary video sequences by using a two-layer convolutional neural network with the temporal slowness constraint. Moreover, we propose an adaptation module to adapt the pre-learned features according to specific target objects. As a result, the adapted features are robust to both complicated motion transformations and appearance changes of specific target objects. Experimental results demonstrate that the learned hierarchical features are able to significantly improve performances of baseline trackers.

\ifCLASSOPTIONcaptionsoff
  \newpage
\fi

\bibliographystyle{IEEEtran}
\bibliography{ref}

\end{document}